\documentclass[final,3p,times]{elsarticle}
\usepackage{amsmath,amsfonts}
\usepackage{algorithmic}
\usepackage{array}
\usepackage[caption=false,font=normalsize,labelfont=sf,textfont=sf]{subfig}
\usepackage{textcomp}
\usepackage{stfloats}
\usepackage{url}
\usepackage{verbatim}
\usepackage{tikz} 
\usepackage{graphicx}
\usepackage[utf8]{inputenc}
\usepackage[percent]{overpic} 

\usepackage{caption}
\def\BibTeX{{\rm B\kern-.05em{\sc i\kern-.025em b}\kern-.08em
    T\kern-.1667em\lower.7ex\hbox{E}\kern-.125emX}}
\usepackage{balance}
\usepackage{xcolor}
\newcommand{\myRed}[1]{\textcolor{black}{#1}}
\newcommand{\mygreen}[1]{\textcolor{black}{#1}}
\newcommand{\minus}{\scalebox{0.5}[1.0]{$-$}}

\usepackage{amssymb}
\usepackage{amsmath}

\journal{Mechanism and Machine Theory}

\begin{document}

\begin{frontmatter}



\title{OmniQuad: a Wheeled-Legged Hybrid Robot with Omnidirectional Wheels}

\author[dii,piaggio]{ Francesco Iotti}
\author[dii,piaggio]{Alok Ranjan}
\author[dii,piaggio]{Franco Angelini}
\author[dii,piaggio]{Manolo Garabini}

\affiliation[dii]{organization={Department of Information Engineering of University of Pisa},
            addressline={Via Girolamo Caruso 16 }, 
            city={Pisa},
            postcode={56122}, 
            country={Italy}}
\affiliation[piaggio]{organization={Research Center Enrico Piaggio},
            addressline={Largo Lucio Lazzarino 1},
            city={Pisa},
            postcode={56122}, 
            country={Italy}}

\begin{abstract}
Combining wheeled and legged locomotion within a single robotic platform offers the potential to harness the advantages of both systems. The development of such hybrid systems remains an active area of research. While legged robots exhibit remarkable adaptability for traversing uneven and rugged terrains, their relatively low speed limits their practicality in indoor environments. Conversely, wheeled robots demonstrate higher efficiency on flat surfaces but often encounter difficulties when navigating obstacles such as steps. This paper introduces \textit{OmniQuad}, a novel hybrid robot that integrates mecanum wheels with legged locomotion to exploit the benefits of both modalities. Mecanum wheels provide omnidirectional mobility, enhancing the robot's manoeuvrability by enabling agile movement in confined spaces and the ability to maintain a consistent orientation during task execution. In this work, we first present the design of a custom \textit{wheeled-foot} featuring mecanum wheels. Subsequently, we describe the complete robot assembly, comprising these wheeled feet and EM-Act actuators. Finally, we detail the control scheme developed to govern the OmniQuad and evaluate its performance through experimental trials conducted in both indoor and outdoor environments.
\end{abstract}



\end{frontmatter}


    
    
    
    
    

\vspace{-2mm}
\section{Introduction}
The demand for mobile robots is substantial across diverse applications in logistics, cleaning, search and rescue, and environmental monitoring \cite{mob_rob_apl}. 
Although the designs of these robots differ from one another, they are still limited in their ability to navigate tight spaces and interact with humans and other robots due to their lack of human-like locomotion capabilities.

Wheeled mobile robots, owing to their simple design and implementation, are extensively used in these settings. Their efficacy ensues from the ease of achieving high speeds and executing movements with good efficiency. These machines were originally designed for planetary exploration \cite{rover_mars, Lunar_rover}, but in recent years, they have also been utilized in warehouse work \cite{warehouse_robot} and homes. Robots equipped with conventional wheels face a significant limitation in manoeuvrability due to their restricted movement capabilities. These robots are only able to move along a direction that is perpendicular to the centre axis of the wheels, which makes it difficult for them to navigate through tight spaces or turn quickly \cite{TAHERI2020103958}.

\begin{figure}[!t]
\centering
\includegraphics[width=3in ,height = 2.3in, trim = 0.2in 0.3in 0.2in 0.2in]{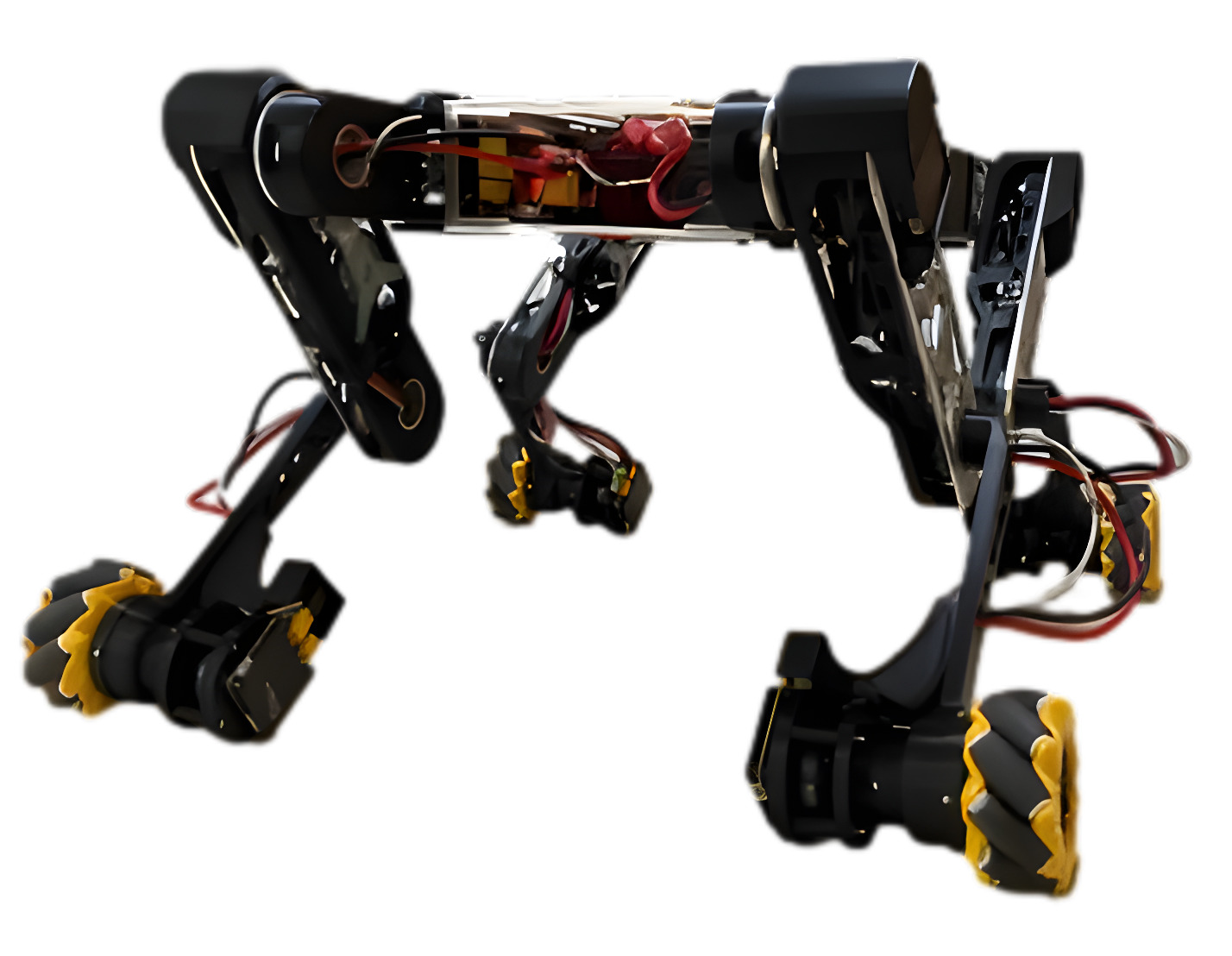}
\caption{OmniQuad hybrid robot.}
\label{omniquad}
\vspace{-2mm}
\end{figure}
Various types of wheels have been developed to facilitate omnidirectional movement, including omni wheels and mecanum wheels. Recent years have seen the integration of omni wheels in different typologies of robots, as referenced in works such as \cite{1242089, three_omni}. These wheels are distinctively mounted at a 45-degree angle concerning their fore axis, setting them apart from other wheel types. Their major problems are the low traction and stability. In addition, they are also susceptible to vibrations and slippage. In \cite{four_omni}, they reduce these kinds of issues by designing a specific suspension system in a four-wheeled omni-wheelchair.

Mecanum wheels \cite{ilon1975wheels} consist of rollers oriented at a 45-degree angle to the fore axis. Their utilization is well-suited for scenarios necessitating high stability and manoeuvrability \cite{mec_intro, SUN2021107128}. Additionally, their capacity to accommodate high load capacities makes them versatile across diverse application domains, such as factories, warehouses and even hospitals, as highlighted in \cite{5980575, mec_apl}. It is essential to note that while omnidirectional wheels offer advantages, they often generate substantial friction forces due to pronounced vibrations during motion. This can result in slippage, impacting overall efficiency during both forward and lateral movements. Attempts to mitigate these challenges include incorporating springs into wheel designs, as addressed in \cite{des_opt_mec}. Moreover, the optimal performance of these wheels is typically observed in flat terrain scenarios. Novel designs for mecanum wheels have been proposed, aiming to optimize motion in rough terrain \cite{5442595}, whereas in \cite{ZHANG201518} they proposed a novel tracked omnidirectional vehicle to achieve omnidirectional motion on challenging terrains. An inherent limitation of wheeled locomotion lies in the inability of such robots to surmount obstacles, a challenge prevalent in both natural and indoor environments, particularly when encountering steps or uneven terrains.

Quadrupedal robots emerge as more adept solutions for rugged terrains due to their ability to overcome obstacles through climbing or jumping manoeuvres \cite{review_quadruped, NI, 6930798}. The major examples of legged robots, also known as quadrupedal robots, are Spot \cite{8913831}, ANYmal \cite{7758092}, Mini Cheetah \cite{8793865}, HyQ \cite{doi:10.1177/0278364915578839}, Oncilla \cite{Sprwitz2018OncillaRA} and Cheeta-cub \cite{Sprwitz2013TowardsDT}. However, in some scenarios, their speed may be slower compared to wheeled robots, and their operational efficiency may fall short \cite{Leg_system}.
Choosing the best locomotion system can be a challenging task. To overcome this, hybrid locomoted robots are being developed to combine the benefits of two different locomotion systems. Currently, there are various types of wheeled-legged hybrid robots \cite{survey} that offer the efficiency of wheeled locomotion and the ability to navigate through rough terrains and climb over obstacles.
Vehicles with omnidirectional wheels and legged systems are often viewed as fundamentally distinct due to their specialized use in different scenarios. Designing a system that seamlessly integrates vehicles with omnidirectional wheels and legged systems poses a complex challenge. It entails not only the mechanical merging of the two components but also the development of a control system that accommodates their unique operational environments. The goal is to create a robot that harnesses the unique advantages of both technologies, enabling it to excel in a diverse array of scenarios beyond the capabilities of individual systems. \mygreen{By combining wheeled and legged locomotion, this integrated approach allows the robot to surpass the performance limitations of conventional single-mode systems across diverse and demanding environments. Specifically, the robot could achieve exceptional maneuverability within structured environments, such as warehouses, factories, or urban infrastructures, where omnidirectional wheels provide rapid, precise positioning and the ability to maintain a consistent orientation, beneficial for operations like inventory handling or inspection tasks. Furthermore, the incorporation of legged locomotion extends the robot's capabilities to negotiate uneven terrains, climb stairs or obstacles, and traverse confined spaces, tasks that are typically challenging for traditional wheeled vehicles. The robot's legs also allow it to dynamically adjust its height, facilitating passage beneath obstacles or through low-clearance areas, enhancing its overall operational versatility.}

The contributions of this paper are as follows: (i) Mechanical design of an Omniwheeled foot that incorporates a motor and a gearbox within the assembly; (ii) a robotic system called OmniQuad, as shown in Fig. \ref{omniquad}, which integrates both mecanum wheels and legs; (iii) a control system that enables the use of both legs and omnidirectional wheels; and (iv) an evaluation of the capabilities of the developed robot through a series of tasks in different environments. These tasks focused on navigating obstacles using various methods, such as passing beneath or scaling over them, taking advantage of hybrid locomotion. The robot's omnidirectional movement capabilities were also examined through tasks requiring fixed orientation while performing various manoeuvres. In outdoor environments, the robot followed a grid-patterned path to simulate an environmental survey.

\vspace{-2mm}
\section{Literature Review}
In recent years, various types of mobile robots have been developed to navigate difficult terrains or overcome obstacles in structured environments, such as steps.

In \cite{omniwheg} a novel robotic system is presented with transformable mecanum wheels that overcome obstacles up to 2.5 times the radius of its wheels. The primary problems with this concept are related to the intricate method needed to overcome barriers, which includes an alignment stage for the transformable wheel to latch onto the obstruction. Furthermore, it appears that the robot has only been tested on obstacles with sharp edges. To simplify the climbing on obstacles in \cite{10502024}, the authors developed a new type of transformable wheel that can passively transform when in contact with the obstacle, avoiding including a motor to transform the wheel. The main limitations are the size and dimension of the obstacle, which can limit the motion even in indoor environments. \mygreen{In \cite{Zheng2024} the authors introduce \textit{$\alpha$}-WaLTR, a four-wheeled robot with passively transformable wheels and spring suspension. This mostly passive setup lets the robot glide over modest obstacles and even tackle stairs using straightforward control. However, the passive design may limit its adaptability on highly uneven or complex terrains.}

Quadruped robots have improved in recent years in terms of their ability to traverse uneven terrains, although problems can still arise indoors. To increase the performance in climbing obstacles or improve the forward velocity of the robot, diverse types of wheeled-legged robots have emerged, each showcasing distinct design and operational characteristics \cite{survey, 9734076}. \mygreen{Currently, advanced quadruped robots, such as the Go2-W from Unitree Robotics\footnote{Unitree Robotics Go2-W: \url{https://www.unitree.com/go2-w}} and the Lynx developed by Deep Robotics\footnote{Deep Robotics Lynx: \url{https://www.deeprobotics.cn/en/index/lynx.html}}, have demonstrated remarkable capabilities in traversing challenging terrains swiftly and reliably compared to traditional legged and wheeled robots.}
In the research community, numerous wheeled-legged robotic systems have been developed to improve locomotion capabilities. In \cite{9204432}, the authors developed a quadruped robot that utilizes passive wheels for roller skating locomotion. However, they do not fully exploit the potential benefits of incorporating wheeled locomotion, and the increase in performance it may offer. In \cite{9794482, 8998367}, the authors present wheel-leg rovers with only one actuated joint per leg. This design is more suitable for movement on slopes because passive joints are easier to adapt to the irregularities of different terrains, but make it impossible for the robot to overcome obstacles by lifting its legs because it has just one actuated joint per leg. In \cite{NI2022} researchers developed a wheel-legged robot with 11 DOF. Although the robot can handle various types of uneven terrain using an IMU sensor-based control system, it lacks the ability to climb obstacles with its legs. The HyTro \cite{Hytro} robot is equipped with wheels placed under the body. Thanks to this design, it can switch between wheeled and legged locomotion by lowering or lifting its legs, but it cannot perform hybrid locomotion. 

\mygreen{Robots that feature wheels on the knees enable switching between legged and wheeled locomotion by lifting or lowering their feet \cite{wheels-knees}; similarly, the robot Max, introduced in \cite{zhou2024}, employs knee-mounted wheels with a specialized mechanism to alternate between agile legged movement and efficient wheeled mobility, yet it shares the limitation of being unable to perform simultaneous hybrid locomotion.}

The robot in \cite{CHEN2021367, bitnaza} is equipped with six wheel-legs. When it encounters an obstacle, it can lift two legs and use the other four for support. However, its high weight limits the efficiency, in terms of electrical power and execution time, of its leg locomotion, so it mainly relies on wheel movements. Another innovation is a hexapod with mecanum wheels \cite{mec_hexa}. The benefit of having six wheels is the total stability of the system and redundancy, which allows it to move even if a limited number of wheels fail. The challenge with this design is the overall structure, as well as the planning of the gait for six legs.

The Centauro \cite{Centauro} and Momaro \cite{momaro} robots are used in disaster management and integrate wheels instead of feet to increase their speed of action. They use 2 degrees of freedom (DOF) wheels to enable omnidirectional movements. Both robots integrate two 7 DOF robotic arms, but Momaro has four 4 DOF legs, whereas Centauro has 5 DOF legs, which contribute to higher manoeuvrability in the legged locomotion to move in uneven terrains. They are almost human-sized and can be bulky for certain applications in confined spaces.

\mygreen{The authors in \cite{Ranjan2025} introduce TraQuad, a multimodal robot combining legged and tracked locomotion, enhancing agility, maneuverability, and traction, especially in outdoor environments with complex and uneven terrains. However, this design is less efficient indoors, where tracked mobility and advanced climbing capabilities might be unnecessary or even disadvantageous. Moreover, the configuration of the robot with 8 degrees of freedom increases mechanical complexity, making it challenging to overcome common indoor obstacles such as stairs or steps.}

Wheel-leg robots with just two legs and two wheels \cite{9561579} are easier to control because of their simple design. However, they encounter difficulties when overcoming obstacles, as they can simply jump over them. The Swiss-mile robot \cite{9636371} is an advancement of the ANYmal robot with torque-controlled and non-steerable wheels. This robot is versatile and can be used both indoors and outdoors, thanks to its hybrid locomotion. Regular wheels enhance the robot's efficiency during forward movements, although they can pose challenges for steering in specific situations, potentially impacting the overall system performance. A novel wheeled-legged robot with magnetic omni wheels has been developed in \cite{omni_feet}, which can climb wall surfaces. The robot stays fixed to an inclined plane with special magnetic omni wheels, but it is more suitable for climbing vertical obstacles than overcoming them. In addition, a new omnidirectional platform with legs has been proposed in \cite{robo_cool}. The platform can carry a weight of up to 80 kg. They developed a novel planetary gear to reduce vibration and achieve agile motion. However, this robot is developed to substitute the legged system in humanoid robots, not to be an independent system.

\mygreen{In this work, we present a novel wheeled-legged robot capable of utilizing both omnidirectional wheels and legs. Our system integrates mecanum wheels to significantly enhance maneuverability, enabling the robot to avoid steering and maintain body orientation without exclusively relying on legs for rotation, as required by the Swiss-mile robot \cite{9636371}, Go series from Unitree\footnote{Unitree Robotics Go2-W: \url{https://www.unitree.com/go2-w}}, and Lynx from Deep Robotics\footnote{Deep Robotics Lynx: \url{https://www.deeprobotics.cn/en/index/lynx.html}}. Additionally, in contrast to robots such as Centauro \cite{Centauro}, Momaro \cite{momaro}, we achieve omnidirectional mobility using wheels with fewer degrees of freedom, thereby simplifying the mechanical design. Furthermore, our legged locomotion addresses the inherent limitations of purely wheeled systems, such as those encountered with transformable wheel designs, by enabling navigation over uneven terrain, obstacle climbing, and height adjustments to traverse under obstacles. This hybrid approach provides distinct advantages in structured environments for precise orientation and versatility in complex, obstacle-rich scenarios.}
 \begin{figure}

    \centering
    \subfloat[]{\includegraphics[width= 1.0 in]{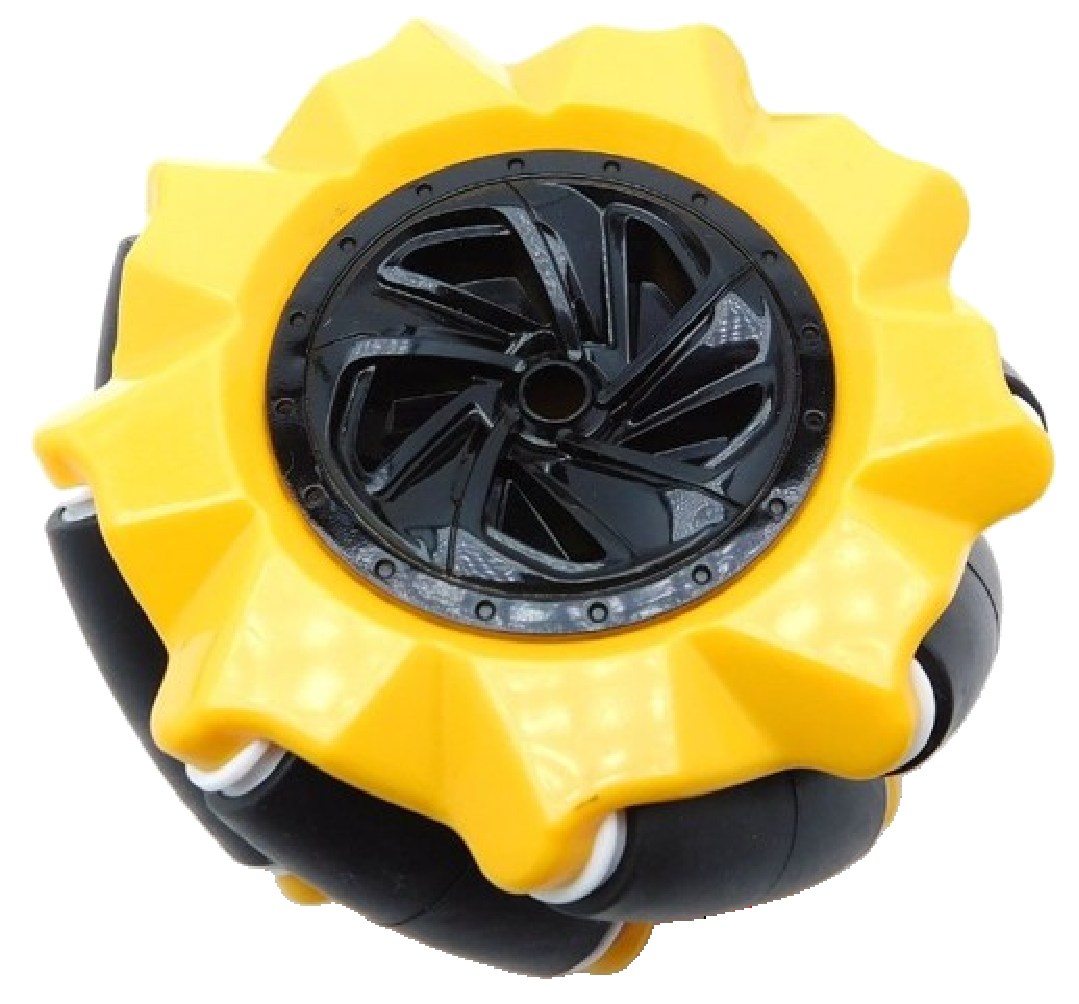} 
    \label{mec_wheel}}\qquad
    \subfloat[]{\includegraphics[width= 1.0 in]{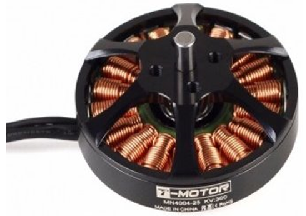}
    \label{t_motor}}\qquad
    \subfloat[]{\includegraphics[width= 1.0 in]{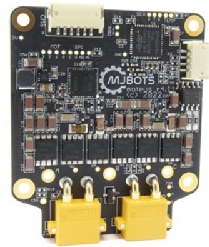} 
    \label{moteus}}\qquad
    \subfloat[]{\includegraphics[width= 1.0 in]{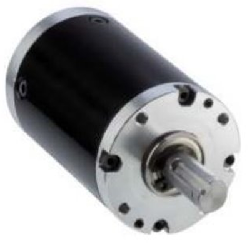} 
    \label{gearbox}}
    
    \caption{\myRed{Components in the actuation system of the wheeled-foot: (a) mecanum wheel; (b) T-motor antigravity 4004; (c) Moteus r.411 ; (d) AASUN AM-28P gearbox.}}
    \label{components}

\end{figure}

\vspace{-2mm}

\section{OmniQuad Design and Electronics}
\label{Design}
This section provides a comprehensive overview of the actuation system and its components, introduces the wheel-foot design with a focus on its features and capabilities, and presents the OmniQuad robot, showcasing all of its characteristics.

\vspace{-2mm}
\subsection{Actuation System}
To enable omnidirectional movement of the robot on wheels, we opt for mecanum wheels instead of Omni wheels due to their ease of implementation, mounting direction, and actuation. To maintain the low weight of the robot, wheels made of a majority of plastic are chosen. An analysis of wheels with varying diameters is conducted to determine the most effective solution in terms of stability and minimizing the total torque to move the robot.

For the motor, we select a brushless option, named Antigravity MN4004 KV300, presented in Fig. \ref{t_motor}, due to its lightweight and high performance. Additionally, to meet the necessary torque requirement, we estimate the total force to move the robot  

\begin{equation}
\label{deqn_ex3}
        F_{tot} = F_{a}+F_{r}, 
\end{equation}
where $F_{a}$ is the acceleration force that can be given as
\begin{equation}
\label{deqn_ex1}
        F_{a} = m\ a.
\end{equation}
In our case, the total mass of the robot is 7 kg, with an acceleration set to 4 m/s\textsuperscript{
2} to achieve the desired nominal speed of 2 m/s. The term \(F_{r}\) denotes the rolling friction force, which is considered in the torque calculations to account for energy losses due to wheel-ground interactions. The rolling friction force can be given as
\begin{equation}
\label{deqn_ex3}
        F_{r} = c\ W.
\end{equation}
Here, $W$ is the normal force perpendicular to the surface on which the wheel is rolling and $c$ is the coefficient of rolling friction. It is challenging to assign a value to this coefficient, due to its dependence on unrelated factors such as the total weight of the wheel, the roller's material, and the deformation of the terrain \textcolor{black}{\cite{Tlale2008}}. Therefore, a safety factor is incorporated, resulting in a total value of 3 for the coefficient of rolling resistance.

The total torque depends on the wheel's radius, which consequently sets a constraint on the wheel's size. We chose a mecanum wheel with an 80mm diameter from DFRobot\myRed{\footnote{DFRobot website: \url{https://www.dfrobot.com/}}} (as shown in Fig. \ref{mec_wheel}), resulting in a total torque of 10.5 Nm. 

A planetary gearbox is selected due to its compact design and high gear ratio. Specifically, we opted for an AASUN AM-28P planetary gearbox, shown in Fig. \ref{gearbox}, with two stages and a total gear ratio of 15. Such a gear ratio allows the system to generate high torque, capable of handling the significant forces that occur when moving with the mecanum wheels.

The moteus r.411 controller\footnote{moteus GitHub page: \url{https://github.com/mjbots/moteus}}, displayed in Fig. \ref{moteus}, is selected to command the motor. This driver allows multiple servo systems to be connected in series using the daisy-chain connection method. This feature enables seamless integration of various servo systems, providing a flexible and efficient solution for controlling multiple servo systems.

\begin{figure}[ht]
    \centering
    \includegraphics[width=0.8\textwidth ]{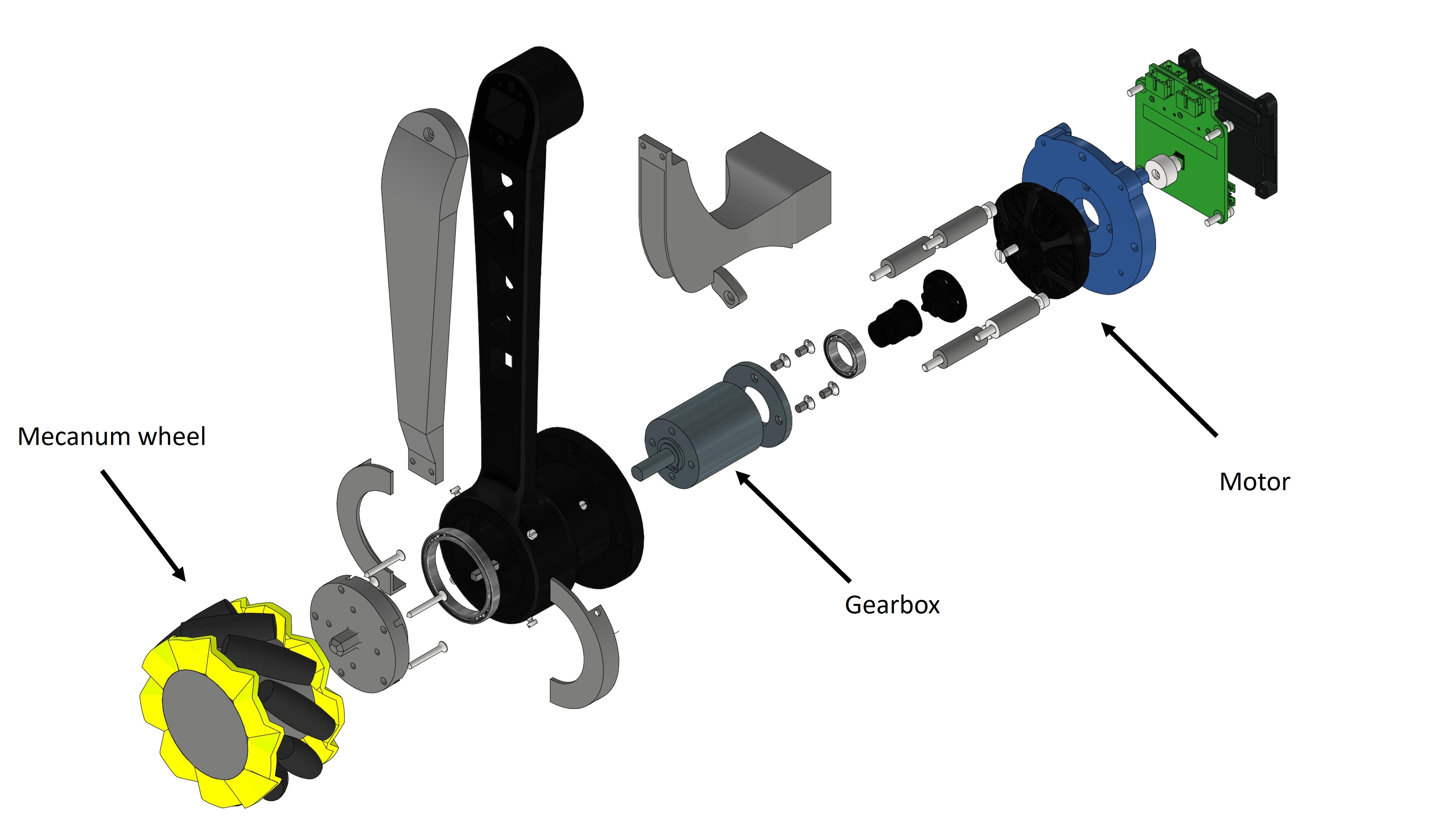}
    \caption{\myRed{Exploded view of the wheeled-foot.}}
    \label{Expl_view}
\end{figure}
\vspace{-2mm}
\subsection{Design of the Wheeled-Foot}
The design incorporates all mechanical components, the motor, the gearbox, and the mecanum wheel, aligned on a single axis. The total width from the controller to the wheels measures 150 mm. The mechanical interfaces are made of ABS plastic, which is appropriately fabricated with FDM technology. In Fig. \ref{Expl_view}, the exploded view of the wheeled foot is depicted. In this solution, the link encapsulates the gearbox. The motor is secured to a plate that is connected directly to the bottom of the link. The controller is fastened to the plate below the motor, and uses a magnet to control it. The motion from the gearbox to the wheel is transmitted through a plastic interface. This interface is attached to the wheel using a hexagonal shaft and screws specifically for plastic. On the other side, the interface is attached to the gearbox shaft using a key made of Aluminum. A bearing is also placed to support the relative motion between the fixed part, the link, and the rotating parts of the actuation system. Additionally, the bearing is used to prevent the wheel from coming out of the system, using flanges to prevent it from moving out. Following the assembly of the mechanical and electronic components, a casing is installed to protect the power and data cables. Finally, the system presents a total length and width of respectively 230 mm and 150 mm, and the total weight is 0.42 kg, as reported in Tab. \ref{spec_wheel-foot}.

\begin{table}[h]
\begin{center}

\caption{\myRed{Mechanical specifications of the wheeled-foot}}
\label{spec_wheel-foot}
\begin{tabular}{ c@{\hspace{1.0cm}}l@{\hspace{2.5cm}} l@{\hspace{2.0cm}} }
\hline
\textbf{S. No} & \textbf{Property} & \textbf{Value}\\
\hline

1 & Length & 230 mm\\

2 & Width & 150 mm\\ 

3 & Weight & 0.420 kg\\

4 & Motor power & 300 W\\

5 & Gearbox ratio & 15:1\\

6 & Max torque & 18 Nm\\
\hline
\end{tabular}
\end{center}
\vspace{-5mm}
\end{table}

\color{black}
\vspace{-2mm}
\subsection{OmniQuad}
\begin{figure*}[!t]
    \includegraphics[width= \linewidth] {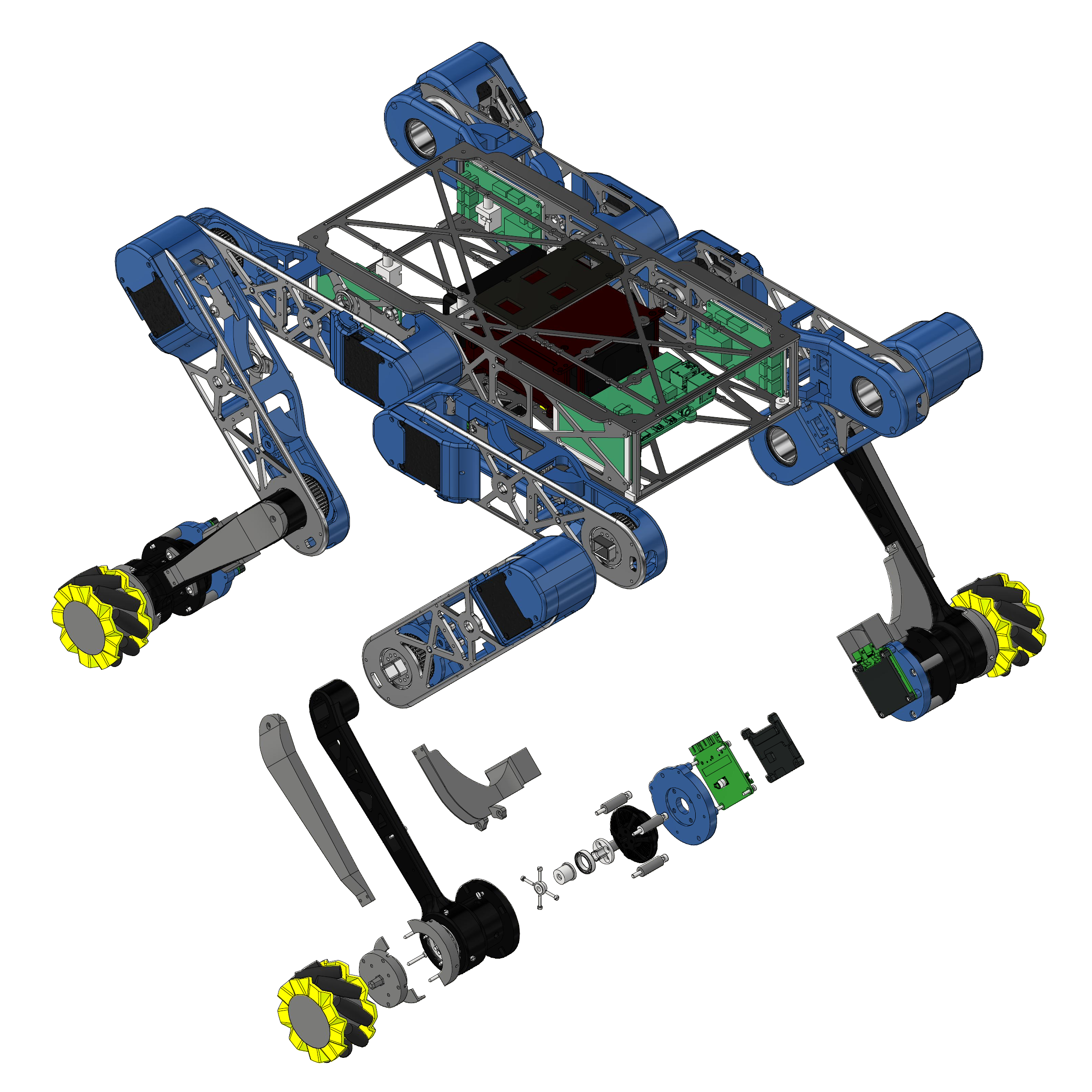}
\vspace{0.1in}
\caption{ 3-D CAD model of the OmniQuad robot, showcasing different components in the leg design.}
\label{cad_robot}
\vspace{-2mm}
\end{figure*}

The complete robotic system is equipped with four legs, each featuring a \textit{Wheeled-Foot} assembly as an end-effector that incorporates a mecanum wheel. Each leg consists of two distinct EM-Act actuators \cite{Em-act}—one to control the hip joint and another to control the knee joint. The Wheeled-Foot assembly includes a third actuator, as illustrated in Fig. \ref{cad_robot}. Integrating mecanum wheels with a 2-DOF leg introduces the capability for lateral movement, a functionality traditionally achieved in quadrupedal robots through the abduction/adduction of the hip joint. 

These components are mechanically interconnected via a rectangular interface, ensuring structural integrity and ease of assembly. Each leg is subsequently mounted onto a cross-linked aluminium chassis with a box-shaped configuration, providing a robust foundation. Within this structural framework, the electronic components necessary for controlling the entire robotic system are mounted. A summary of the design specifications is presented in Table \ref{spec_omni}.

\begin{table}[h]
\begin{center}

\caption{\myRed{Mechanical specifications of the OmniQuad robot}}
\label{spec_omni}
\begin{tabular}{  c@{\hspace{1.0cm}}l@{\hspace{3.0cm}} l@{\hspace{2.5cm}} }
\hline
 \textbf{S. No.} & \textbf{Property} & \textbf{Value}\\
\hline
1 & Nominal length & 542 mm\\

2 & Width & 498 mm\\ 

3 & Maximum height & 460 mm\\

4 & Minimum height & 180 mm\\

5 & Weight & 7 kg\\

6 & Peak leg joint torque & 19.6 Nm\\

7 & Peak wheel joint torque & 18 Nm\\
\hline
\end{tabular}
\end{center}
\vspace{-5mm}
\end{table}


\vspace{-2mm}

\subsection{Electronics}
The OmniQuad employs a total of twelve motors, three on each leg, and the power requirements for the entire system are met by a DJI 47D battery\myRed{\footnote{DJI store website: \url{https://www.dji-store.it/}}}, with a capacity of 4500mAh. To distribute electrical power evenly among the three motors of each leg, four different power distributors are used. These distributors are called Mjbots power dist r4.5b\myRed{\footnote{mjbots-power-dist-r4-5b: \url{https://mjbots.com/products/mjbots-power-dist-r4-5b}}}. Each leg's two motors, responsible for leg motion, are directly linked to the dedicated power distributor for that particular leg. Meanwhile, the motor of the mecanum wheel is connected to the motor of the knee in a daisy-chain manner.


The Raspberry Pi 4 within the mjbots pi3hat r4.5\footnote{mjbots-pi3hat-r4-5: \url{https://mjbots.com/products/mjbots-pi3hat-r4-5}} \color{black} serves as the central control unit for the entire system. Specifically, the Pi3hat facilitates the implementation of low-level control over the robot's joints and enables data acquisition from sensors interfaced with the driver. The three drivers within each leg are interconnected in a daisy chain configuration through JST data cables. Consequently, each leg is linked to a channel of the Pi3hat for control and data acquisition. Leveraging the Wi-Fi module of the Raspberry Pi, it is possible to establish a connection with a computer and execute Python code to individually control each motor and acquire sensor data from the driver.

\section{Modelling and Control}
In order to gain a comprehensive understanding of the robot's motion, it is important to examine its kinematics in two distinct modes of operation. One mode involves the use of wheels for movement, while the other mode involves the use of legs. The following section provides an analysis of these two distinct kinematic models that are employed in the entire system.
Additionally, the challenges associated with transitioning the robot between different types of locomotion are examined. In conclusion, the dynamic model of the entire system is introduced.
\vspace{-2mm}
\subsection{Omnidirectional Vehicle}
\begin{figure}[!t]
\centering
\includegraphics[width=3.8in]{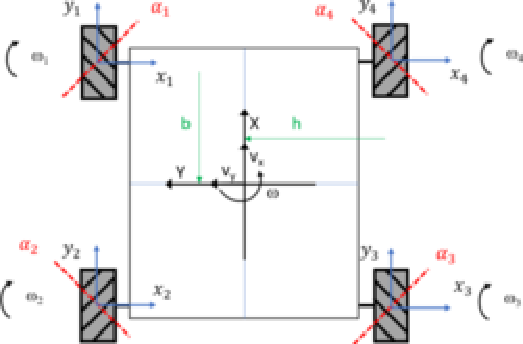}
\caption{\myRed{Kinematic model of a four mecanum-wheeled vehicle.}}
\label{Mec_veh}
\end{figure}
An omnidirectional vehicle is equipped with 4 mecanum wheels as shown in Fig. \ref{Mec_veh}. To facilitate both forward and lateral movements, the wheels with the same roller's rotational axis are positioned along the diagonals of the robot. Notably, every wheel is independently actuated by a dedicated motor. The angular velocities of these mecanum wheels are denoted by $\omega_{i} (i = 1, 2, 3, 4)$, and collectively they govern the platform's overall angular and linear velocity, denoted respectively as $\omega$ and $V$, which is decomposed, along $X$ and $Y$ axis, in $V_{x}$ and $V_{y}$. Whereas the directions of the roller's velocity are denoted by $\alpha_{i}$, which determines the force vector of each wheel. Essential parameters governing the wheel dynamics enclose the angle of the rollers ($\gamma$), the radius of the wheel ($R$), and the radial distance of each wheel from the robot's centre ($r$), decomposed in ($b$), along the $x$ axis and ($h$), along the $y$ axis. The forward kinematic equations characterizing the locomotion of a Four-Wheeled Mecanum vehicle \cite{9305643} are reported in (\ref{wheel_kin}). 
\begin{equation}
\label{wheel_kin}
\left[
\begin{array}{c}
     V_{x}  \\
     V_{y} \\
     \omega
\end{array} \right]=\frac{R}{4}
\left[
\arraycolsep=1.5pt\def\arraystretch{1.2} 
\begin{array}{cccc}
1 & 1 & 1 & 1 \\
\minus\tan(\gamma) & \tan(\gamma) & \minus\tan(\gamma) & \tan(\gamma) \\
k(\gamma) & k(\gamma) & \minus k(\gamma) & \minus k(\gamma)
\end{array} \right]
\left[
\begin{array}{c}
     \omega_{1}  \\
     \omega_{2} \\
     \omega_{3} \\
     \omega_{4}
\end{array} \right],
\end{equation}
\myRed{where} $k(\gamma)$ is defined as 
\begin{equation*}
    k(\gamma) = \frac{1}{b - h \tan(\gamma)}.
\end{equation*}
\myRed{Each} wheel's force vector is perpendicular to the orientation of its mecanum rollers, allowing the combined effect of all wheels to confer omnidirectional mobility to the platform. The configuration of all four wheels imparts the capability for forward, lateral, and rotational motions, affording a comprehensive range of navigational possibilities.

\begin{figure}[t]
\centering
\begin{minipage}[t]{0.45\textwidth}
\centering
    \subfloat[]{\includegraphics[width=0.99\linewidth]{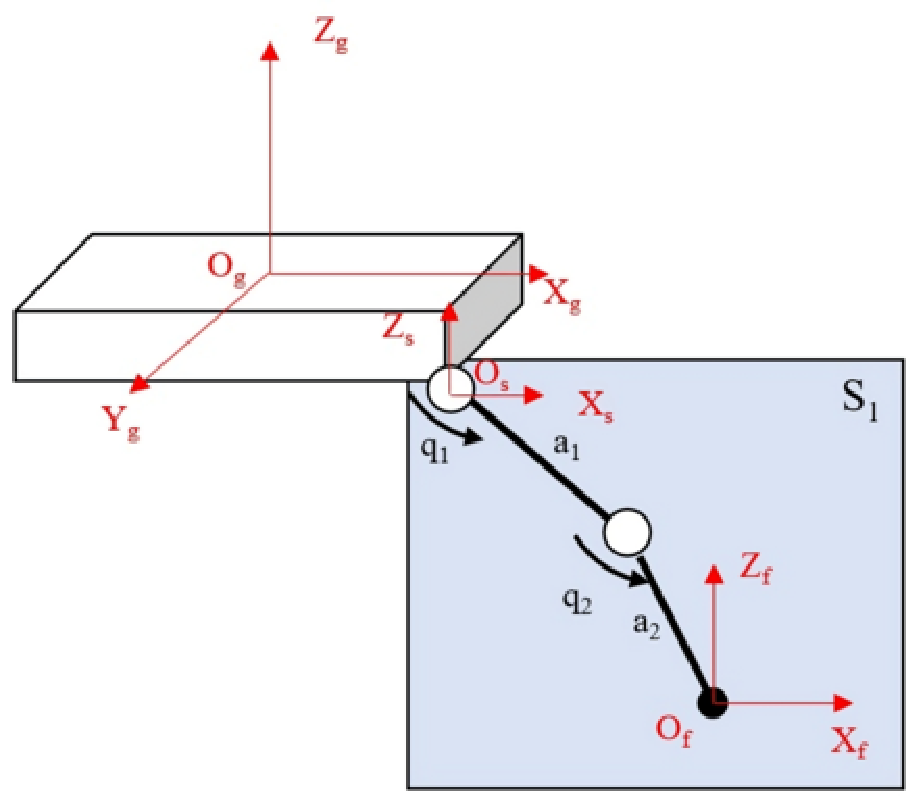}
\label{leg_sys}}
\end{minipage}
\hfill
\begin{minipage}[t]{0.45\textwidth}
\centering
    \subfloat[]{\includegraphics[width=0.99\linewidth]{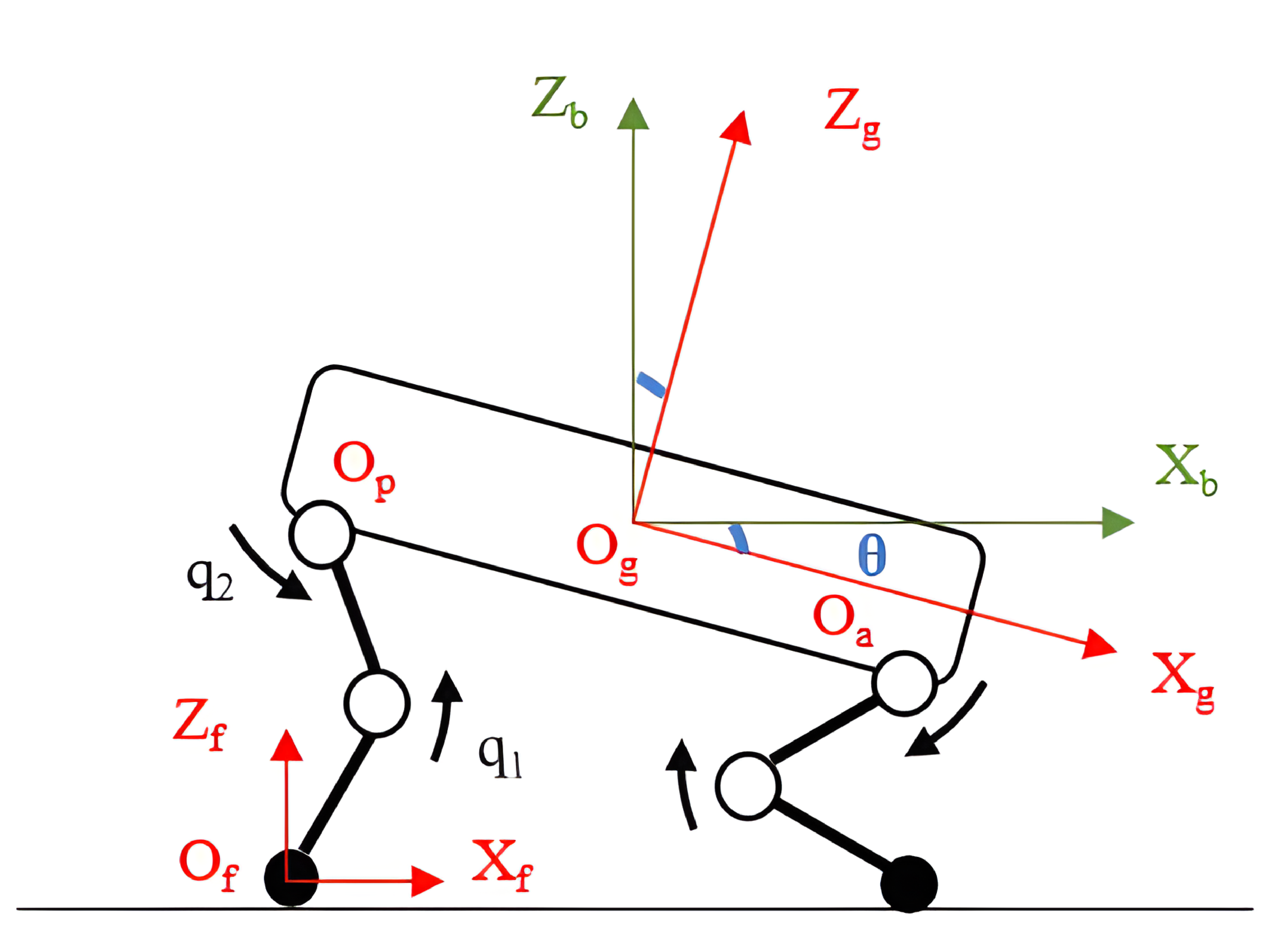}
    \label{pitch_fig}}
    
\end{minipage}
\caption{(a) Kinematic diagram of a single leg of the robot, (b) Kinematic diagram of the legged system when the base changes its pitch angle.}
\label{legged}
\vspace{-2mm}
\end{figure}
\vspace{-2mm}

\subsection{Legged System}
\label{legged}
The robotic system's capacity to surmount obstacles and vary its base elevation is facilitated by its legged architecture, characterized by two revolute joints per leg. As shown in Fig. \ref{leg_sys}, the centre of gravity of the base \(O_{g}\) can move in a three-dimensional space, while a single leg operates on plane $\mathbf{S}_{1}$ independently.

When all four legs are synchronized in their movement, the base's height can be adjusted, thereby mitigating any tilting or imbalance. Alternatively, by independently controlling the motion of a single leg, the robot can deliberately raise one of its feet, allowing it to overcome obstacles. Executing a singular leg movement at a time enables the robot to sustain a stable configuration during its task. A primary objective is to determine the leg pose to position the foot effectively within the designated task space. Within the two-dimensional plane, $\mathbf{S}_{1}$, the forward kinematic model of a two-revolute joint manipulator is reported below.
\begin{equation}
\label{for_kin_rr}
{O_{f}^{s}} = 
    \begin{cases}
    l_{1} \cos(q_{1}) + l_{2} \cos(q_{1}+ q_{2}), & \text{along $X$ axis} \\
    l_{1} \sin(q_{1}) + l_{2} \sin(q_{1}+ q_{2}), & \text{along $Z$ axis} 
    \end{cases}
\end{equation}
\myRed{where} $l_1$ and $q_1$ are respectively the length and the joint angle of the first link, whereas $l_2$ and $q_2$ refer to the second link. Using the Denavit-Hartenberg convention \cite{10.1115/1.4011045}, it is possible to refer to the coordinates of the foot in the shoulder frame $O_{s}X_{s}Z_{s}$. To solve the inverse kinematics problem within this same two-dimensional plane, $\mathbf{S}_{1}$, a possible approach involves the application of the Law of Cosines \cite{robot}. 

Additionally, the coordinated motion of the robot's four legs can stabilise its centre of mass to prevent it from falling. The pitch angle of the base can be adjusted, which is particularly useful when the robot needs to stabilise itself while only three legs are in contact with the ground. In this scenario, the configuration of the legs is reversed: the end-effector is now the shoulder joint, $O_{p}$ and $O_{a}$, which is attached to the robot's base. To compute the pitch angle, the combined motion of the front and back legs is required, Fig. \ref{pitch_fig}. Using \eqref{for_kin_rr}, we can calculate the $X$ and $Z$ components of the robot's shoulder concerning the foot frame $O_{f}X_{f}Z_{f}$. The positions of the shoulders, $O_{p}$ and $O_{a}$, can be defined in a frame located at the robot's centre of mass, denoted as $O_{b}X_{b}Z_{b}$. This frame is oriented in the same direction as the robot when the pitch angle is zero. It is important to note that in this scenario, the foot frame and the robot frame share the same orientation. They are merely translated relative to each other based on the dimensions of the robot's base. With this information, the actual pitch angle $\theta$ can be computed as:
\begin{equation}
\label{pitch}
\theta = \arctan{\frac{Z_{o_{a}}^{b} - Z_{o_{p}}^{b}} {X_{o_{a}}^{b} - X_{o_{p}}^{b}}}.
\end{equation}

\myRed{It} is also possible to adjust the robot's height by synchronising the motion of all four legs simultaneously, setting the same $Z$ components for all the legs of the robot.

\subsection{Hybrid System}

The transition from legged locomotion to omni-wheeled movement presents significant challenges in ensuring stability and efficiency. Conversely, the reverse transition is straightforward, as the four wheels will generally remain in contact with the ground, allowing the robot to start moving its legs from a stable condition. The most critical aspect of the transition between legged and wheeled actuation is accurately maintaining proper contact between all four wheels and the ground. Given that Mecanum wheels are composed of multiple rollers, which are convex cylinders arranged around the wheel, the resultant ground-contact interface follows a non-uniform trajectory. This can make it particularly difficult to guarantee stable omnidirectional movement, as any misalignment of the wheels, such as failing to be upright to the surface, can lead to a loss of required movement.

The integration of leg mechanisms substantially improves the ability to overcome these challenges. The legs play a crucial role not only in positioning the wheels correctly but also in acting as a suspension system that helps absorb impact forces \cite{Em-act}. This is advantageous in hybrid locomotion, where maintaining the proper ground-wheel contact is essential. Having a 2 DOF leg is also beneficial in flat terrain scenarios as it constrains the foot to be in the upright position and does not allow it to tilt, which could otherwise alter the wheel’s effective radius of rotation. This structural simplicity reduces the complexity of the control strategy needed to maintain the desired wheel orientation,  which is also a preferred pose of application for this robot.

\begin{figure}[t]
    \centering
    \includegraphics[width=0.5\linewidth]{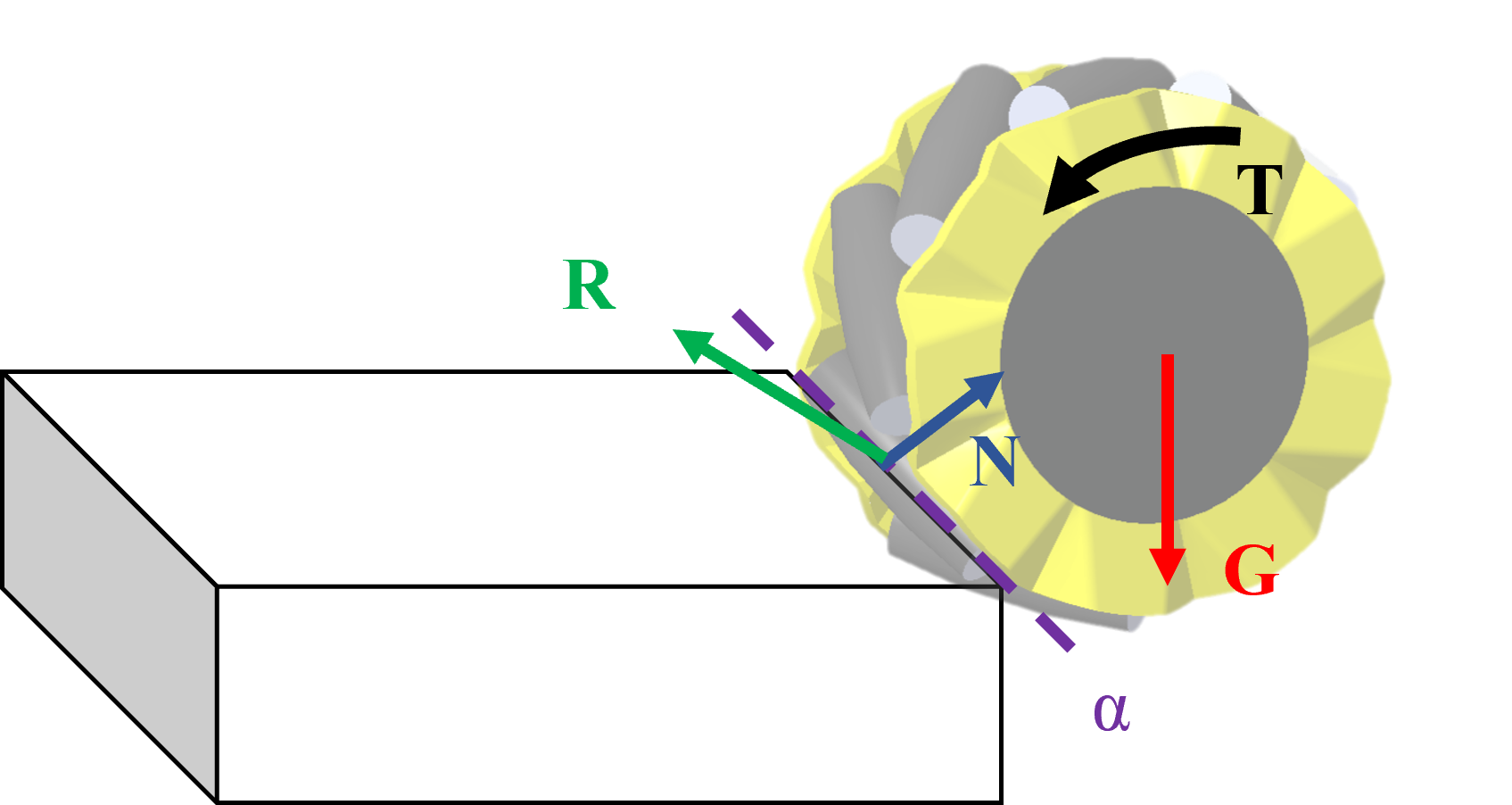}
    \caption{Mecanum wheel on the edge of the obstacle.}
    \label{fig:mec_on_the_edge}
\end{figure}

As illustrated in Figure \ref{fig:mec_on_the_edge}, when a wheel is placed partially on an obstacle, the presence of the free-rotating rollers can cause instability due to a contact force that is not normal to the ground, causing the foot to slip on the edge of the obstacle due to a potential loss of traction as the roller rotates. To prevent this, the implemented control strategy moves the wheel forward beyond its radius before making contact, ensuring that the contact point aligns properly with the wheel's center and the gravitational force direction.

\color{black}
\subsection{Dynamic Model}
The dynamics of a quadruped robot stem from its characterization as a planar rigid body affected by the forces generated by its legs. This system is conceptualized as a floating-base system, indicating that it has no fixed point in space \cite{gor2013}.
\begin{equation}
    \label{robot_vectors}
    q = [q_{b}^{\textit{\myRed{T}}} \quad q_{j}^{\textit{\myRed{T}}}]^{\textit{\myRed{T}}},
    \dot{q} = [v_{b}^{\textit{\myRed{T}}} \quad \dot{ q}_{j}^{\textit{\myRed{T}}}]^{\textit{\myRed{T}}}, 
    \ddot{q} = [a_{b}^{\textit{\myRed{T}}} \quad \ddot{q}_{j}^{\textit{\myRed{T}}}]^{\textit{\myRed{T}}}
\end{equation}
\myRed{to} formulate the quadruped robot’s dynamics, we first describe the robot’s configuration using its generalised coordinates. The pose of the robot’s base, which includes both position and orientation, is denoted by \(q_b \in SE(3)\), and the joint angles of the four legs are denoted by \(q_j \in \mathbb{R}^{8}\). Together, these form the full generalized coordinate vector, \(q = (q_b,q_j)\). The linear and angular velocity and acceleration are represented by \(v_b\) and \(a_b\), respectively. Using this notation, we can write the equations of motion of the robot, which incorporate both the motion of the base and the motion of each joint
\begin{equation}
    \label{dyn_model_quad}
    M(q)\ddot{q} + C(q,\dot{q}) + G(q) = F
\end{equation}
\myRed{where} $M(q) \in \mathbb{R}^{(6+8)\times(6+8)}$ is the inertia matrix, $C(q,\dot{q}) \in \mathbb{R}^{14}$ represents the Coriolis and centrifugal forces, $G(q) \in \mathbb{R}^{14}$ represents the gravitational forces, and $F \in \mathbb{R}^{14}$ is the vector of external forces that include the joints torque and the contact forces of the feet.
When considering an omnidirectional vehicle, it is also possible to write the dynamic equations in the vectorial form as \eqref{dyn_model_quad}. In this case, we consider the robot as a rigid body moving on a plane, with the coordinates of its centre of mass $x$ and $y$ and the yaw orientation\cite{lin2013}, in this case the vector \(q \in \mathbb{R}^{3 + n}\), depending on the space dimension and number of wheels \(n\), that is equal to \(4\) in our case.
The total dynamic model of the OmniQuad robot can be derived from the integration of the dynamic equations of the omnidirectional vehicle into the equations of the quadruped model. In conclusion, the total model can be expressed in the state space form as \eqref{dyn_model_quad}, in which the \(q \in \mathbb{R}^{6 + 8 + 4}\) vector includes also the angular position of the wheels, $M(q) \in \mathbb{R}^{(6+12)\times(6+12)}$, $C(q,\dot{q}) \in \mathbb{R}^{18}$, $G(q) \in \mathbb{R}^{18}$, and $F \in \mathbb{R}^{18}$.

\subsection{Control System}
This section describes the control system and the full robot's control diagram. It begins with a direct derivation of the inverse kinematics. Following this, there is an explanation of the two modules used for obstacle detection and avoidance. Lastly, it provides a detailed understanding of the control diagram.
\subsubsection{Inverse Kinematics module}
Similar to the overall model of the full robot, the inverse kinematics can be structured into two distinct subsystems, one for the legs and another for the wheels.

In detail, starting from the kinematic model \eqref{for_kin_rr} of the leg, the inverse kinematics equations are derived as
\begin{align}
q_{2} &= \pm \arccos\,\!\Bigl(
     \frac{x^{2} + z^{2} - l_{1}^{2} - l_{2}^{2}}{2\,l_{1}\,l_{2}}
   \Bigr), \label{eq:q1}\\[6pt]
q_{1} &= \arctan2\,\bigl(z,\,x\bigr)
         \;-\;
         \arctan2\,\!\Bigl(
              l_{2}\,\sin(q_{2}),\;l_{1} + l_{2}\,\cos(q_{2})
        \Bigr).\label{eq:q2}
\end{align}
\
To compute $q_{2}$, two possible solutions exist for a given foot position. The correct choice depends on the desired shape of the leg in motion. The two most common possibilities are the C-shape gait, in which the knee points backward, and the X-shape gait, in which the legs cross over. By selecting the appropriate sign of $q_{2}$, it is possible to ensure the leg follows the intended trajectory.

On the other hand, starting from equation \eqref{wheel_kin}, it is possible to derive the inverse kinematics of the omnidirectional vehicle, which takes as input the linear and angular velocity of the base:

\begin{equation}
\begin{pmatrix}
\omega_{1}\\[2pt]
\omega_{2}\\[2pt]
\omega_{3}\\[2pt]
\omega_{4}\\[2pt]
\end{pmatrix}
\;=\;
{
  \frac{1}{\,R\,}
  \begin{pmatrix}
    1 & -\cot(\gamma) & \tfrac{1}{k(\gamma)} \\[6pt]
    1 & \cot(\gamma)  & \tfrac{1}{k(\gamma)} \\[6pt]
    1 & -\cot(\gamma) & -\tfrac{1}{k(\gamma)} \\[6pt]
    1 & \cot(\gamma)  & -\tfrac{1}{k(\gamma)}
  \end{pmatrix}
}
\begin{pmatrix}
V_x\\[3pt]
V_y\\[3pt]
\omega_{z}
\end{pmatrix},
\label{eq:omni_inverse_kinematics} 
\end{equation}

where $k(\gamma)$ depends on the x-position of the foot, while the y-coordinate remains fixed. Specifically $k(\gamma)$ is given by
\begin{equation}
    k(\gamma,q_{1},q_{2}) = \frac{1}{b(q_{1},q_{2}) - h \tan(\gamma)} ,
\end{equation}
where $b(q_{1},q_{2})$ represents the distance along the $X$ axis from the centre of mass to the wheel contact point. Considering the nominal case $b_{0}$ in which the wheels are positioned directly below the shoulder, $b(q_{1},q_{2})$ is expressed as
\begin{equation}
     b(q_{1},q_{2}) = b_{0} + l_{1} \cos(q_{1}) + l_{2} \cos(q_{1}+ q_{2}) .
\end{equation}

\subsubsection{Obstacle Detection and Obstacle Climbing modules}
\label{obstacle_detect}

The Obstacle Detection Module is responsible for identifying obstacles and estimating their height using torque measurements from the wheel's current sensor. Meanwhile, the Obstacle Climbing Module ensures the system maintains its centre of mass within the support polygon and plans the required motion to either climb over or avoid obstacles. This module relies on IMU readings to adjust the base's orientation appropriately.

When a wheel encounters an obstacle and gets stuck, the torque increases significantly due to resistance, activating the obstacle detection module. To overcome the obstruction, the Obstacle Climbing Module initiates a wheel-lifting manoeuvre while keeping the wheel rotating to maintain elevated torque. This sustained torque helps the system accurately detect when the wheel is still in contact with the obstacle. As the wheel clears the obstacle, the torque drops to near zero, signalling that lifting is no longer needed. Once the robot is free from the obstacle, it can plan a trajectory to avoid it and continue with the task.

When two wheels get stuck, the robot must employ a legged locomotion strategy to climb the obstacle. The Obstacle Climbing Module first decides which leg to lift and then computes the new centre of gravity based on the remaining supported legs. The robot begins to unbalance its base by shifting its position and, if necessary, applying pitch and/or roll compensation in height, as illustrated in Section \ref{legged}. Once the two front wheels reach the top of the obstacle, the robot stabilizes itself using IMU readings and continues rotating the wheels until it has traveled a distance equivalent to the length of its body. When the front and hind wheels are at different heights, the robot must adjust its pitch angle to move forward to the centre of gravity. Before lifting the last leg, it performs a roll adjustment to ensure proper stepping onto the obstacle.

\vspace{-2mm}
\begin{figure}[!t]
\centering
\includegraphics[width=6.4in]{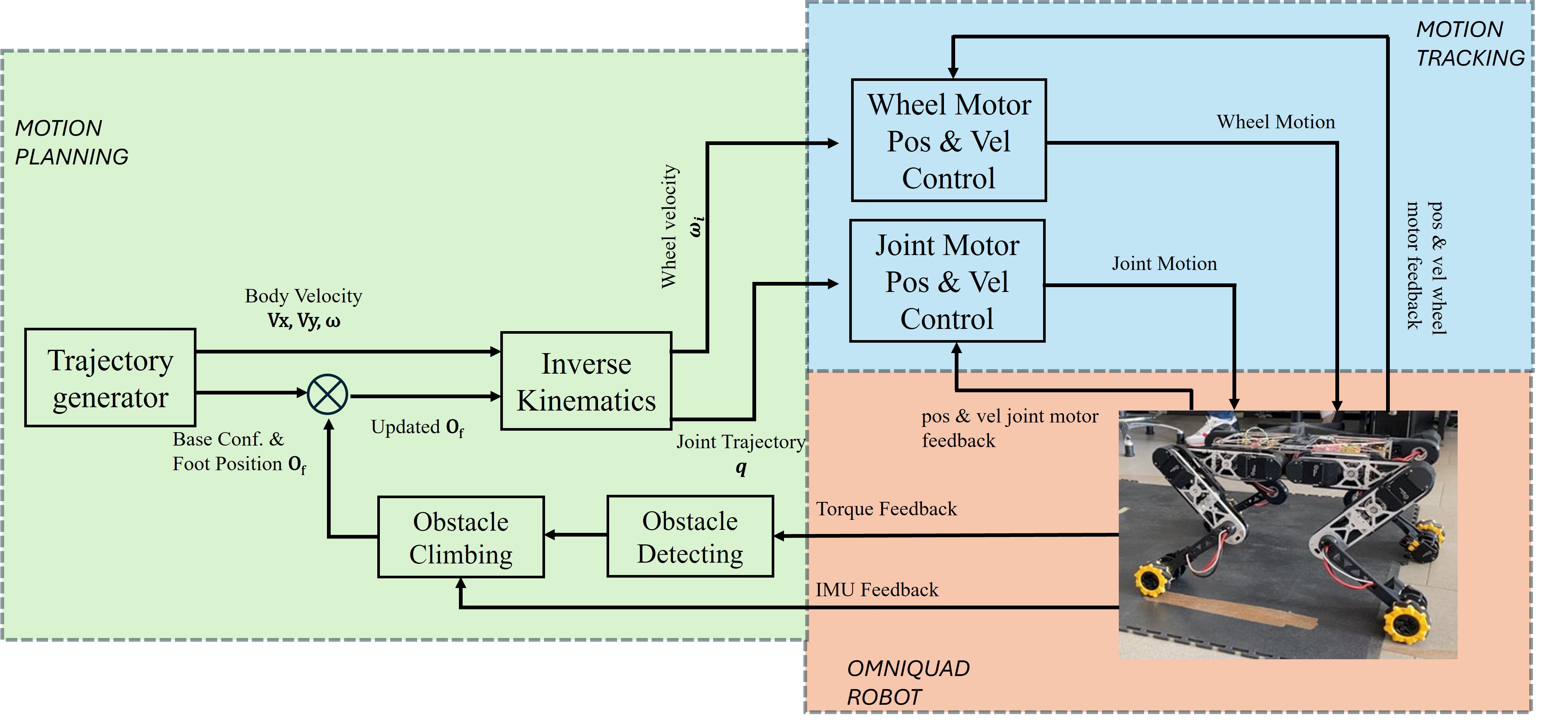}
\caption{Control Scheme of OmniQuad.}
\label{con_scheme}
\vspace{-2mm}
\end{figure}

\color{black}
\subsubsection{Control Diagram}
The control scheme of the entire system is shown in Fig. \ref{con_scheme}. The primary motion objective entails maintaining the centre of mass of the robot within the support polygon while successfully executing a predetermined trajectory.
The motion planning module generates the trajectory of the leg joints and the motion of the robot's body. At first, computing the linear and angular velocity of the robot depends on the trajectory it needs to follow, which will be used to determine the trajectory of the wheels. In this part, also the base configuration and the feet position are estimated, which will be used to generate the joint trajectory based on position and velocity references.
We have also integrated a planning module that receives commands from a gamepad. This setup allows users to set the desired linear velocities \(V_{x}\) and \(V_{y}\), angular velocity \(\omega\), and the height adjustment rate of the floating base. \myRed{Before computing the inverse kinematics, the obstacle detection and obstacle climbing modules updated the wheels' positions based on the readings from the current sensors present in the encoders of the wheels and IMU data to overcome possible obstacles that the robot may encounter.
After computing the joint references using the inverse kinematics equations \eqref{eq:q1}, \eqref{eq:q2}, \eqref{eq:omni_inverse_kinematics}, the motion tracking module ensures precise execution of the desired joint trajectories. To achieve this, a dedicated ROS2 hardware interface is developed to enable joint actuation through FD-CAN, utilizing a Raspberry Pi 4 in combination with a Pi3Hat for seamless motor control. This interface allows users to set motor parameters—including position, velocity, effort, current limits, and motor gains—through a configurable file, ensuring adaptability to various operational requirements.}

\section{Experimental Validation}
A series of experiments was conducted to assess the functional capabilities of the OmniQuad robot comprehensively. \myRed{ Initially, we focused on tuning the PD controllers for the robot’s joints—a process that proved extremely helpful for setting appropriate gains in all subsequent trials. Next, we carried out tests to evaluate the performance of the wheel-foot design, specifically looking at slippage, friction, and the actuation system’s behaviour on different terrains.}

Following this, the robot's abilities in two distinct obstacle-overcoming scenarios were examined. The first experiment focused on passing under an overhead obstacle while maintaining the same orientation, demonstrating the robot's omnidirectional movements. The second experiment aimed at assessing the robot's legged locomotion capabilities, achieved by climbing over an obstacle. Moreover, the joystick controller was also tested in an outdoor environment. 

\myRed{Subsequently, stress tests were performed. Primarily, we observed the obstacle detection module in action, with the robot successfully operating on only three wheels. Furthermore, a durability test is conducted, designed to evaluate the robustness of the overall hardware. Finally, we assessed the maximum payload the robot can carry using its legs}. To view the experimental tests, please also refer to the attached video.

\subsection{Compatibility hardware software test}
\begin{figure*}[t]
\begin{center}

    \subfloat[\scriptsize Results of the tuning of the wheel.]{
        \includegraphics[width=0.32\textwidth,trim={3.6cm 0.4cm 4.4cm 1.7cm}, clip]{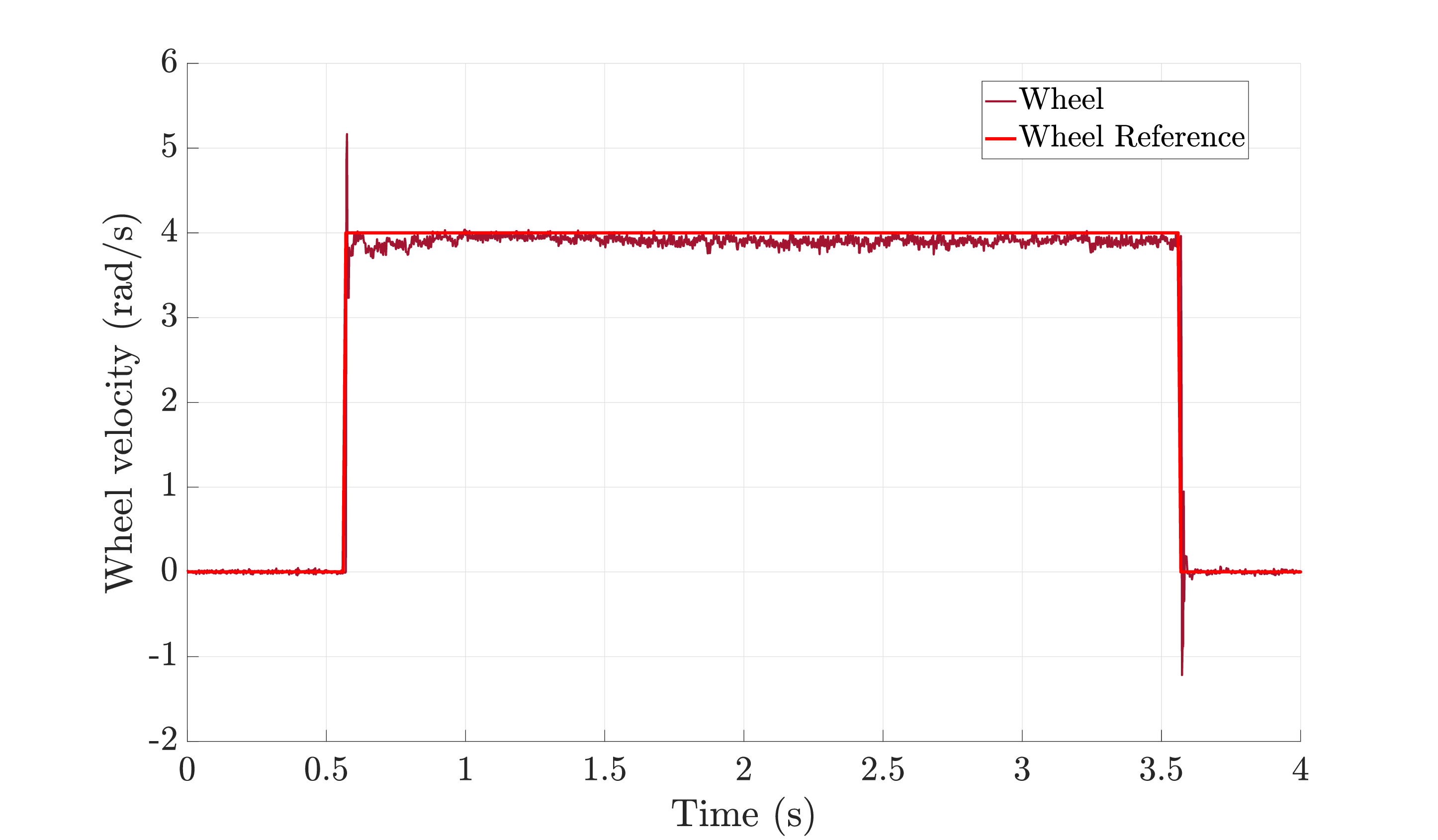}
        \label{fig:tune_wheel}}\hfill
    \subfloat[\scriptsize Results of the tuning of the hip joint.]{
        \includegraphics[width=0.32\textwidth,trim={3.1cm 0.4cm 4.45cm 1.7cm},clip]{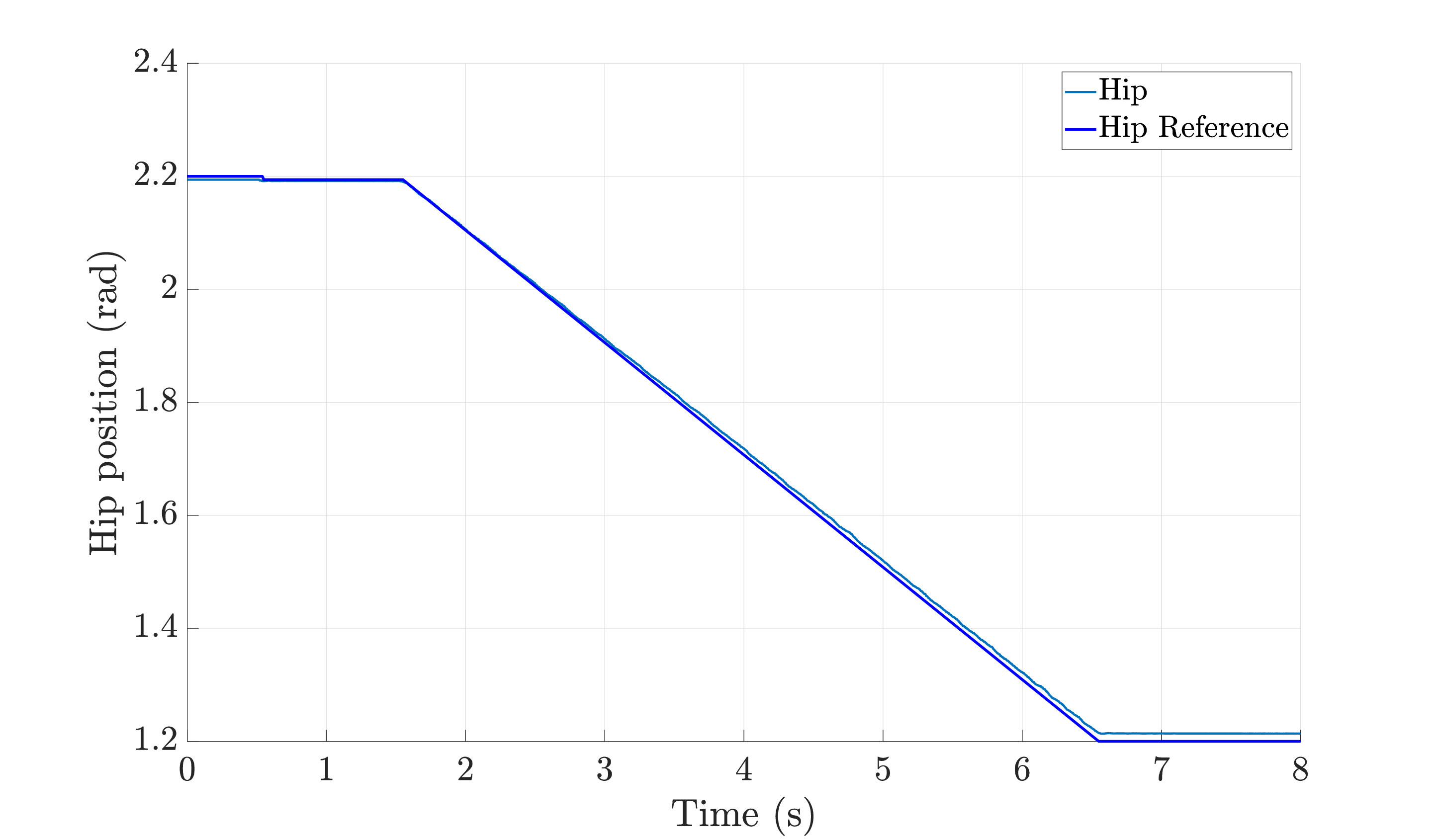}
         \label{fig:tune_hip}}\hfill
    \subfloat[\scriptsize Results of the tuning of the knee joint.]{
        \includegraphics[width=0.32\textwidth, trim={2.6cm 0.4cm 4.4cm 1.7cm},clip]{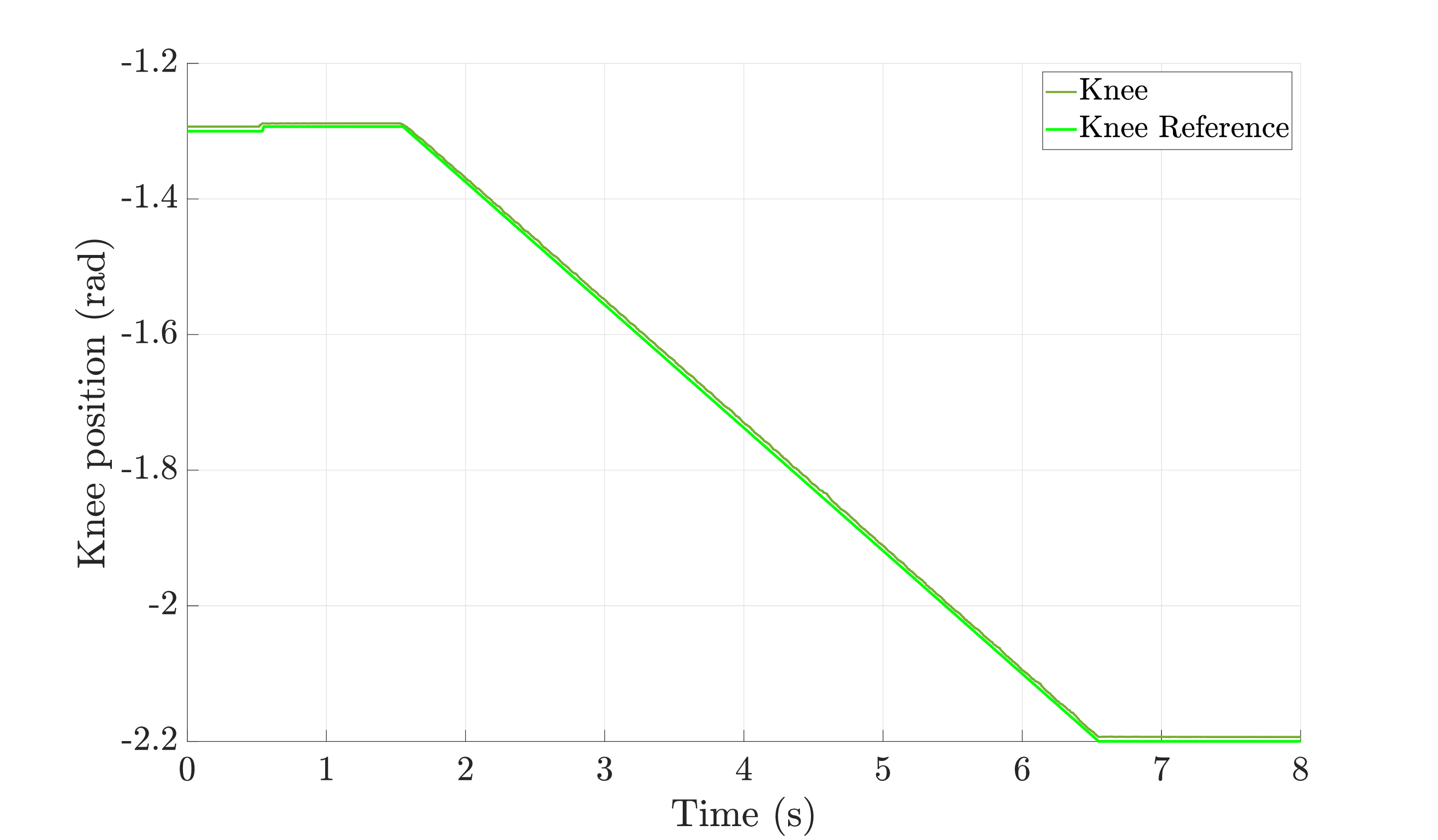}
         \label{fig:tune_knee}}
\caption{PD tuning test.}
\label{tuning}

\end{center}
\end{figure*}

During the compatibility testing phase of the OmniQuad robot, the PD controllers were carefully tuned. Proper PD tuning is essential as it directly affects the robot’s stability and responsiveness, ensuring smooth motion and precise control over its actuators. A proportional gain of 150 Nm/rad and a derivative gain of 0.5 Nms/rad were established for the leg joints. These values provide a balanced trade-off—keeping joint error relatively low while allowing the actuators to remain sufficiently compliant, Figs. \ref{fig:tune_hip}, \ref{fig:tune_knee}, rather than overly rigid. This elasticity helps the robot adapt better to dynamic tasks. Conversely, since the wheel joints rely primarily on velocity control, only the derivative term was set at 0.95 Nms/rad, resulting in consistent and controlled wheel movement throughout all experiments, Fig. \ref{fig:tune_wheel}.

\subsection{Actuation system performance test}
\begin{figure}[t]

    \centering
    \begin{minipage}[b]{0.3\textwidth}
        \centering
        \includegraphics[width=\textwidth, trim={3.6cm 0.4cm 4.4cm 45cm}, clip]{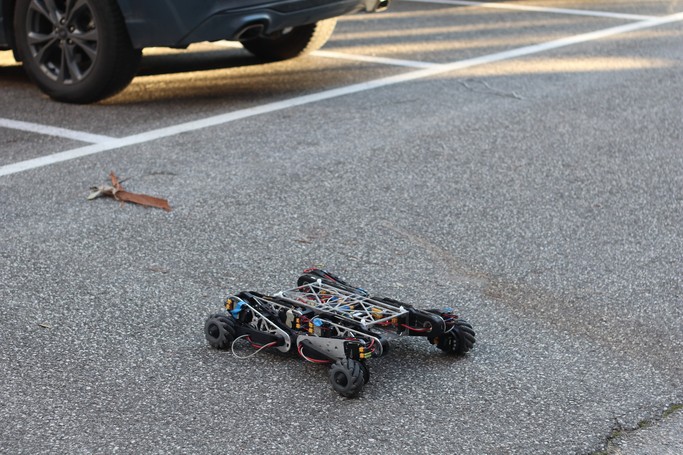}
        \captionof{figure}{\myRed{Experiment on asphalt street terrain.}}
        \label{fig:set_asphalt}
    \end{minipage}\hfill
    \begin{minipage}[b]{0.3\textwidth}
        \centering
        \includegraphics[width=\textwidth, trim={3.6cm 20cm 4.4cm 25cm}, clip]{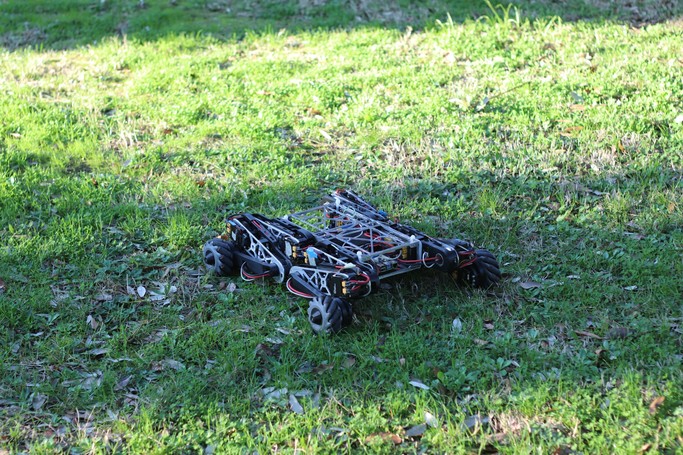}
        \captionof{figure}{\myRed{Experiment on grass terrain.}}
        \label{fig:set_grass}
    \end{minipage}\hfill
    \begin{minipage}[b]{0.3\textwidth}
        \centering
        \includegraphics[width=\textwidth, trim={2.0cm 0.0cm 1.0cm 0.0cm}, clip]{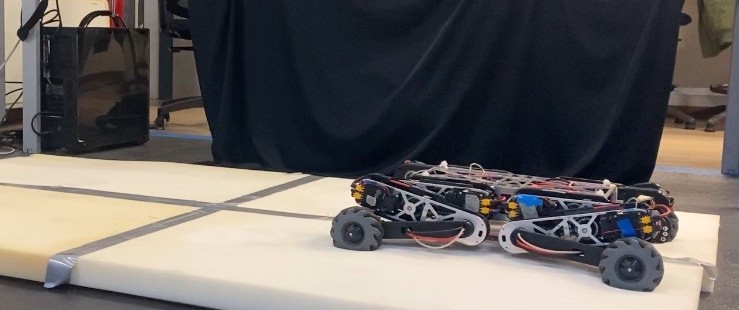}
        \captionof{figure}{\myRed{Experiment on soft terrain.}}
        \label{fig:set_soft}
    \end{minipage}

\end{figure}
\subsubsection{Experimental setup}
To test the performance of the wheel-foot actuation system, we conducted a series of experiments in both indoor and outdoor environments. 
Indoors, we focused on evaluating forward and lateral movements to determine the range of velocities in which the robot could operate. Gymnastic mats were used to minimize surface irregularities and increase traction between the wheels and the floor.

\begin{figure}[!t]
\centering
\includegraphics[width=0.5\linewidth, trim={0.5cm 5cm 0.5cm 2cm}, clip]{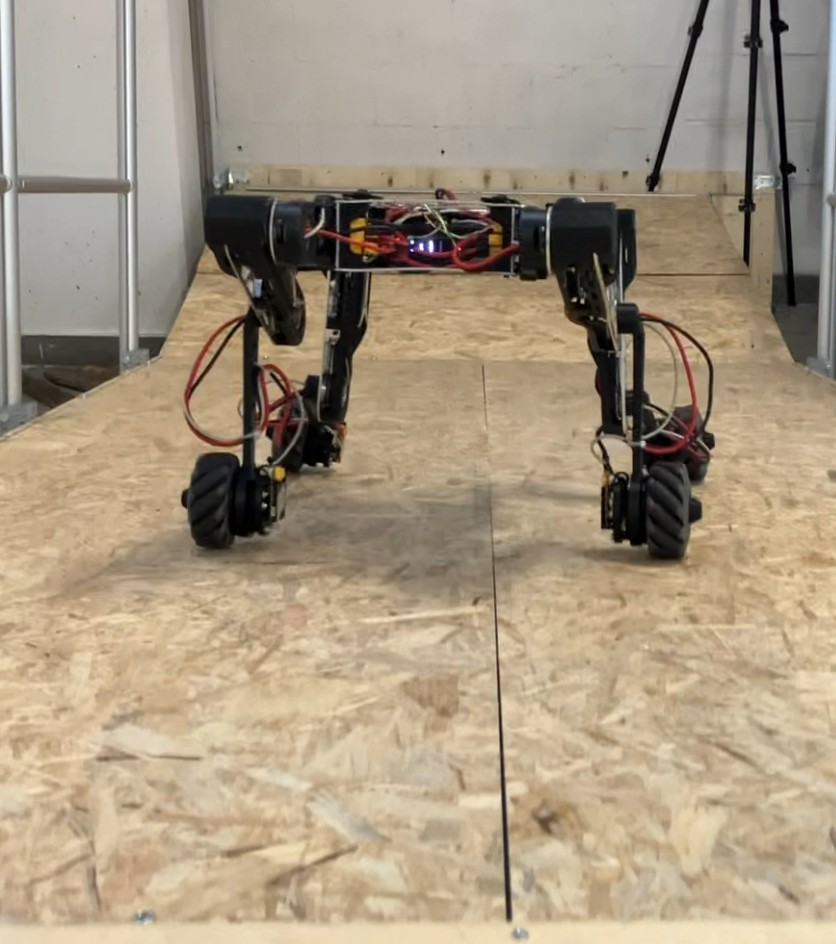}
\caption{\myRed{Robot on a 20$^\circ$ slope.}}
\label{fig:set_slope}
\vspace{-2mm}
\end{figure}

Afterwards, we tested the entire actuation system—including the motor, gearbox, and wheels—across various outdoor terrains, specifically asphalt (Fig. \ref{fig:set_asphalt}) grass (Fig. \ref{fig:set_grass}) and soft polyurethane foam (Fig. \ref{fig:set_soft}). Finally, we evaluated the performance of the system on a sloped surface to demonstrate its capability when operating on ramps, which is presented in Fig. \ref{fig:set_slope}.

\subsubsection{Test description}
To test the performance of the wheels, we designed a velocity trajectory with an initial acceleration phase, followed by a constant phase, a deceleration phase to zero, and then a negative velocity phase mirroring the first half of the trajectory in reverse. We applied this velocity profile in both the forward and lateral directions. During outdoor testing, the robot reached a forward velocity of 1 m/s and then reversed to -1 m/s. We chose this speed because the robot could exhibit the same behaviour in the forward direction, although we were unable to replicate it in lateral motion. Finally, in the slope experiment, we commanded a velocity reference of 0.5 m/s in the forward direction and then reversed at the same speed over a shorter time, to prevent the robot from falling off the ramp.
\subsubsection{Results and discussion}
During forward movements, the robot reached a top speed of 2 m/s without any slippage. In lateral movements, it achieved a maximum speed of 1.5 m/s, also without slippage. However, higher friction forces and vibrations were observed during lateral movements, limiting performance compared to forward motion. 

The total torque measured in these experiments, as shown in Fig. \ref{torque_exp}, indicates that the highest torque occurs during the acceleration phase of forward motion. In lateral motion, the torque remained consistently high, even at constant velocity, due to the vibrations and friction forces characteristic of omnidirectional wheels. Furthermore, the total torque magnitude aligns closely with the estimate provided in Section\ref{Design} when considering a forward velocity of 2 m/s.

The corresponding results are illustrated in Figs. \ref{fig:street}, \ref{fig:grass}, and \ref{fig:soft}. On asphalt street terrain, the maximum torque appears at the start of the task, as expected, and then gradually decreases. On grass, the maximum torque also peaks at the beginning but fluctuates irregularly, likely due to the uneven ground. In contrast, soft terrain produces the highest maximum torque of all tests, remaining elevated even during movement. Despite these variations, the velocity remained stable overall, with no significant issues during either forward or backward movements.

We encountered limitations in wheel motion, particularly during lateral movements. For example, in the grass experiment, the primary challenge was terrain irregularity, as the wheels performed lateral motions more effectively in some cases than in others. On soft terrain, the maximum lateral velocity achieved was 0.1 m/s. The results of this test are summarized in Table \ref{wheels_capa}, and additional details can be observed in the attached video.

\noindent
\begin{minipage}[t]{0.48\textwidth}
  \centering
  \includegraphics[width=\linewidth, trim={3cm 0cm 4cm 1cm}, clip]{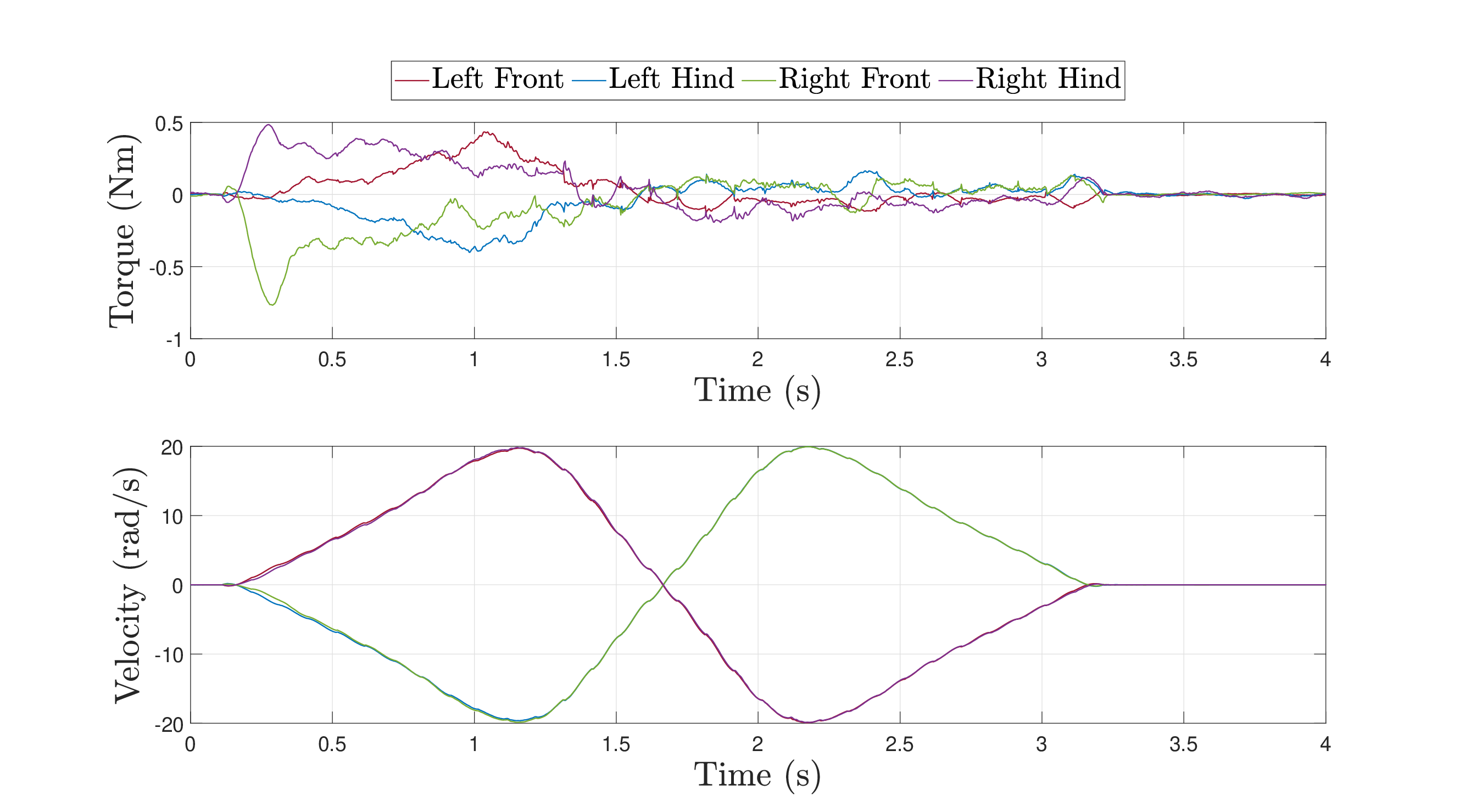}
  \captionof{figure}{\myRed{Torque and velocity on asphalt street.}}
  \label{fig:street}
\end{minipage}\hfill
\begin{minipage}[t]{0.48\textwidth}
  \centering
  \includegraphics[width=\linewidth, trim={3cm 0cm 4cm 1cm}, clip]{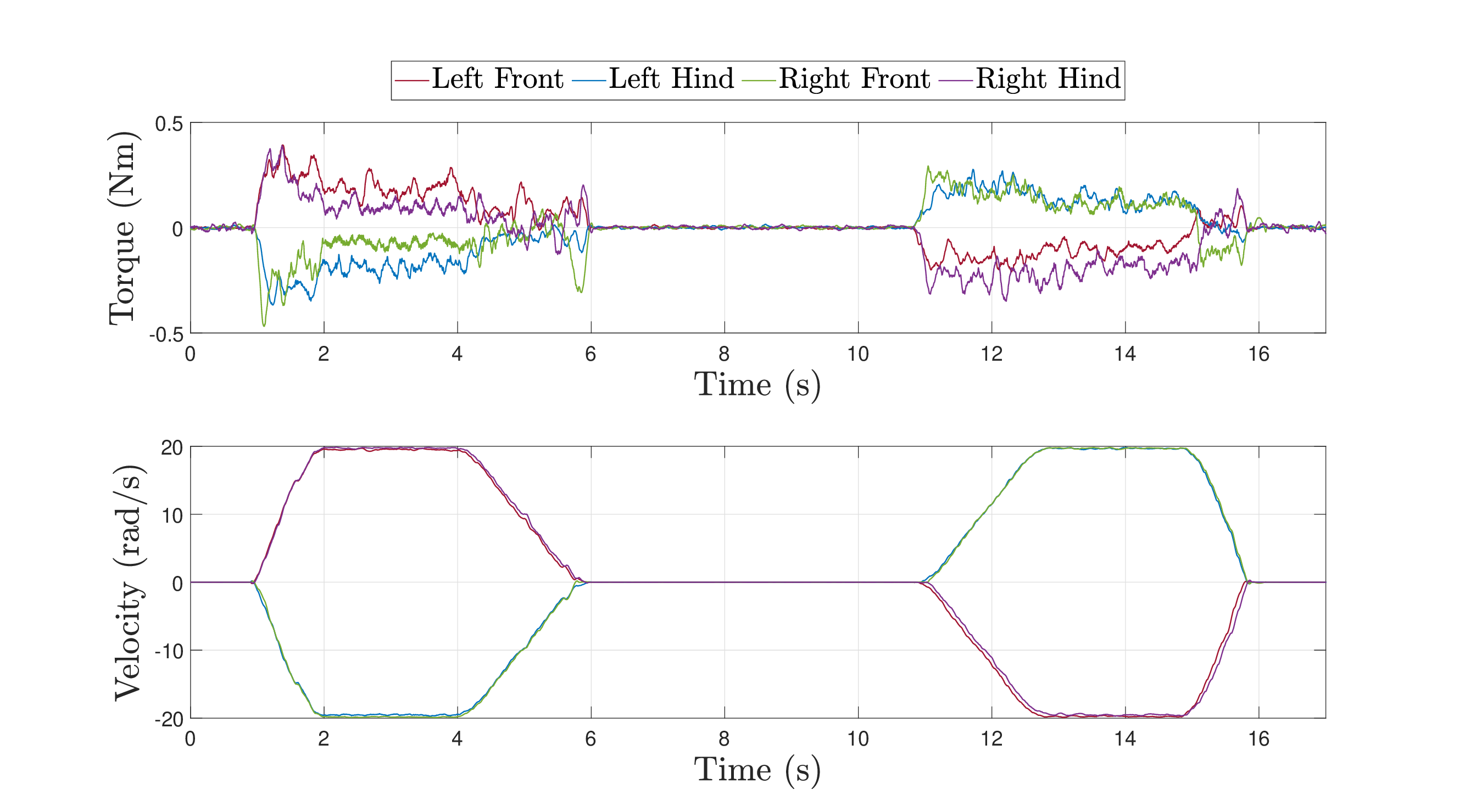}
  \captionof{figure}{\myRed{Torque and velocity on grass.}}
  \label{fig:grass}
\end{minipage}

\vspace{1em}

\noindent
\begin{minipage}[t]{0.48\textwidth}
  \centering
  \includegraphics[width=\linewidth, trim={3cm 0cm 4cm 1cm}, clip]{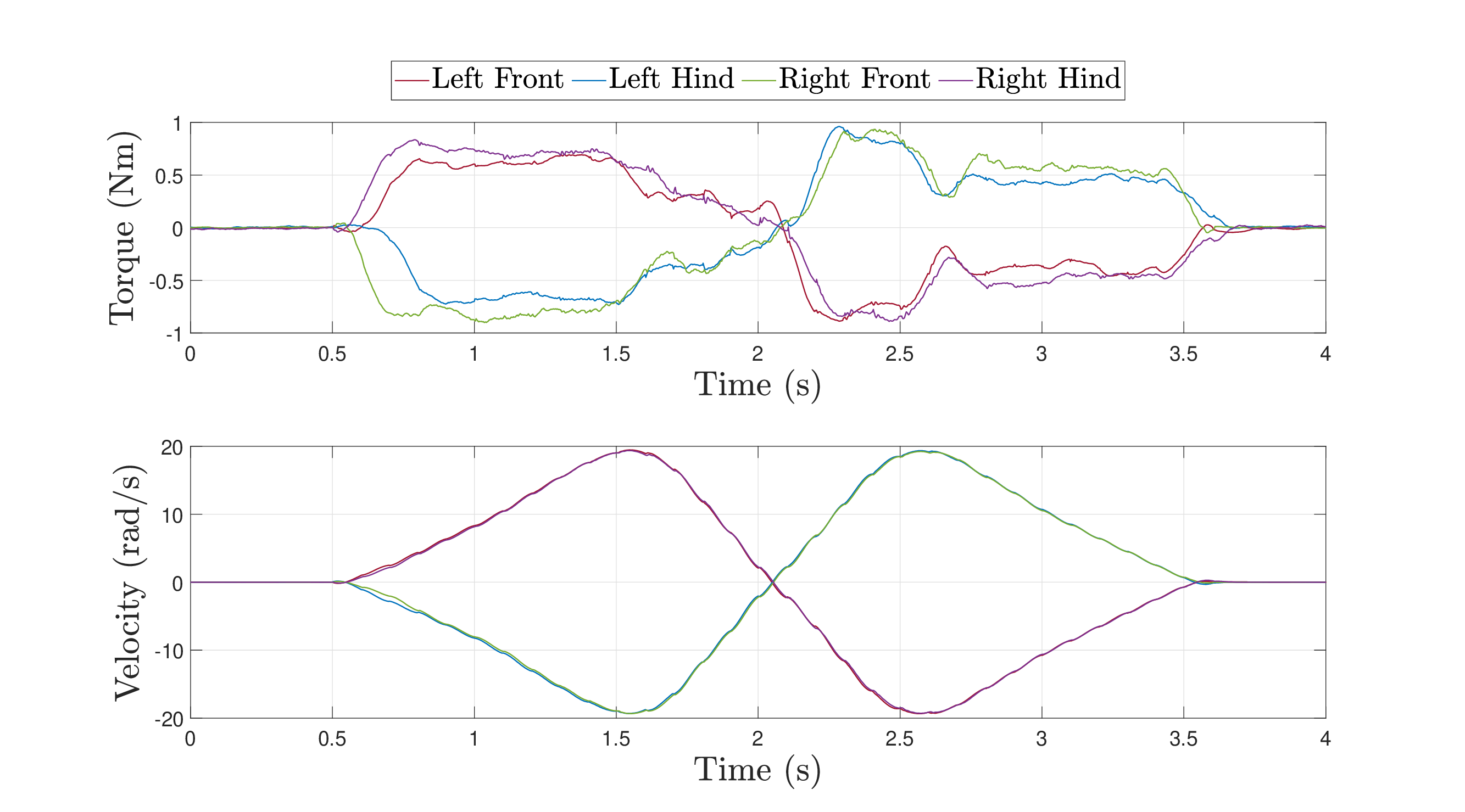}
  \captionof{figure}{\myRed{Torque and velocity on soft terrain.}}
  \label{fig:soft}
\end{minipage}\hfill
\begin{minipage}[t]{0.48\textwidth}
  \centering
  \includegraphics[width=\linewidth, trim={3cm 0cm 4cm 1cm}, clip]{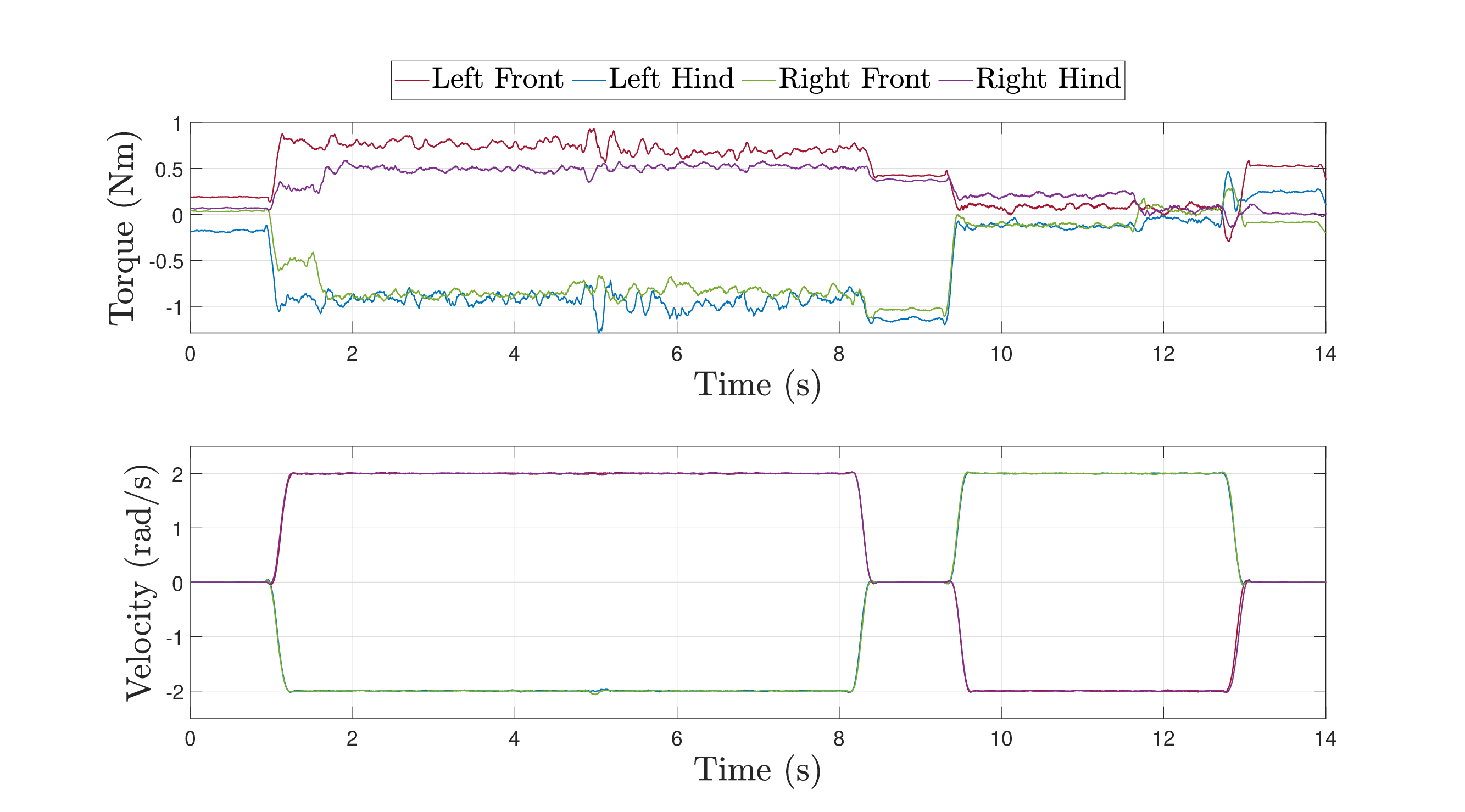}
  \captionof{figure}{\myRed{Torque and velocity climbing the 20$^\circ$ slope.}}
  \label{slope}
\end{minipage}

\vspace{1em}

\begin{center}
  \begin{minipage}[t]{0.48\textwidth}
    \centering
    \includegraphics[width=\linewidth]{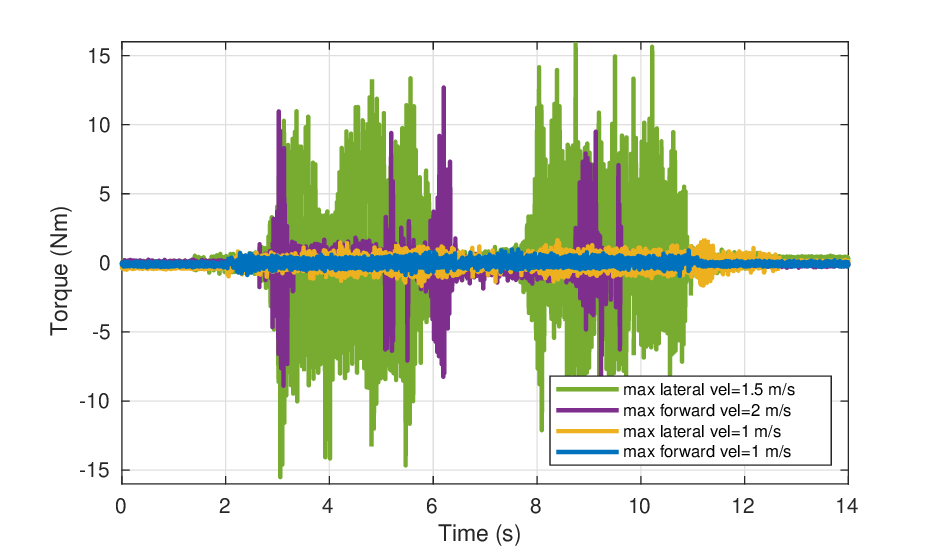}
    \captionof{figure}{\myRed{Torque registered during the experiment on testing forward and lateral velocity.}}
    \label{torque_exp}
  \end{minipage}
\end{center}

\begin{table}[h]
\renewcommand{\arraystretch}{1.2} 
\setlength{\tabcolsep}{8pt} 
\small

\centering
\caption{Wheel performance on different terrains}
\label{wheels_capa}
\begin{tabular}{c c c c}

\hline
\textbf{Type of Terrain} & \textbf{Max Forward Velocity} & \textbf{Max Lateral Velocity} & \textbf{Max Torque} \\
\hline
Asphalt Road & 2 & 1.5  & 0.8 \\[0.2em]

Grass  & 2 & 1    & 0.5 \\[0.2em]

Soft (Polyurethane Foam)  & 2 & 0.1  & 1.3 \\
\hline

\end{tabular}
\end{table}

We also tested performance on slopes up to 20$^\circ$, as shown in Fig. \ref{slope}. We successfully climbed the slope at a forward velocity of 0.1m/s, which generated sufficient torque for a stable ascent. Afterwards, the robot descended at -0.1m/s. For a more detailed overview, please refer to the attached video.


\color{black}
\vspace{-3mm}
\subsection{Overcome obstacle: going below}
\subsubsection{Experimental setup}
To test the capabilities of the robot, we imitated a realistic scenario in which it could operate. In this scenario, we used a rod longer than the robot itself, which was supported at both ends by two sets of boxes positioned at a predetermined height of 200 mm.

\subsubsection{Test description}
The primary objective is to avoid the obstacle and consequently overcome it while maintaining the same orientation of the robot. This experimental assessment is divided into three key phases. In the initial phase, the robot is set in forward motion while concurrently lowering its base to a specific height of 180 mm. Subsequently, in the second phase, the robot executes lateral movement, successfully passing beneath the horizontal rod. Lastly, the robot proceeds to raise its base back to its initial pose in the third phase of the evaluation.

\subsubsection{Results and discussion}
\begin{figure*}[!t]
\centering
\subfloat[\scriptsize Sequence image of the task completed by Omniquad]{\includegraphics[width= \textwidth,  trim = 0in 0in 0in 0in] {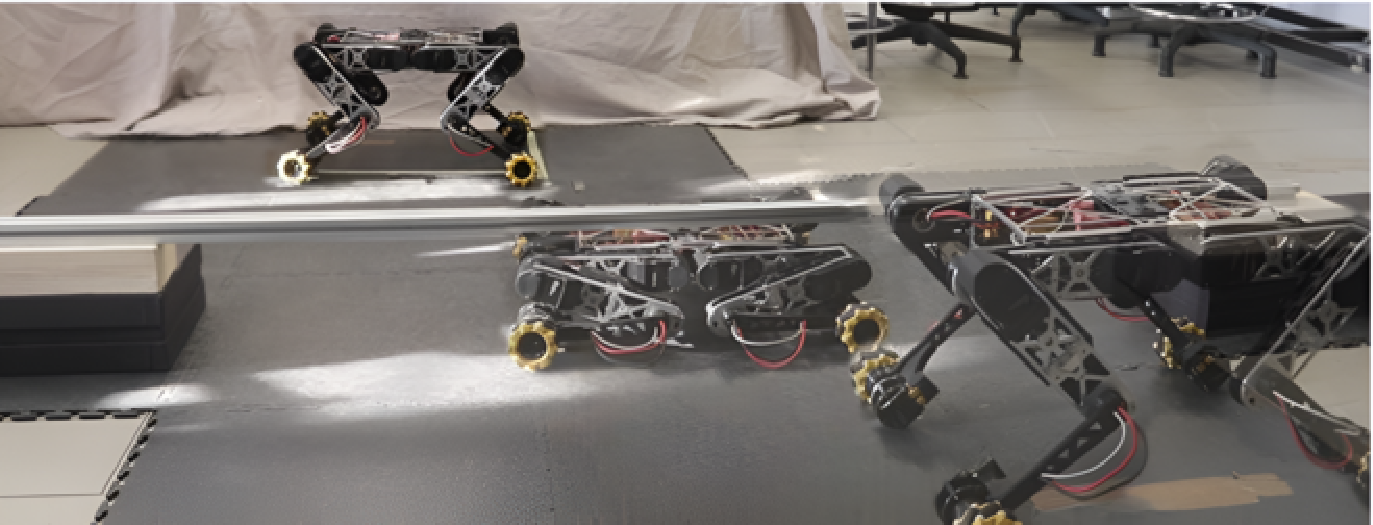}}
\label{seq_task1}\\
\subfloat[\scriptsize Wheels velocity trajectory]{\includegraphics[width= 0.45\textwidth,  trim = {0in 0.0in 0in 0.0in}] {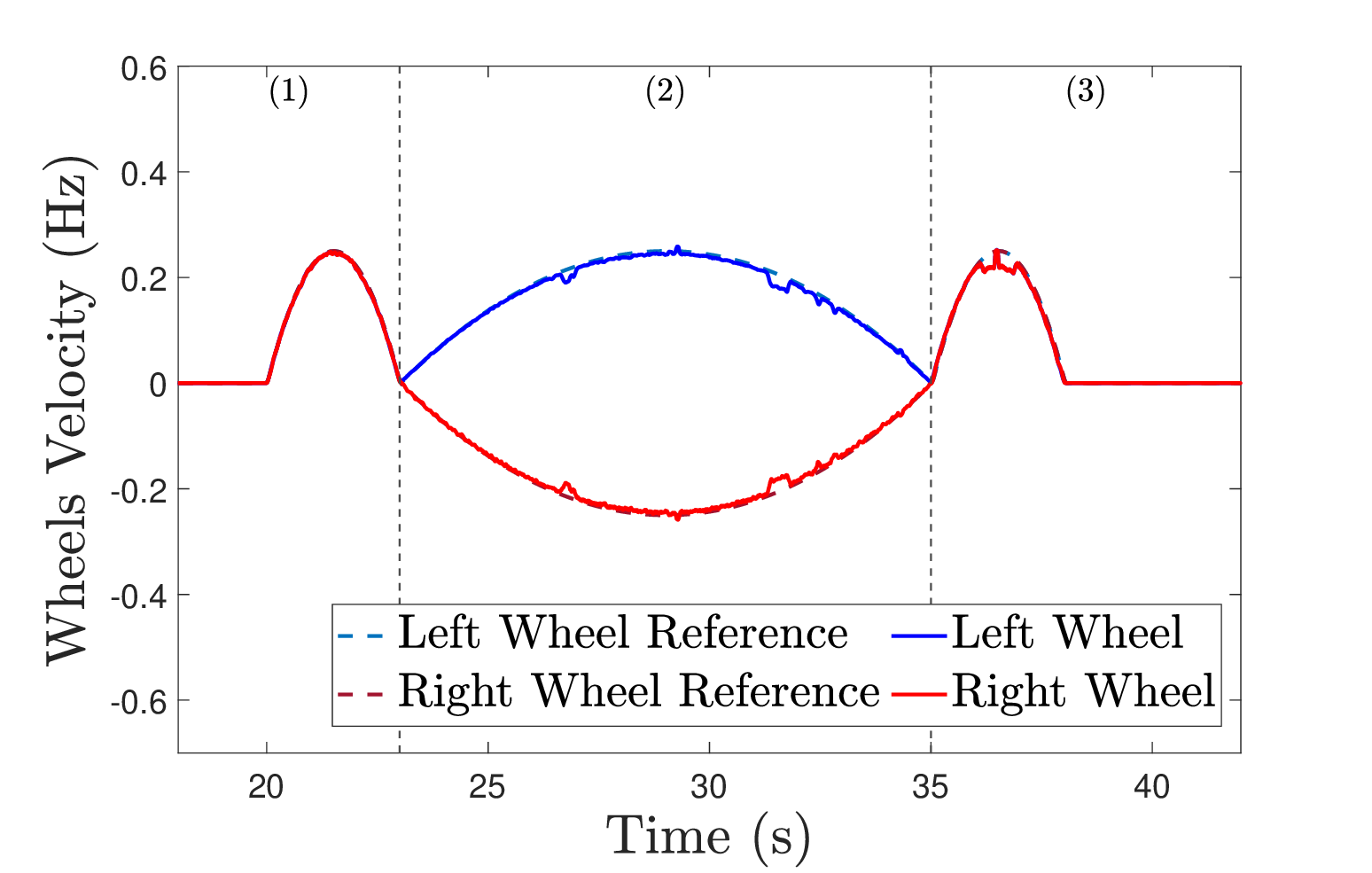}
\label{wheel_task1}}
\subfloat[\scriptsize Height of the robot during the task]{\includegraphics[width= 0.45\linewidth,  trim = 0.0in 0.0in 0.0in 0.0in] {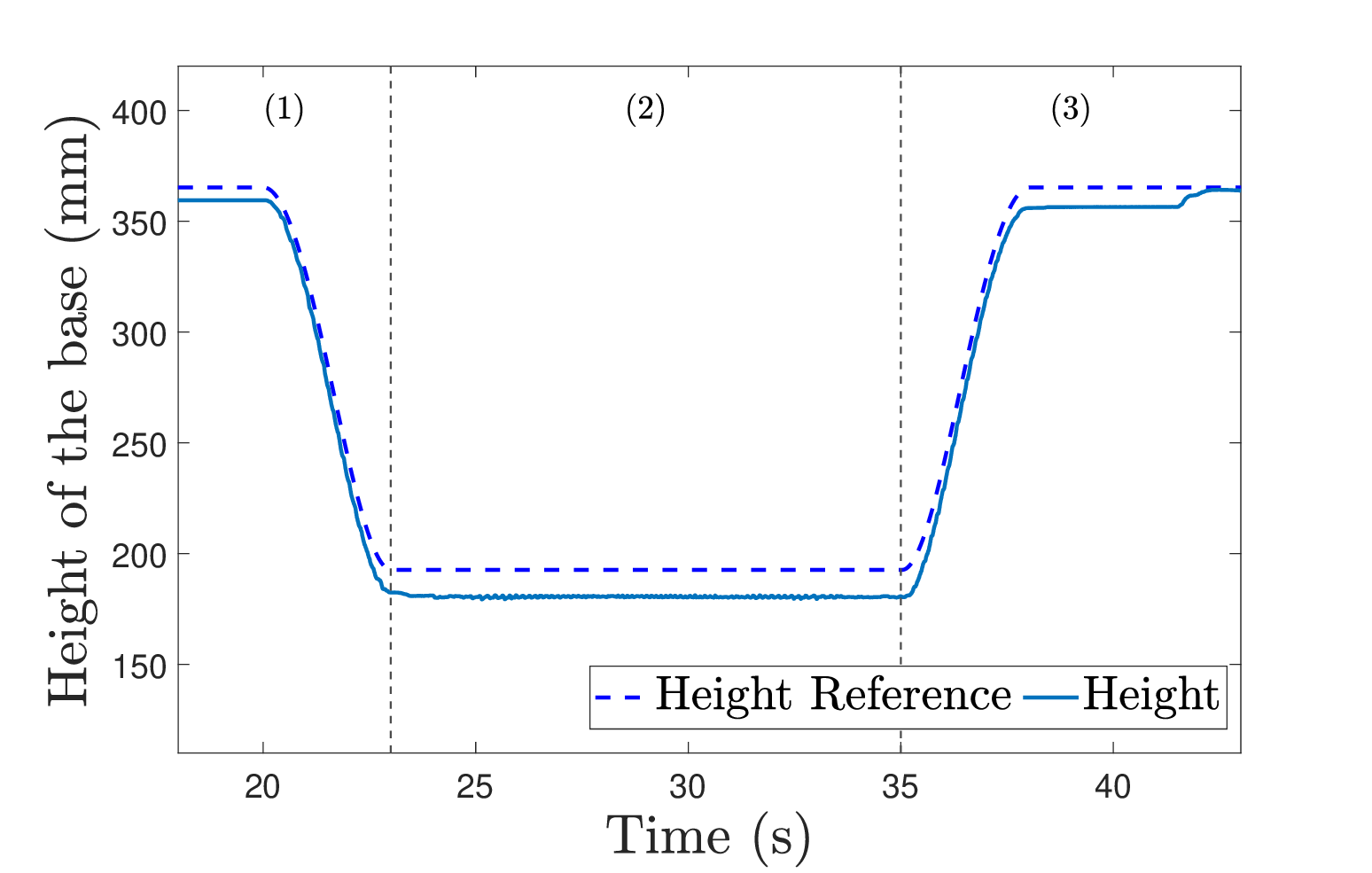}
\label{height_Task1}}\\
\caption{\small Results of the task "going below the obstacle":  (1) the robot is in the default configuration,  (2) it lowers the base and moves laterally under the obstacle, (3) it lifts its base and returns in homing configuration.}
\label{task1}
\vspace{-1mm}
\end{figure*}

The findings of this experiment are visually presented in Fig. \ref{task1}. In Fig. \ref{height_Task1}, the height of the robot and its reference are presented during the whole task. In the phase where the robot is shorter, the error is 20 mm due to the weight of the robot contributing to a decrease in the total height, but without compromising the success of the experiment the observed variance between the reference and actual joint trajectories is sufficiently minimal to ensure the successful execution of the task at hand; in detail, the maximum recorded error is around $3^{\circ}$. Right and left wheels, Fig. \ref{wheel_task1}, display rotational motion in opposite directions when the robot is moving laterally. Notably, during lateral movement, the system operates smoothly without any discernible vibrations due to the discontinuity of the rollers. Nevertheless, as the velocity is increased, some vibrations become perceptible; however, it is crucial to underscore that the overall system remains operationally unaffected by these vibrations, which are quite common when operating with mecanum wheels.

\vspace{-3mm}
\subsection{Overcome obstacle: going over}

 \begin{figure*}[!t]
\centering
\subfloat[\scriptsize Sequence image of the task completed by OmniQuad]{\includegraphics[width= \textwidth,  trim = 0in 0in 0in 0in] {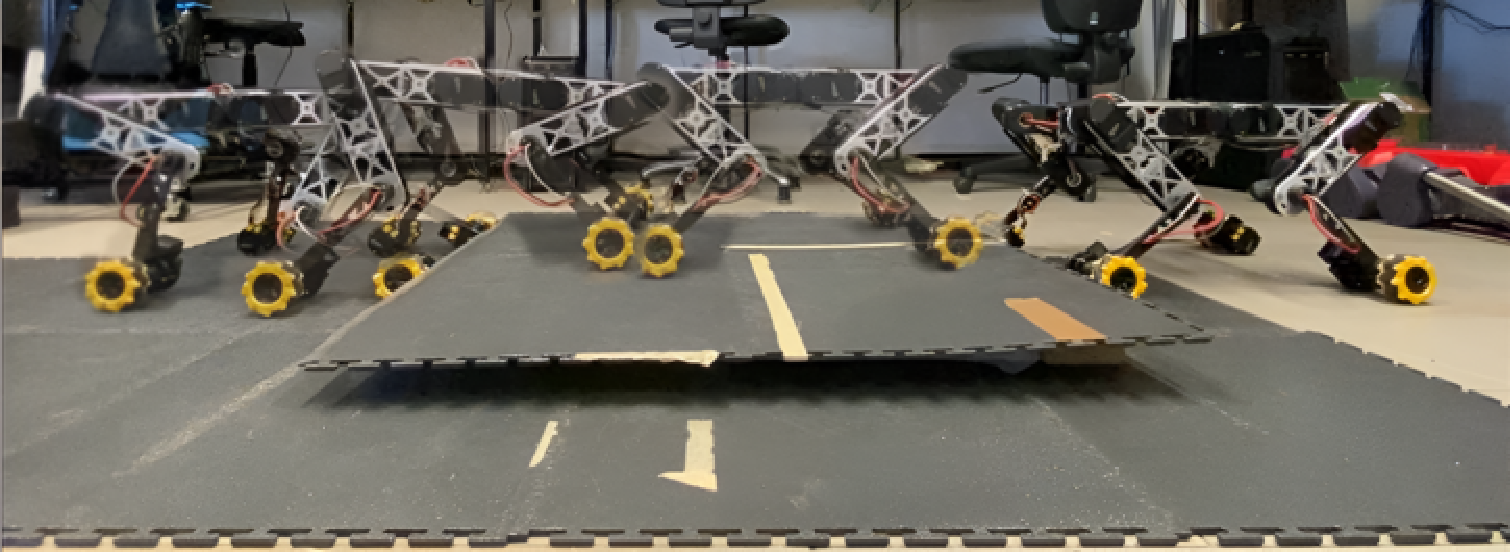}}
\label{seq_task2}\\

\subfloat[\scriptsize Front Legs trajectory]{\includegraphics[width=0.9\textwidth] {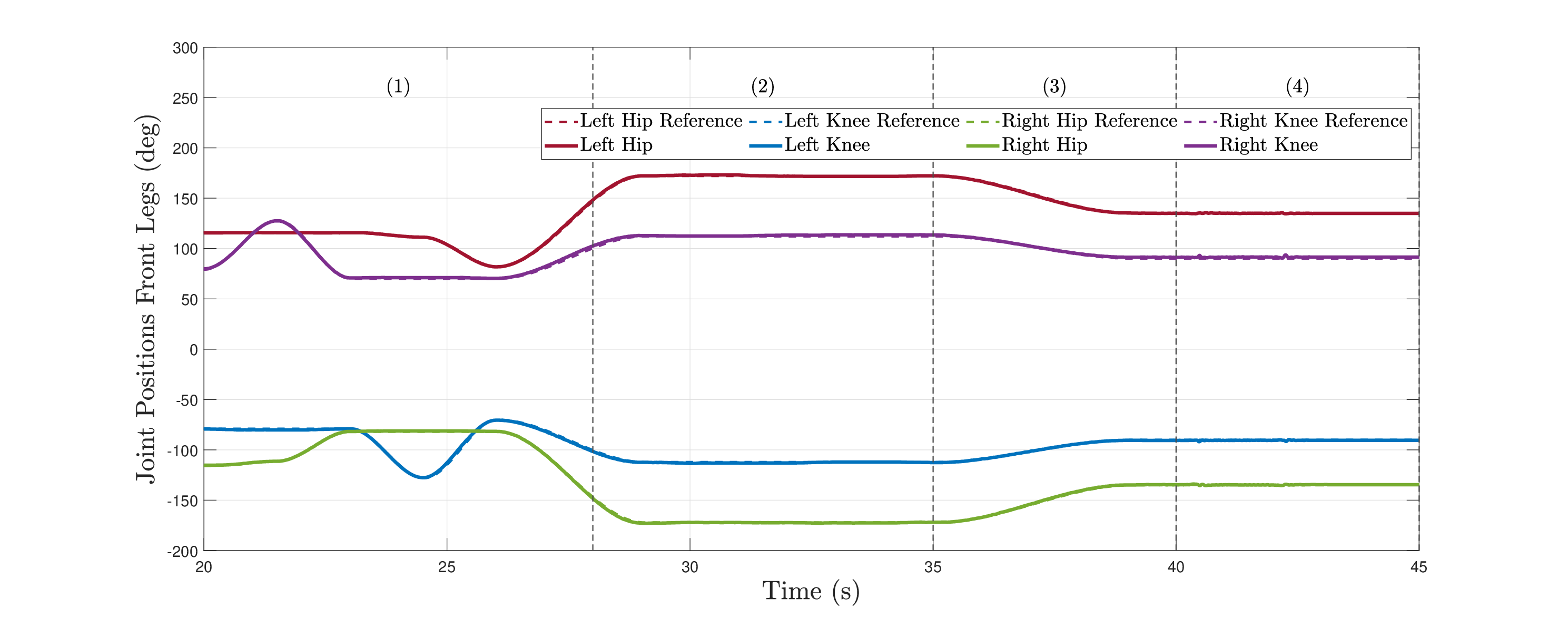}
\label{Fjoint_task2}}\\
\vspace{-2mm}
\subfloat[\scriptsize Hind Legs trajectory]{\includegraphics[width=0.9\linewidth, trim = 0in 0in 0in 0in] {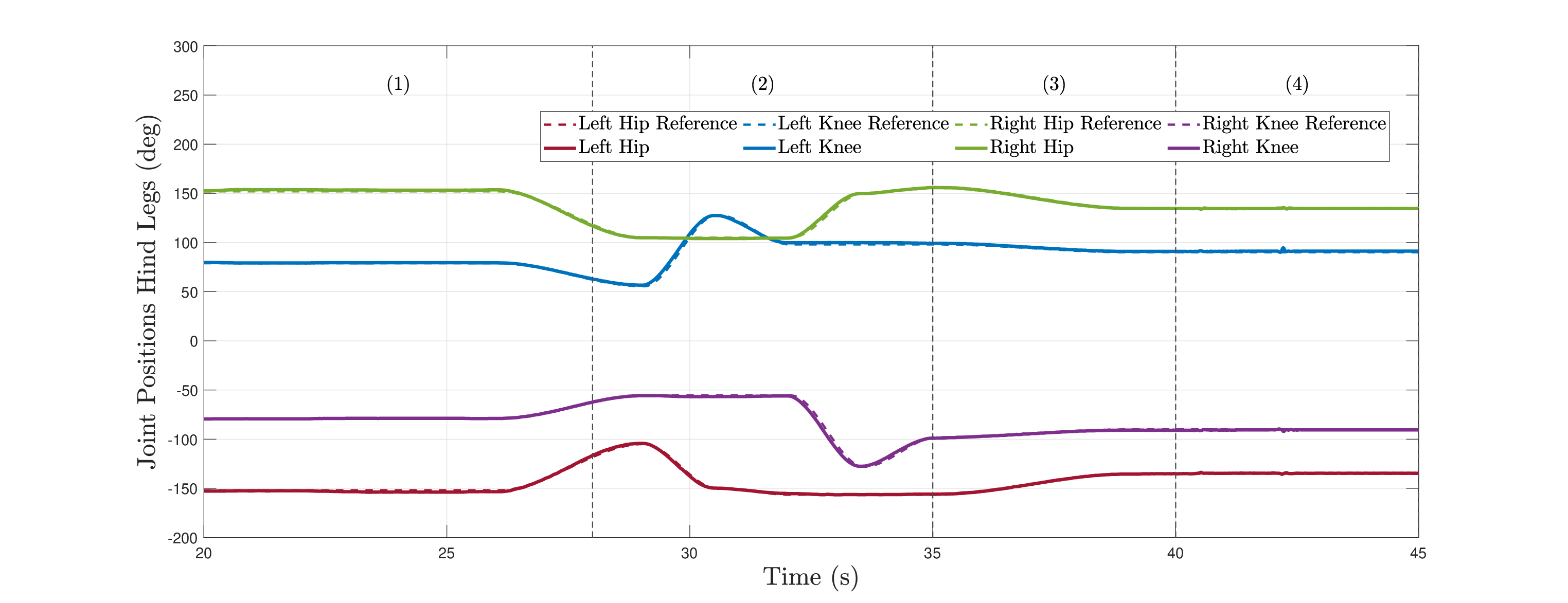}
\label{Hjoint_Task2}}
\caption{\small Results of the task "going over the obstacle": (1) the robot lifts the front legs and places them over the obstacle, (2) it tilts the base and climbs the obstacle with the hind legs, (3) it returns in default configuration and start moving forward with the wheels, (4) it descends the obstacle using the wheels.}
\label{task2}
\vspace{-1mm}
\end{figure*}

\subsubsection{Experimental setup}
 To complete the task of climbing on the obstacle, rigid boxes were employed to emulate a step structure of specified height. These boxes were cushioned by the application of gymnastic mats, not only on top of the boxes but also in the immediate vicinity, with the intent of minimizing terrain irregularities, ensuring a uniform surface, and even keeping the robot safe from any sudden failure.

\subsubsection{Test description}
 The experiment is divided into three distinct phases. Initially, the robot is positioned in front of a step obstacle, and it methodically moves one foot at a time, initiating the process of ascending a 100 mm high obstacle with its front legs. In the subsequent phase, the robot employs its wheels to manoeuvre and approach the hind legs with the step, allowing it to effectively surmount the obstacle. Finally, the robot readjusts the orientation of its base and overcomes the obstacle, this last part is executed by the robot using the wheels.

\subsubsection{Results and discussion}
 Fig. \ref{task2} illustrates the outcomes of this experiment, with a specific focus on front legs, Fig. \ref{Fjoint_task2}, and hind legs, Fig. \ref{Hjoint_Task2}. To ensure the robot's stability, the Center of mass orientation needs to be changed to ensure that it will be contained in the support polygon, also in a configuration with just three legs in contact. This can be achieved by tilting the robot so that, while lifting any foot, it  will not make the robot fall, regarding the fact that when the robot has just three feet in contact with terrain, the centre of gravity changes. Notably, the joint trajectories corresponding to the front legs and hind legs exhibit a noteworthy degree of similarity, due to the symmetric configuration of the legs. The error between the reference and the measurement is less than $1.5^{\circ}$ for each leg throughout the task.
 
\vspace{-3mm}
 \subsection{Joystick Experiments}
  \begin{figure*}[!t]
\centering
\subfloat[\scriptsize Sequence image of the task completed by OmniQuad]{\includegraphics[width=\textwidth,  trim = 0in 0in 0in 0in] {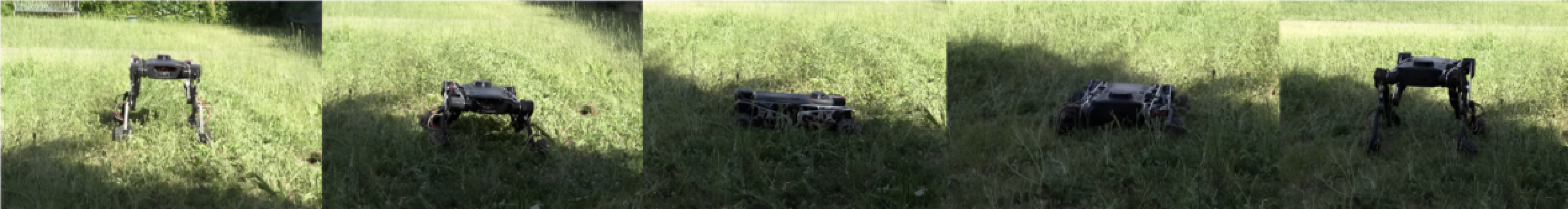}}
\label{out_task}\\
\subfloat[\scriptsize Height of the base trajectory]{\includegraphics[width= 0.45\linewidth,  trim = 0in 0in 0in 0in] {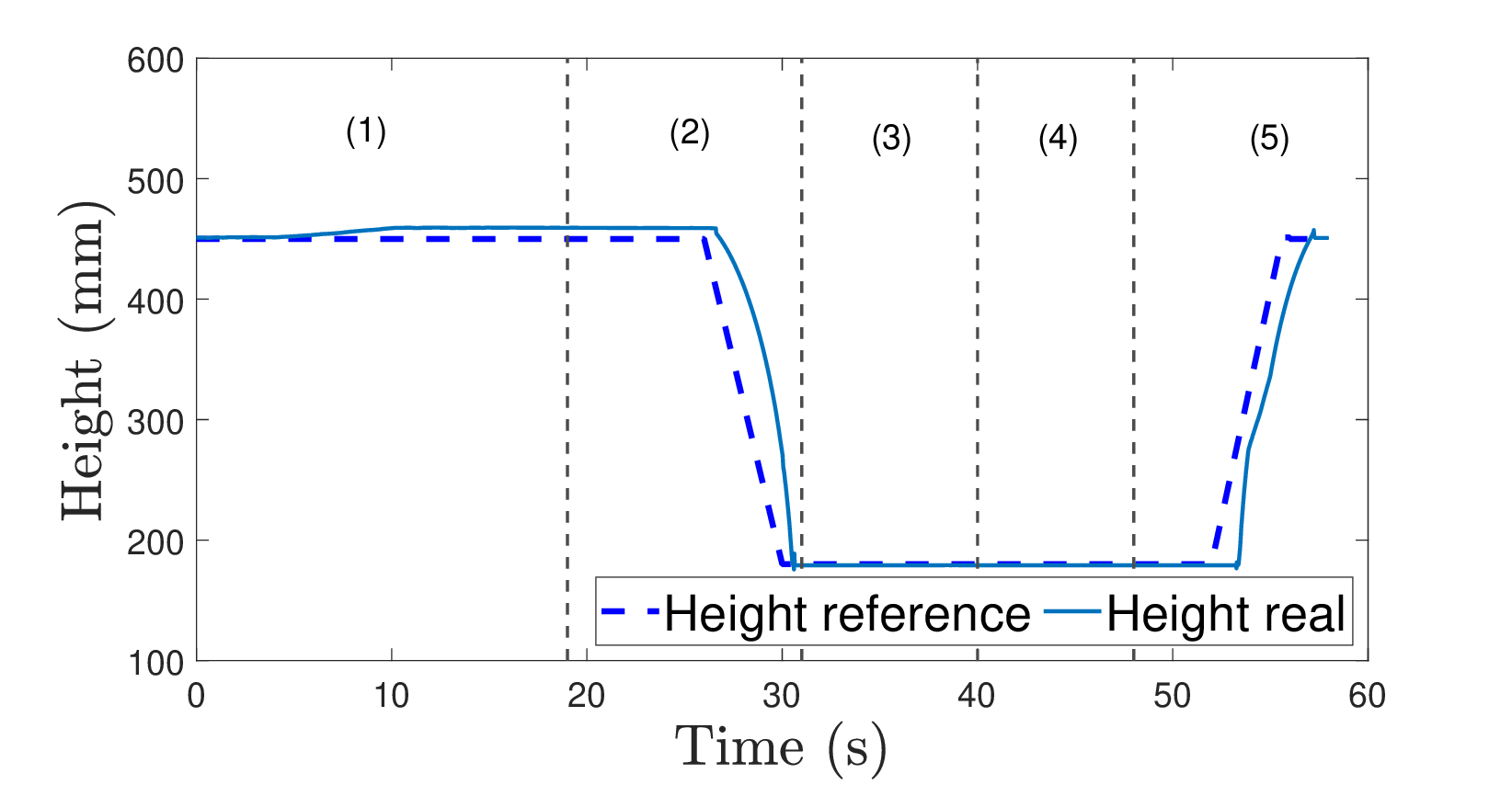}
\label{out_height}}
\subfloat[\scriptsize Linear and angular velocity trajectory]{\includegraphics[width= 0.45\linewidth,  trim = 0in 0in 0.15in 0in] {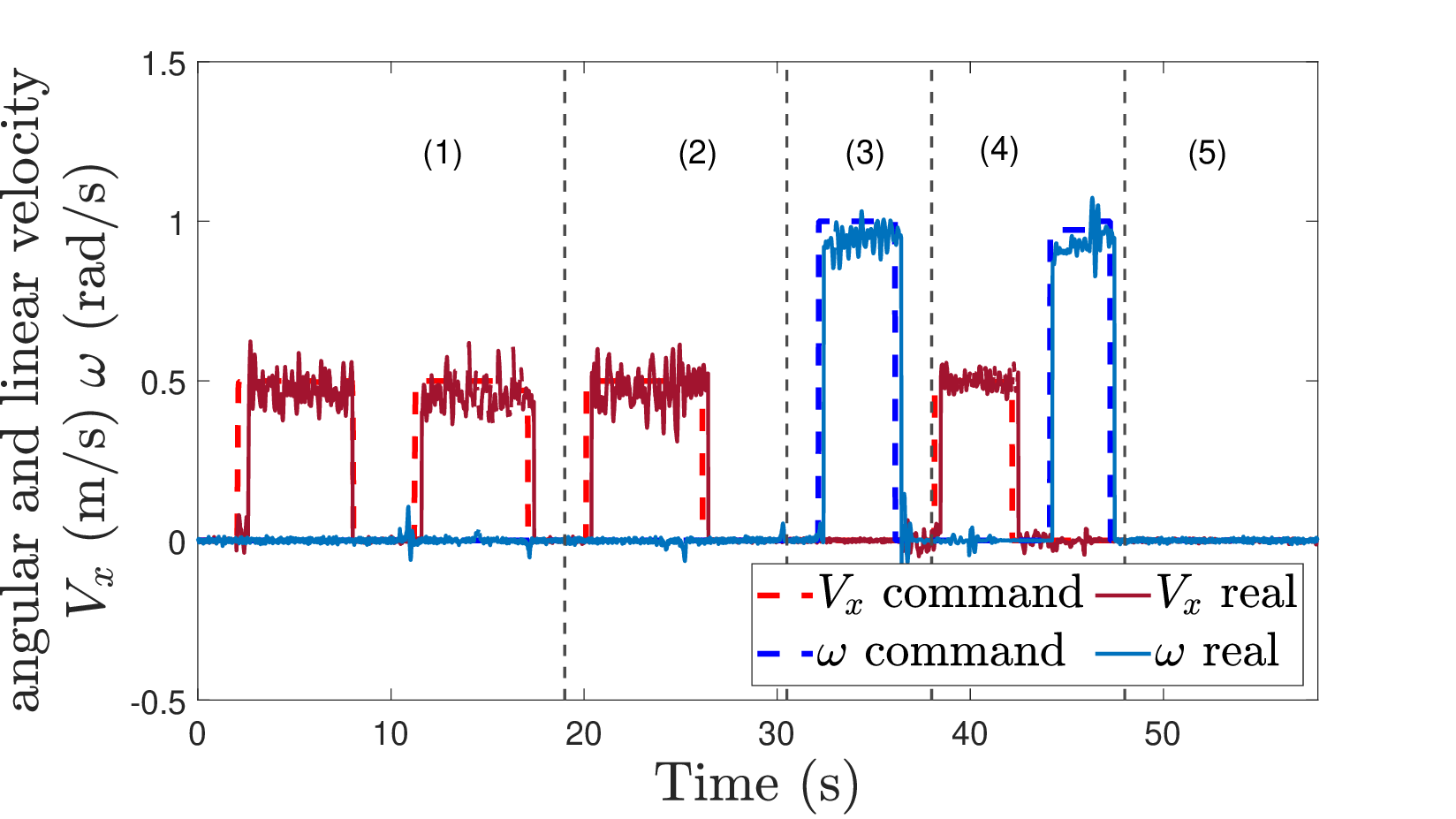}
\label{out_vel}}
\caption{\small Results in natural environment using the joystick: (1) the robot moves forward, (2) the robot starts lowering its base, (3) it turns right, (4) it moves forward and turns right again, (5) it lifts the base.}
\label{outdoor}
\vspace{-2mm}
\end{figure*}
 \subsubsection{Experimental setup}
To thoroughly assess the capabilities of the OmniQuad robot, we conducted experiments to assess the correct functionality of the joystick command controller. These trials occurred on grass, a natural surface frequently encountered in human environments, to showcase robustness when encountering uneven terrains. 
 
\subsubsection{Test description}
The objective was to create a 3 m $\times$3 m grid where the robot could operate at both maximum and minimum heights, performing omnidirectional movements such as moving forward and rotating in place, as illustrated in Fig. \ref{outdoor}. The commands were provided via a gamepad to demonstrate its functionality. The robot began at its maximum height of approximately 450 mm, moved forward at a speed of 0.5 m/s, and stopped after covering $1$ meter. It then resumed forward movement before lowering its base to the minimum height and making a $90^{\circ}$ turn. After turning, it moved forward again, executed another $90^{\circ}$ turn, and finally raised its base back to the maximum height. The maximum angular velocity reached was 1.0 rad/s.

\subsubsection{Results and discussion}
Fig. \ref{out_vel} displays the reference trajectory for both angular and linear velocities, while Figure \ref{out_height} illustrates the height reference throughout the experiment. For the sake of brevity, only a partial reference to the experiment is provided due to the repetition of the task. These results demonstrate that, despite the irregular terrain and increased vibrations, the robot completed the task, indicating its potential for effective performance in outdoor environments.

\subsection{Obstacle detection experiment}

\begin{figure*}[t]
    \centering
    
    \begin{minipage}{0.24\textwidth}
        \centering
        \begin{overpic}[width=0.95\linewidth, trim={5cm 10cm 25cm 0cm}, clip]{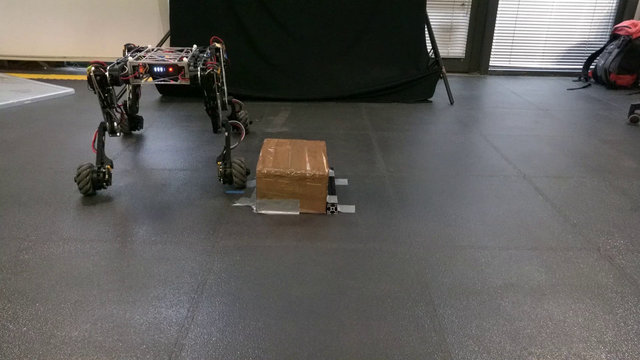}
            \put(2,4){\color{white}\bfseries (1)}
        \end{overpic}
        \label{fig:omni_seq1}
    \end{minipage}\hfill
    \begin{minipage}{0.24\textwidth}
        \centering
        \begin{overpic}[width=0.95\linewidth, trim={5cm 10cm 25cm 0cm}, clip]{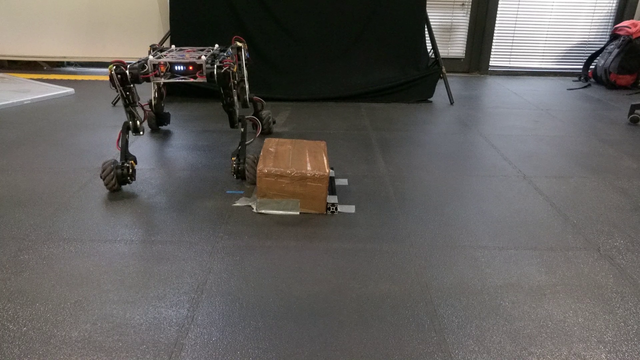}
            \put(2,4){\color{white}\bfseries (2)}
        \end{overpic}
        \label{fig:omni_seq2}
    \end{minipage}\hfill
    \begin{minipage}{0.24\textwidth}
        \centering
        \begin{overpic}[width=0.95\linewidth, trim={5cm 10cm 25cm 0cm}, clip]{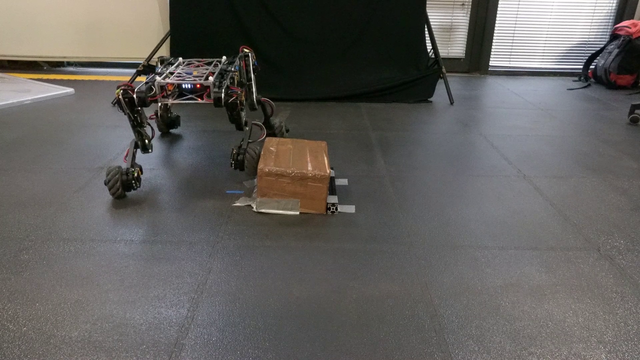}
            \put(2,4){\color{white}\bfseries (3)}
        \end{overpic}
        \label{fig:omni_seq3}
    \end{minipage}\hfill
    \begin{minipage}{0.24\textwidth}
        \centering
        \begin{overpic}[width=0.95\linewidth, trim={5cm 10cm 25cm 0cm}, clip]{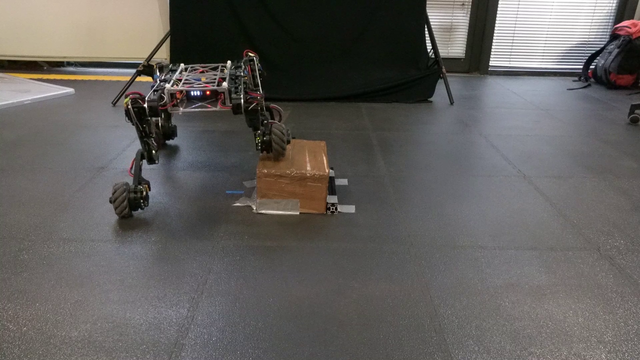}
            \put(2,4){\color{white}\bfseries (4)}
        \end{overpic}
        \label{fig:omni_seq4}
    \end{minipage}

    \vspace{1em}

    \begin{minipage}{0.24\textwidth}
        \centering
        \begin{overpic}[width=0.95\linewidth, trim={5cm 8cm 25cm 2cm}, clip]{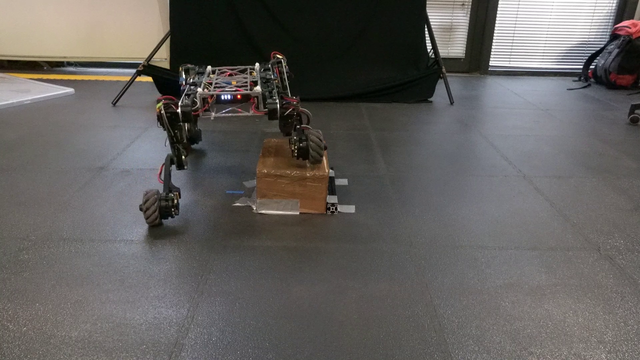}
            \put(2,4){\color{white}\bfseries (5)}
        \end{overpic}
        \label{fig:omni_seq5}
    \end{minipage}\hfill
    \begin{minipage}{0.24\textwidth}
        \centering
        \begin{overpic}[width=0.95\linewidth, trim={15cm 8cm 15cm 2cm}, clip]{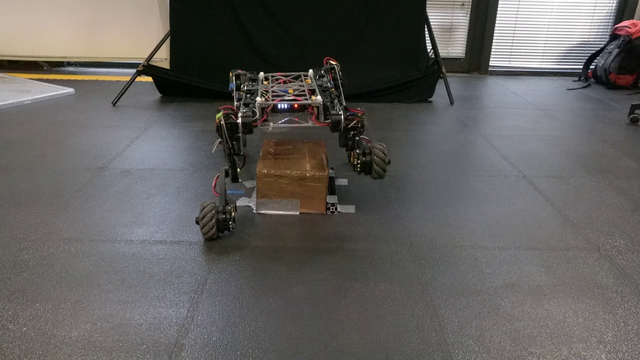}
            \put(2,4){\color{white}\bfseries (6)}
        \end{overpic}
        \label{fig:omni_seq6}
    \end{minipage}\hfill
    \begin{minipage}{0.24\textwidth}
        \centering
        \begin{overpic}[width=0.95\linewidth, trim={15cm 8cm 15cm 2cm}, clip]{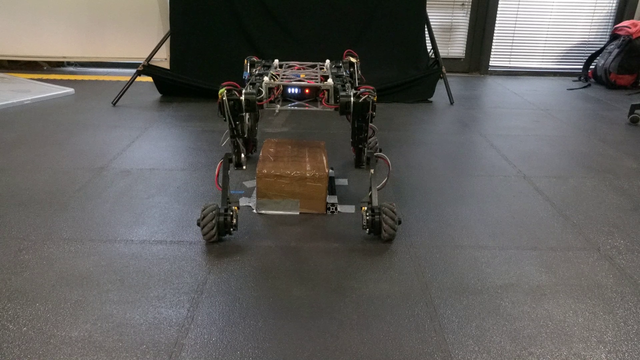}
            \put(2,4){\color{white}\bfseries (7)}
        \end{overpic}
        \label{fig:omni_seq7}
    \end{minipage}\hfill
    \begin{minipage}{0.24\textwidth}
        \centering
        \begin{overpic}[width=0.95\linewidth, trim={25cm 8cm 5cm 2cm}, clip]{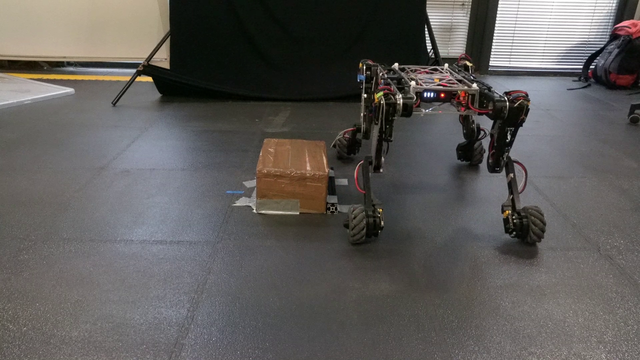}
            \put(2,4){\color{white}\bfseries (8)}
        \end{overpic}
        \label{fig:omni_seq8}
    \end{minipage}
    
    \caption{\myRed{Sequence of the motion of the OmniQuad robot during the "obstacle detection experiment"; (1) the robot starts moving left, (2) it touches the obstacle with the front left wheel, (3) it starts lifting the front left leg, (4) the leg of the robot is above the obstacle, (5) the robot starts moving using only three wheels, (6) it stops and start placing the front left wheel down to the ground, (7) it starts again moving left,(8) it overcomes the obstacle.}}
    \label{fig:omni_expe_extended}
\end{figure*}

\begin{figure}[t]
\centering
\includegraphics[width=\linewidth]{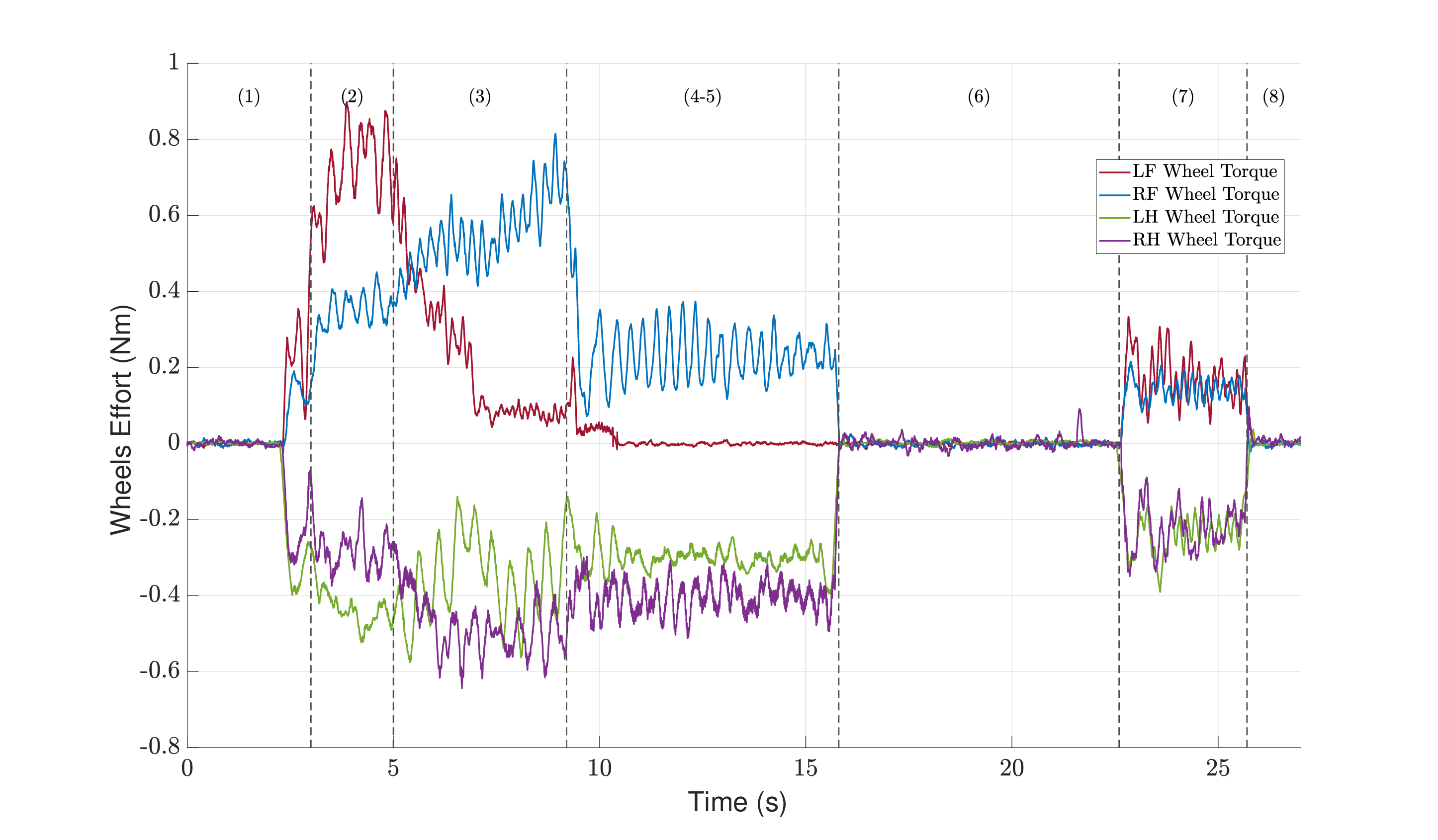}
\caption{\myRed{Torque of the four wheels during the obstacle detection experiment.}}
\label{wheel_torque}

\end{figure}

\subsubsection{Experimental setup}
This experiment was conducted to assess the robot’s capability to overcome obstacles during lateral locomotion, relying on its obstacle detection module. The primary objective was to evaluate the robot’s ability to autonomously detect the obstruction, adjust its posture, and execute a controlled ascent over the obstacle. In this test, the obstacle presents a height of 12 cm and has a box shape with dimensions 250 mm $\times$ 150 mm. 

\subsubsection{Test description}
The test commenced with the robot performing lateral motion until it encountered the obstacle, causing it to get stuck. Upon initial contact, an increase in wheel torque is experienced, which alerts the obstacle-detecting module that an obstruction has occurred. After this, the obstacle climbing module is active and starts the procedure described in section \ref{obstacle_detect}. When the torque of the wheel reaches zero starts the new phase starts in which the robot moves to avoid the obstacle. The system moves diagonally, using three wheels, considering that one is lifted. Once the traversal was completed, the robot restored its nominal configuration and resumed its lateral motion.

\subsubsection{Results and discussion}
Fig. \ref{wheel_torque} highlights the peak torque experienced by the front-left wheel, which was subsequently lifted to facilitate the obstacle-overcoming manoeuvre. During the traversal phase, the robot maintained effective locomotion with only three wheels in contact with the ground, demonstrating the robustness of the mecanum wheels and the whole actuation system. This ability to maintain stability despite one wheel being non-functional or lifted underscores the system’s capability to compensate for wheel irregularities.

\subsection{Stress Test}
\begin{figure}[t]
    \centering
    \begin{tikzpicture}
        \node[anchor=south west, inner sep=0] (image) at (0,0){%
            \includegraphics[width=\linewidth]{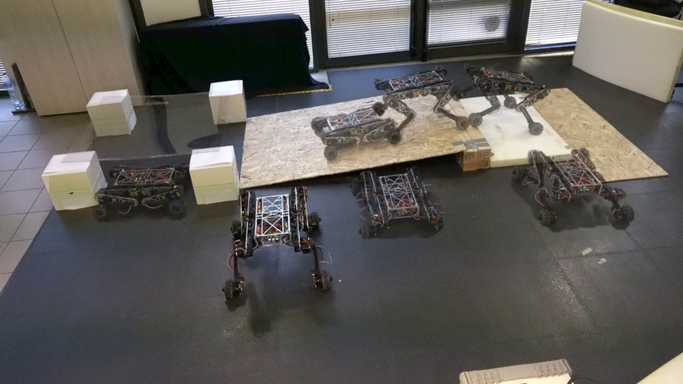}};
        
        \begin{scope}[x={(image.south east)}, y={(image.north west)}]
            \node[draw, circle, fill=white, text=black] at (0.2, 0.8) {1};
            \node[draw, circle, fill=white, text=black] at (0.4, 0.6) {2};
            \node[draw, circle, fill=white, text=black] at (0.7, 0.53) {3};
        \end{scope}
    \end{tikzpicture}
    \caption{\myRed{"Durability experiment": In location~1, the robot goes thorough a 25\,cm underpass; 
    in location~2, climbs a 5$^{\circ}$ slope; and in location~3, descends a 15\,cm step.}}
    \label{fig:long_expe}
\end{figure}

 \subsubsection{Experimental setup}
To evaluate the reliability and performance of our system, we conducted a series of experiments aimed at testing the robustness of both the hardware and the controller. These tests were designed to assess the system’s ability to withstand varying operational conditions, unexpected disturbances, and prolonged usage. 
We performed a durability experiment in which the robot navigated a series of challenging obstacles designed to test both its mechanical resilience and control robustness. 
Furthermore, we tested the maximum payload capacity during legged locomotion. The experiment involved navigating a single 3.5 cm step while carrying an incrementally increasing load after each successful trial.

\subsubsection{Test description}
During the durability experiment, the robot navigated a series of challenging obstacles designed to test both its mechanical resilience and control robustness. It first moved beneath a transparent panel before ascending a 5$^\circ$ slope. Upon reaching a 15 cm step, the robot executed a jump down, assessing its impact absorption and balance upon landing. Occasionally, instead of following the intended path, it deliberately descended laterally from the slope, demonstrating its ability to recover from unexpected orientations, Fig. \ref{fig:long_expe}. Further along the course, it lay down on a foam mattress, adapting to a soft, uneven surface. At times, the robot actively attempted to push obstacles in its path, testing its ability to resist external forces and interact dynamically with its surroundings. Throughout the trial, it was controlled via a joystick, highlighting the adaptability and robustness of both its hardware and control systems.

\begin{figure*}[t]
    \centering
    
    \begin{minipage}{0.24\textwidth}
        \centering
        \begin{overpic}[width=0.95\linewidth, trim={0cm 2cm 8cm 1cm},clip]{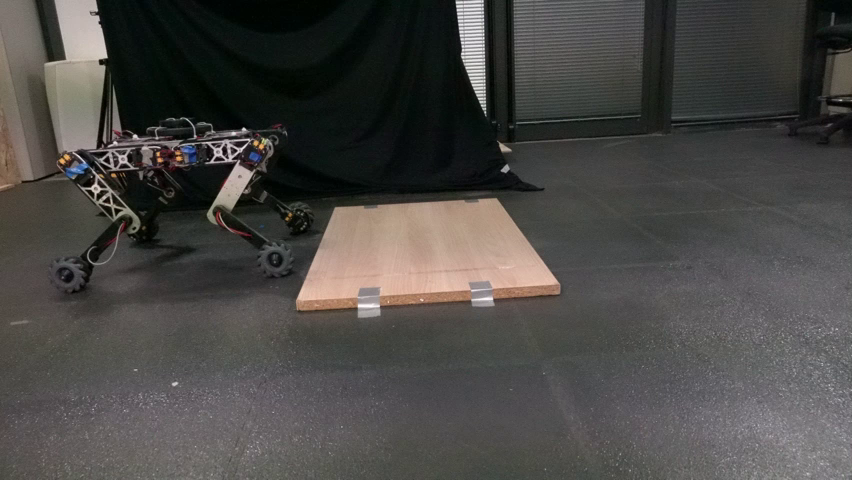}
            \put(2,4){\color{white}\bfseries (1)}
        \end{overpic}
        \label{fig:payload_seq1}
    \end{minipage}\hfill
    \begin{minipage}{0.24\textwidth}
        \centering
        \begin{overpic}[width=0.95\linewidth, trim={0cm 2cm 8cm 1cm},clip]{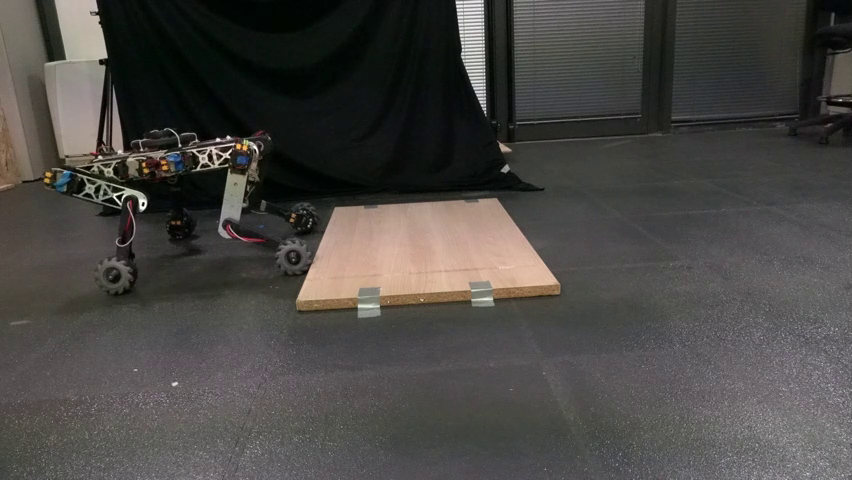}
            \put(2,4){\color{white}\bfseries (2)}
        \end{overpic}
        \label{fig:payload_seq2}
    \end{minipage}\hfill
    \begin{minipage}{0.24\textwidth}
        \centering
        \begin{overpic}[width=0.95\linewidth, trim={0cm 2cm 8cm 1cm},clip]{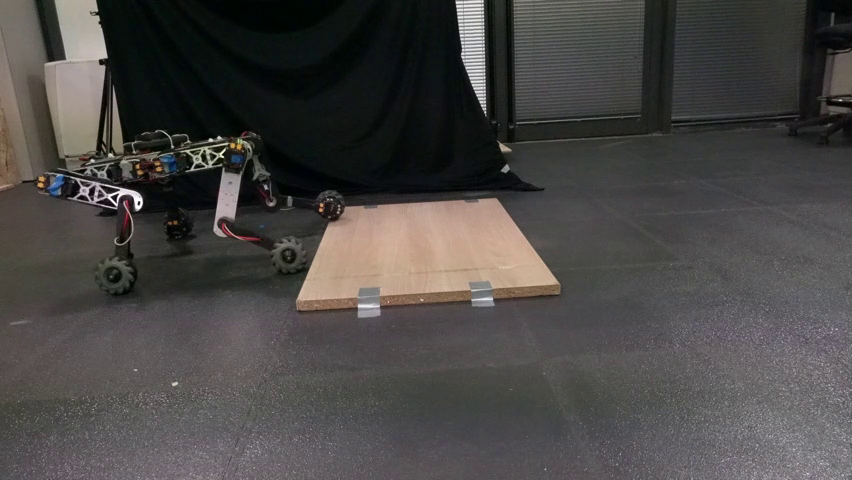}
            \put(2,4){\color{white}\bfseries (3)}
        \end{overpic}
        \label{fig:payload_seq3}
    \end{minipage}\hfill
    \begin{minipage}{0.24\textwidth}
        \centering
        \begin{overpic}[width=0.95\linewidth, trim={0cm 2cm 8cm 1cm},clip]{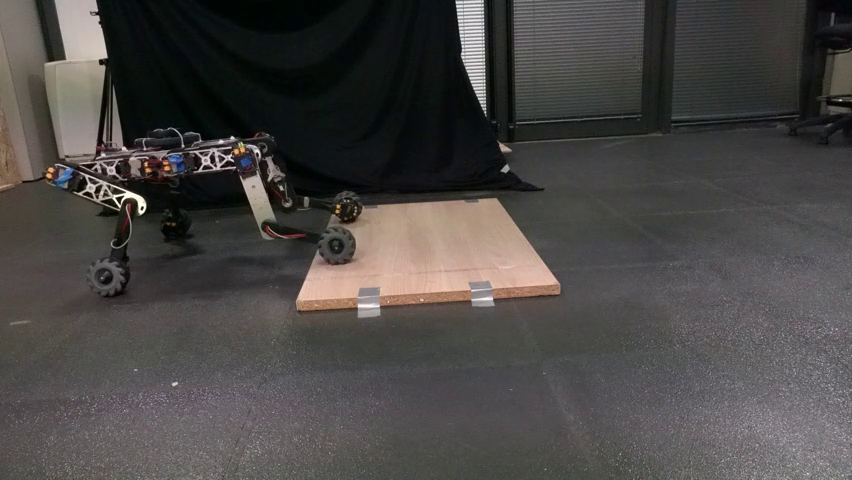}
            \put(2,4){\color{white}\bfseries (4)}
        \end{overpic}
        \label{fig:payload_seq4}
    \end{minipage}

    \vspace{1em}

    \begin{minipage}{0.24\textwidth}
        \centering
        \begin{overpic}[width=0.95\linewidth, trim={1cm 2cm 7cm 1cm},clip]{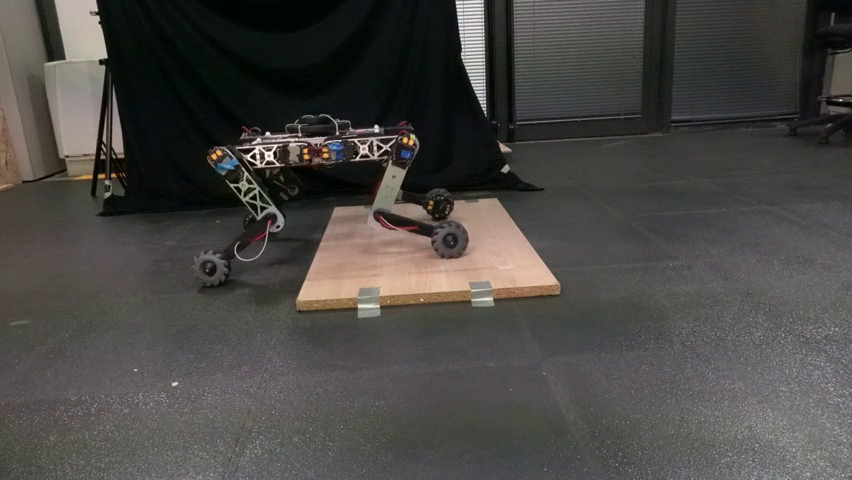}
            \put(2,4){\color{white}\bfseries (5)}
        \end{overpic}
        \label{fig:payload_seq5}
    \end{minipage}\hfill
    \begin{minipage}{0.24\textwidth}
        \centering
        \begin{overpic}[width=0.95\linewidth, trim={2cm 2cm 6cm 1cm},clip]{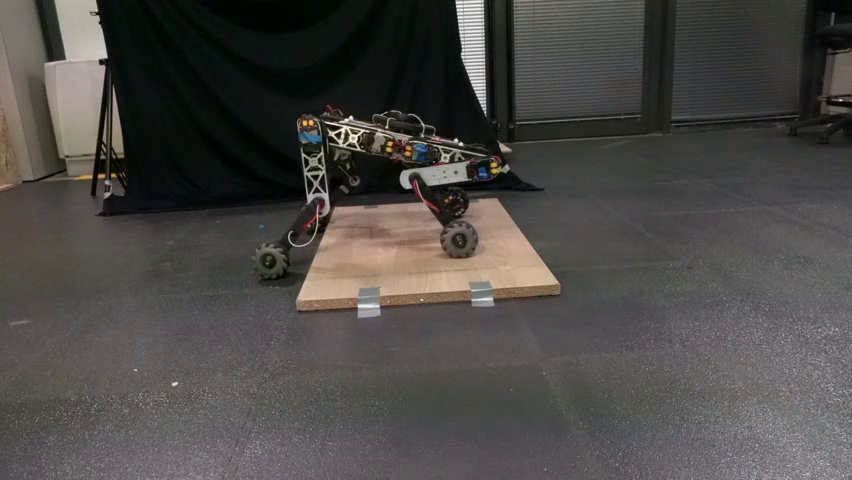}
            \put(2,4){\color{white}\bfseries (6)}
        \end{overpic}
        \label{fig:payload_seq6}
    \end{minipage}\hfill
    \begin{minipage}{0.24\textwidth}
        \centering
        \begin{overpic}[width=0.95\linewidth, trim={2cm 2cm 6cm 1cm},clip]{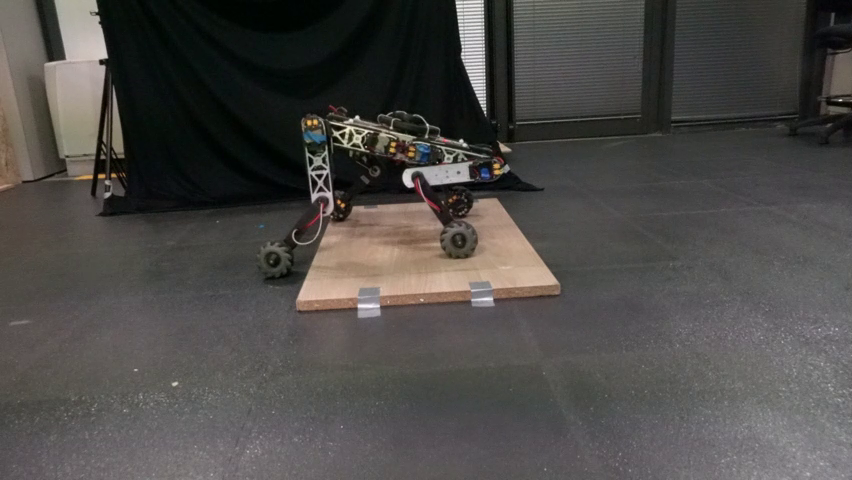}
            \put(2,4){\color{white}\bfseries (7)}
        \end{overpic}
        \label{fig:payload_seq7}
    \end{minipage}\hfill
    \begin{minipage}{0.24\textwidth}
        \centering
        \begin{overpic}[width=0.95\linewidth, trim={3cm 2cm 5cm 1cm},clip]{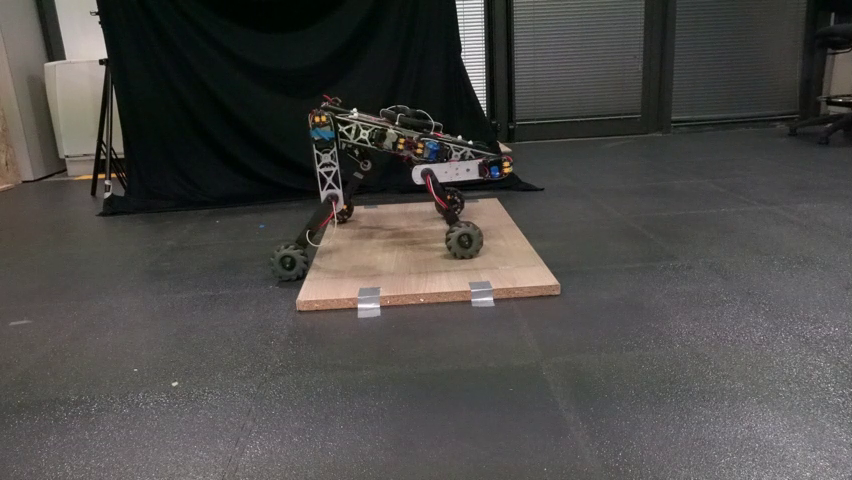}
            \put(2,4){\color{white}\bfseries (8)}
        \end{overpic}
        \label{fig:payload_seq8}
    \end{minipage}

    \vspace{1em}

    \begin{minipage}{0.24\textwidth}
        \centering
        \begin{overpic}[width=0.95\linewidth, trim={3cm 2cm 5cm 1cm},clip]{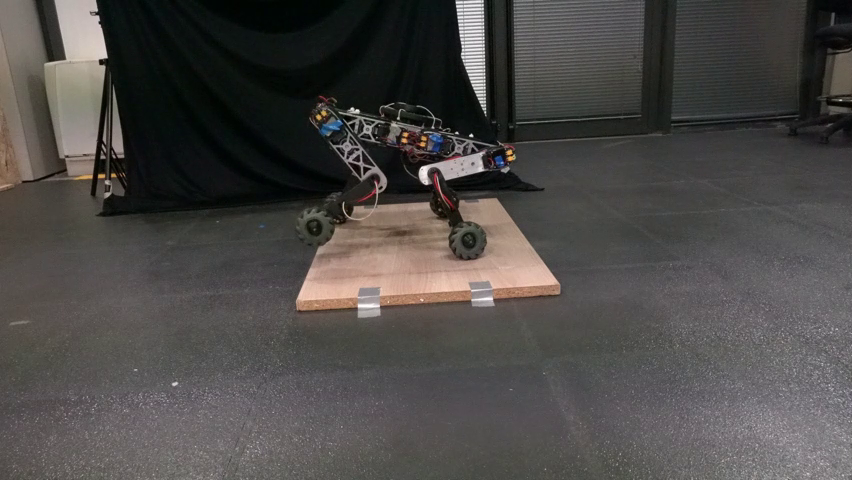}
            \put(2,4){\color{white}\bfseries (9)}
        \end{overpic}
        \label{fig:payload_seq9}
    \end{minipage}\hfill
    \begin{minipage}{0.24\textwidth}
        \centering
        \begin{overpic}[width=0.95\linewidth, trim={4cm 2cm 4cm 1cm},clip]{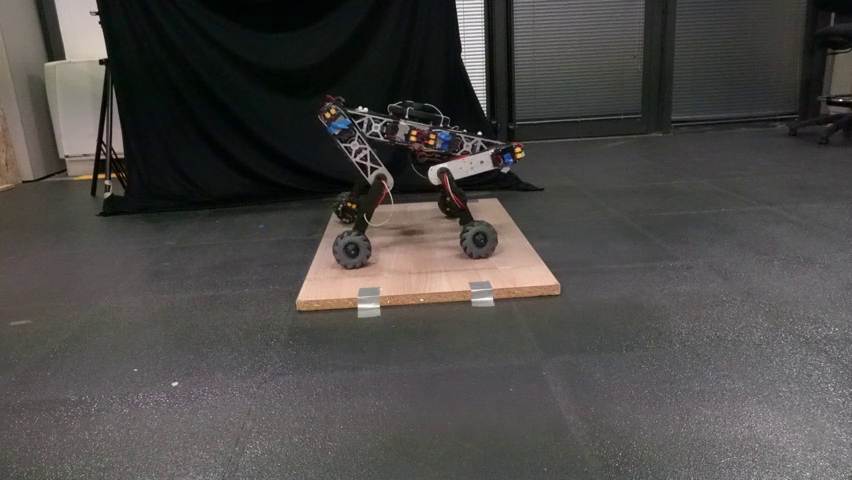}
            \put(2,4){\color{white}\bfseries (10)}
        \end{overpic}
        \label{fig:payload_seq10}
    \end{minipage}\hfill
    \begin{minipage}{0.24\textwidth}
        \centering
        \begin{overpic}[width=0.95\linewidth, trim={6cm 2cm 2cm 1cm},clip]{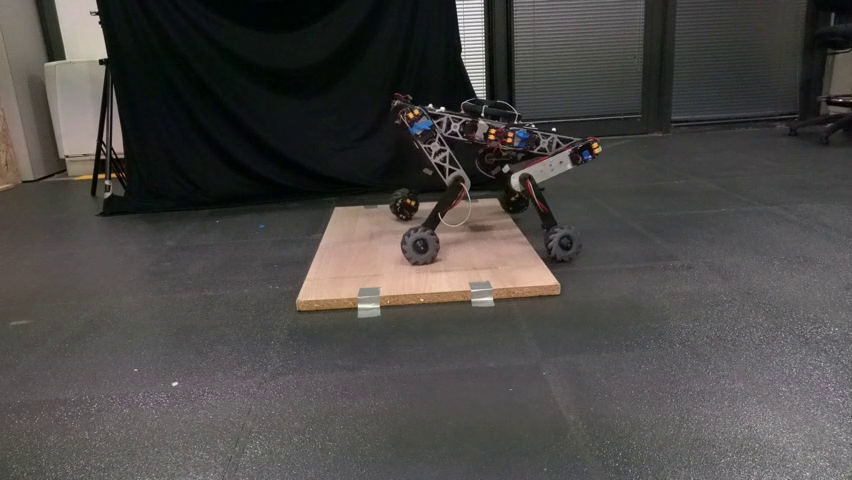}
            \put(2,4){\color{white}\bfseries (11)}
        \end{overpic}
        \label{fig:payload_seq11}
    \end{minipage}\hfill
    \begin{minipage}{0.24\textwidth}
        \centering
        \begin{overpic}[width=0.95\linewidth, trim={8cm 2cm 0cm 1cm},clip]{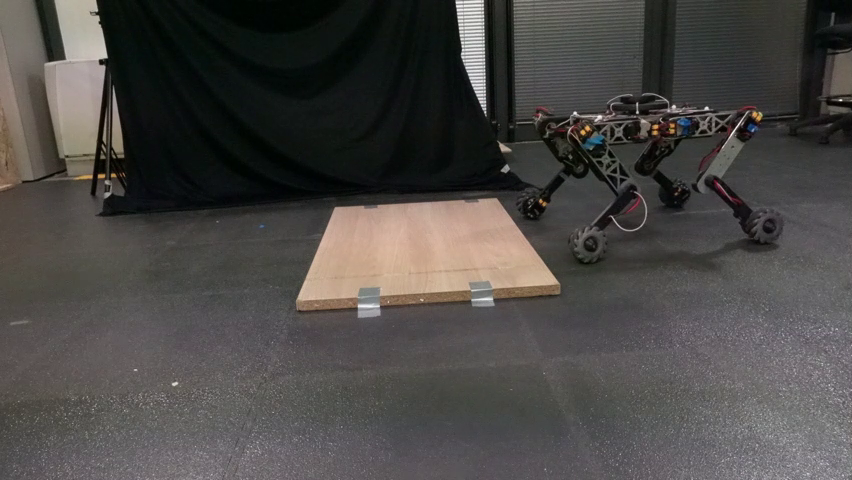}
            \put(2,4){\color{white}\bfseries (12)}
        \end{overpic}
        \label{fig:payload_seq12}
    \end{minipage}

    \caption{\myRed{Sequence of the motion of the OmniQuad robot overcoming a step of 3.5 cm height with an additional payload of 2.5 kg. (1) the robot approaches the obstacle, (2)  shifts the base backwards, (3)  places the front left wheel on the top of the step, (4)  places the front right wheel on the top of the step, (5)  moves forward using wheels and adjusts the base, (6)  tilts the base forward, (7)  places the hind left wheel on the top of the step, (8), tilts the base to the left, (9) lifts the hind right wheel, (10) places it on the step, (11) moves forward and adjusts the base, (12) brings the base back to horizontal.}}
    \label{fig:payload_extended_1}
\end{figure*}

In the payload experiments, a series of tests were conducted to evaluate the maximum payload capacity of the robot during legged locomotion. The load was progressively increased from 0.5 kg to 2.5 kg to assess its effect on mobility and stability. During the experiment, the obstacle detecting and climbing stay on, allowing the robot to successfully overcome the step, Fig. \ref{fig:payload_extended_1}.

\subsubsection{Results and discussion}

The trials of the durability tests were conducted continuously, with the robot performing two full experimental runs lasting 5 and 6 minutes, respectively, without requiring intermediate breaks. Throughout both trials, the system operated without any hardware malfunctions or software failures, confirming the reliability of the control algorithms and the structural integrity of the mechanical components. The joystick-controlled interface remained responsive, ensuring precise command execution and real-time adaptability to environmental variations.

\begin{figure*}[t]
\begin{center}

    \subfloat[\scriptsize Torque on the front legs]{
        \includegraphics[width=0.48\textwidth,trim = {3.3cm 0.5cm 4.2cm 1cm},clip]{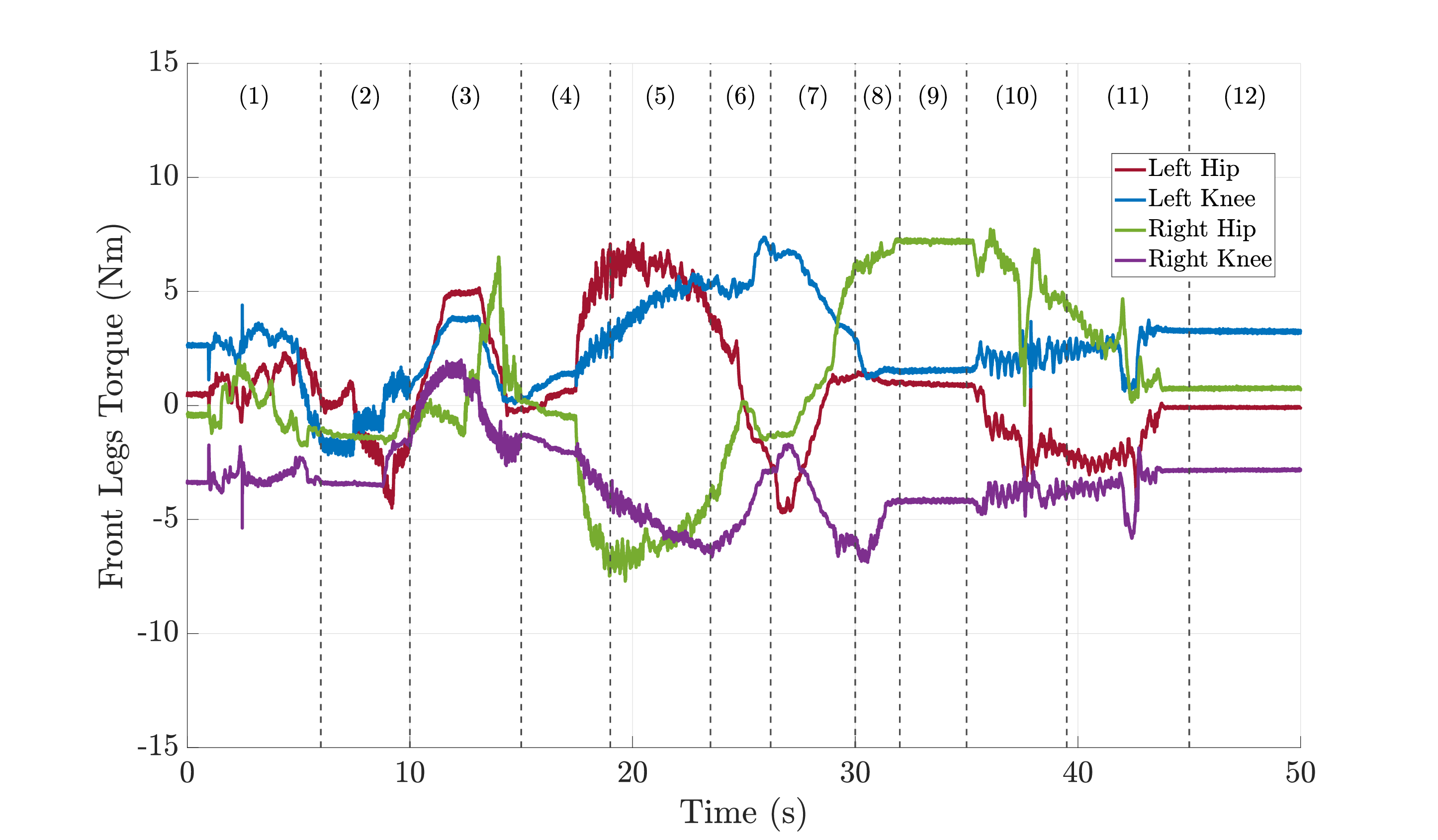}
        \label{fig:load_front}}\hfill
    \subfloat[\scriptsize Torque on the hind legs]{
        \includegraphics[width=0.48\textwidth,trim = {3.3cm 0.5cm 4.2cm 1cm},clip]{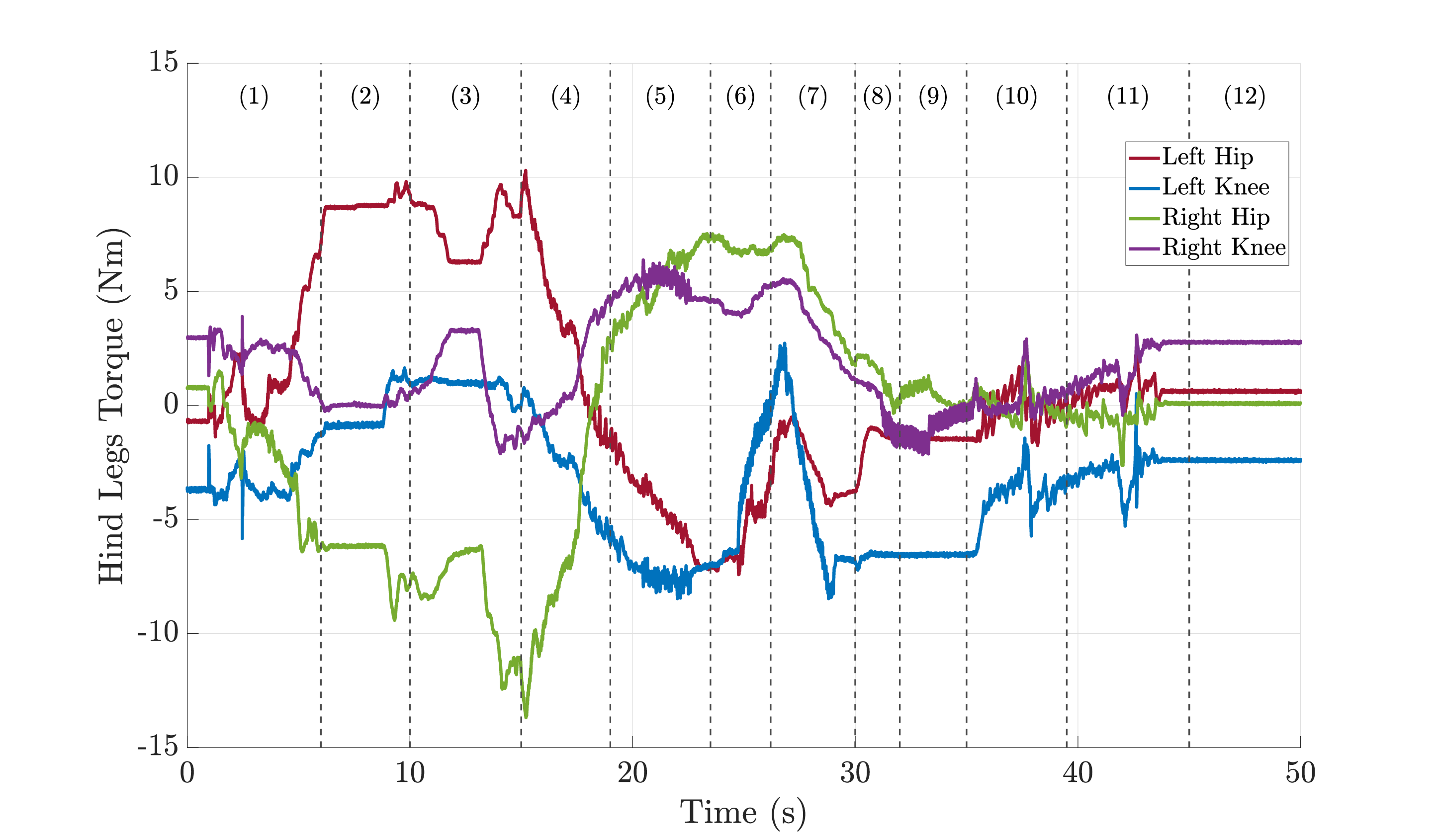}
         \label{fig:load_hind}}

\caption{Torque on the robot joints, while the robot is performing the 3.5 cm step with a 2.5 kg payload.}
\label{payload_results}
\end{center}
\end{figure*}

During the payload test, the robot successfully stepped on the obstacle carrying a total load of 2.5 kg. Fig. \ref{payload_results}  shows the torque affecting the leg joint, which remains for the majority of the task below 10 Nm; the maximum peak torque of the EM-Act actuators is 19.6 Nm. However, when attempting to carry a 3 kg payload, the system experienced a critical failure. The motor driver overheated, leading to a loss of control and an eventual fall.

\subsection{Energy Consumption}

Energy consumption analysis is effectively conducted by monitoring battery power usage over time. Experimental observations highlight significant variations in battery life dependent upon the robot's locomotion mode, specifically when comparing wheeled versus legged configurations. On flat terrain, the wheels primarily facilitate propulsion, while the legs maintain the robot's base at a predefined height.

\begin{figure}[htbp]
    \centering
    \subfloat[\mygreen{\scriptsize Mean values of the current for the leg motors when the robot maintains the base fixed. The average current per leg motor is approximately = $6.05~\text{A}$.}\label{fig:legs-fix}]
    {
        \includegraphics[width=0.3\linewidth, trim=2cm 0cm 3.8cm 0.5cm, clip]{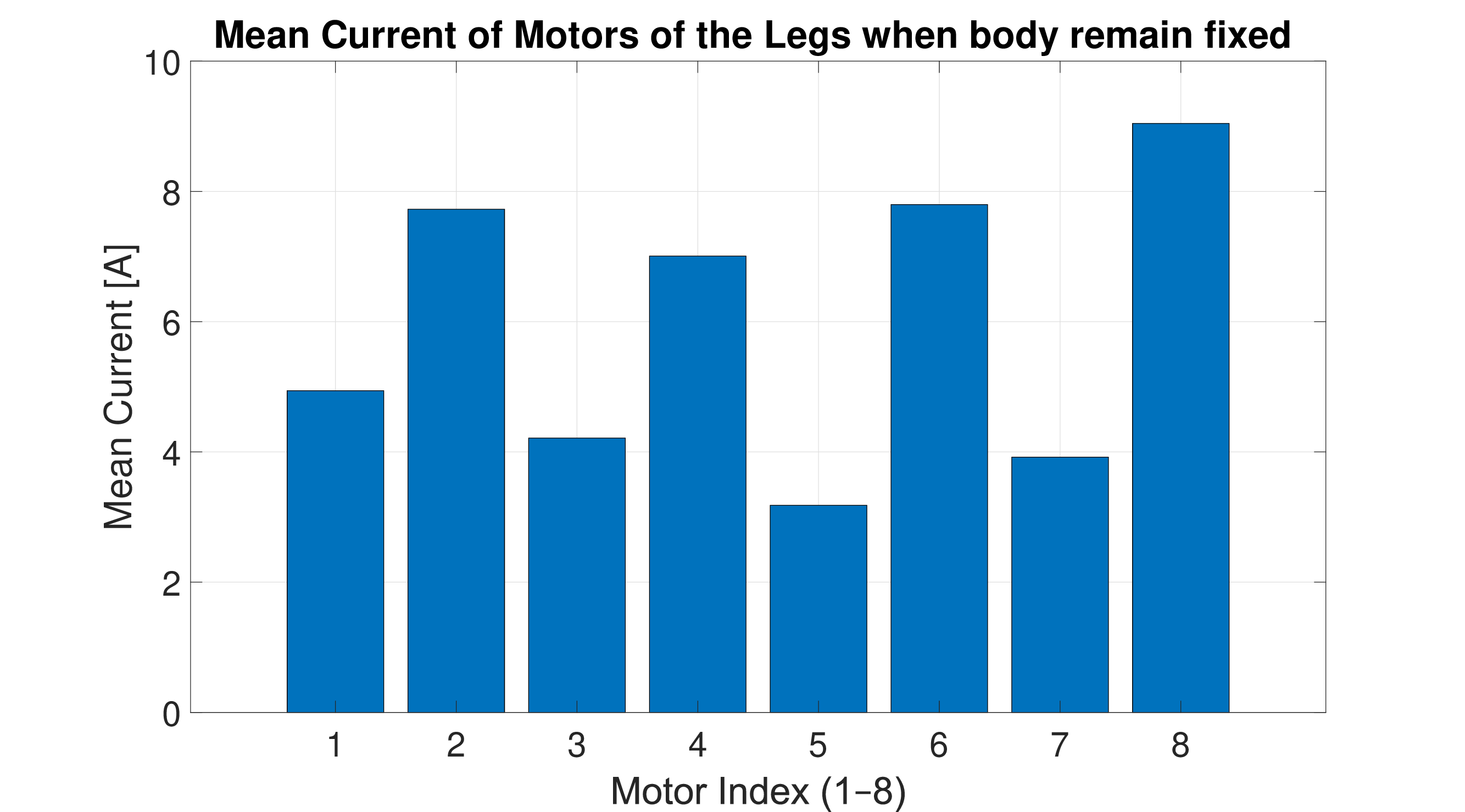}
    }
    \hfill
    \subfloat[\mygreen{\scriptsize Mean values of the current for the wheel motors when the robot is moving with the wheels. The average current per wheel motor is approximately = $0.44~\text{A}$.}\label{fig:wheels}]
    {
        \includegraphics[width=0.3\linewidth, trim=2cm 0cm 3.8cm 0.5cm, clip]{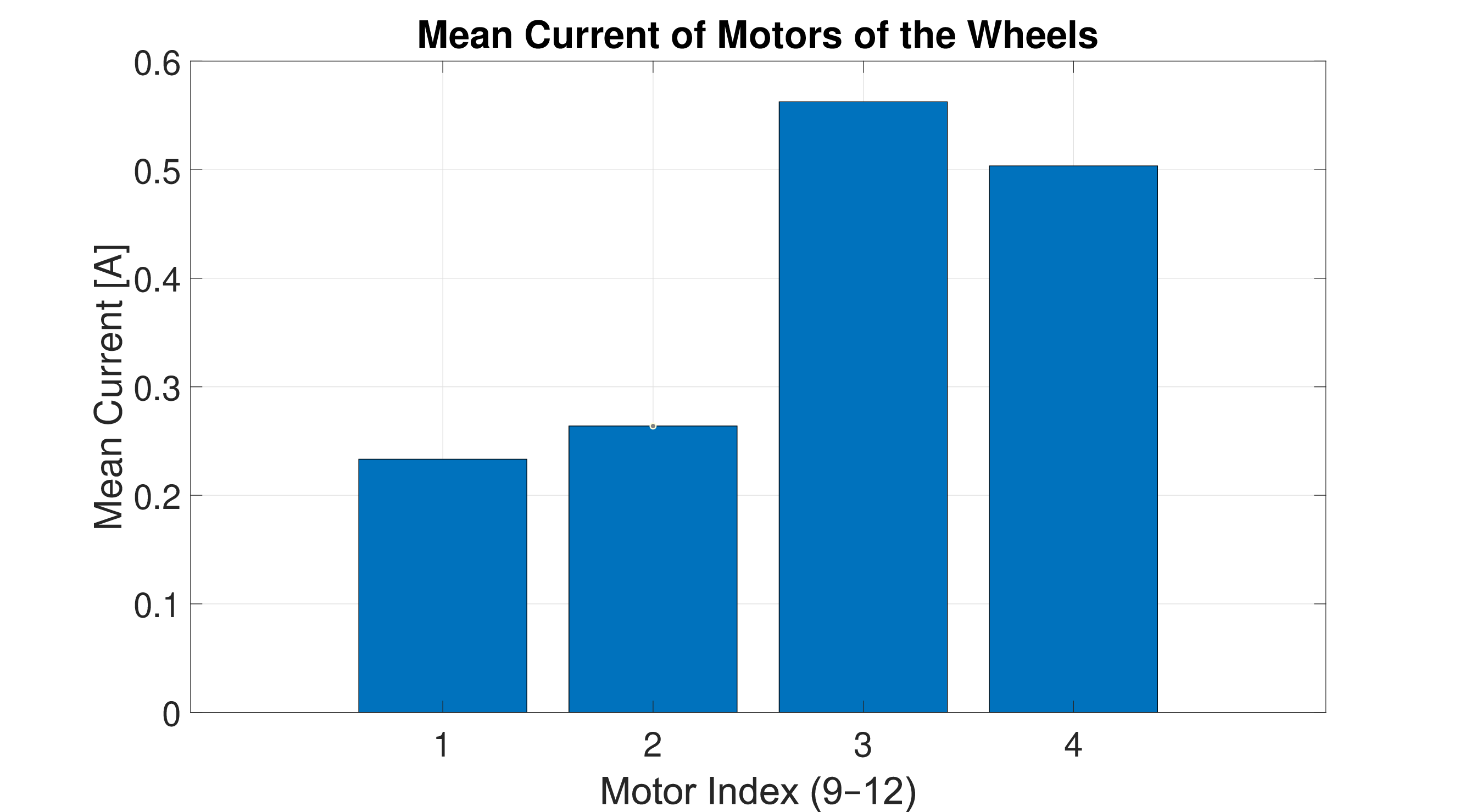}
    }
    \hfill
    \subfloat[\mygreen{\scriptsize Mean values of the current for the leg motors when the robot lifts or lowers the base. The average current per leg motor is approximately = $9.46~\text{A}$.}\label{fig:legs}]
    {
        \includegraphics[width=0.3\linewidth, trim=2cm 0cm 3.8cm 0.5cm, clip]{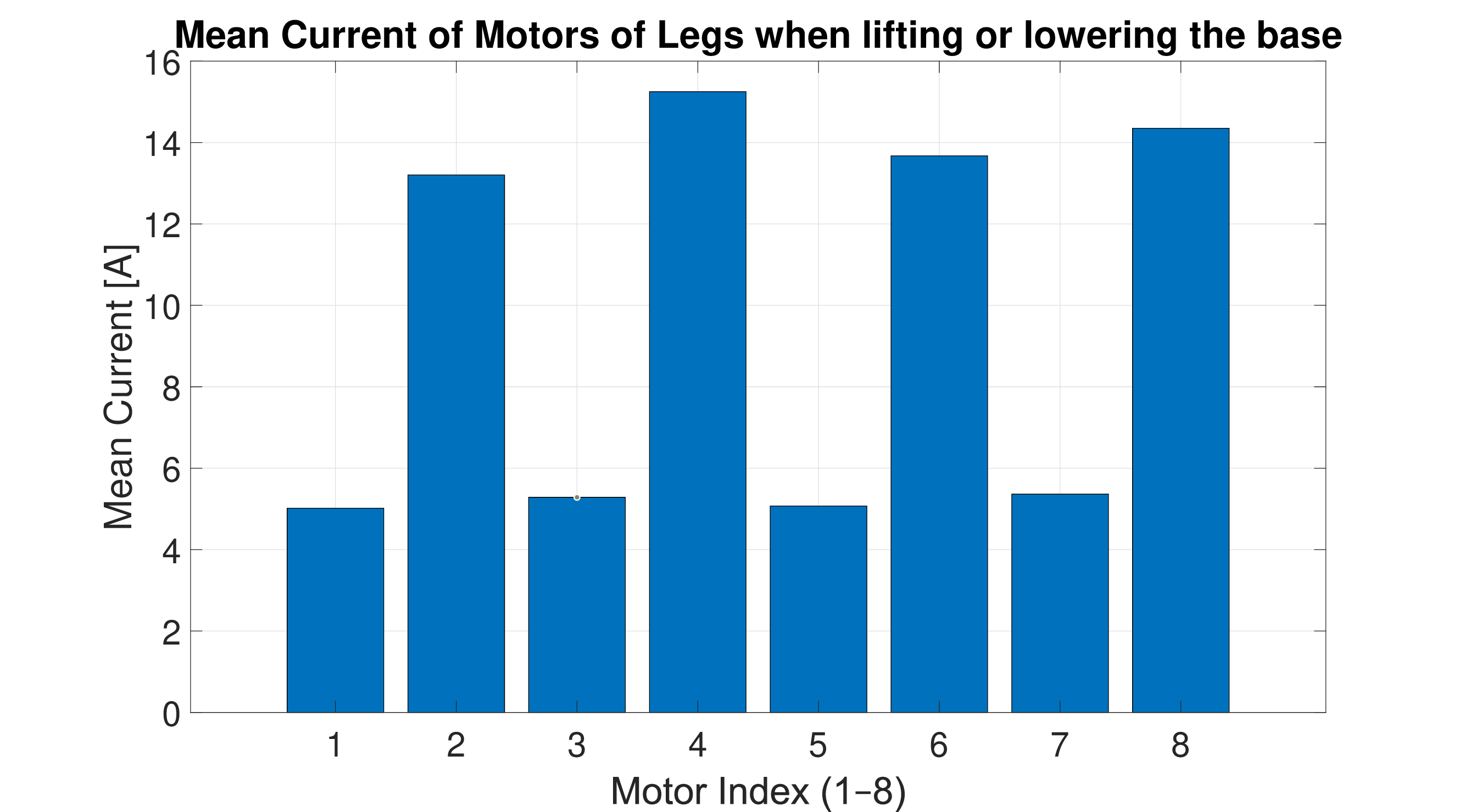}
    }
    \caption{\mygreen{Mean values of the motors' current when the robot is moving on flat terrain.}}
    \label{fig:combined}
\end{figure}

Fig.s~\ref{fig:legs-fix} and~\ref{fig:wheels} illustrate the average current drawn by each motor driver over a 100-second interval, clearly showing that the leg motors consume notably higher current than the wheel motors.
The mean current across all motors is calculated as

\begin{equation}
I_{\text{m}} = \frac{1}{N} \sum_{i=1}^{N} \left( \frac{1}{T_0} \int_{0}^{T_0} I_{\text{mot}, i}(t) \, dt \right),
\end{equation}

where \textit{N} is the number of motors and \textit{I\textsubscript{mot},i} is the instantaneous current drawn by the \textit{i}‑th motor at time \textit{t}, and \textit{T\textsubscript{o}} is the duration of the measurement interval.

To estimate battery usage, the mean current is converted to energy consumption in Watt-hours (Wh) using

\begin{equation}
P_{\text{c}} = V \times I_{\text{m}} \times \frac{T_{\text{o}}}{3600},
\end{equation}

where \textit{P\textsubscript{c}} denotes the energy consumed in watt-hours, \textit{V} is the supply voltage, \textit{T\textsubscript{o}} is the operation time in seconds, and \textit{I\textsubscript{m}} is the average current in Ampere across all motors.
As an example, using the values reported in the graphs, Fig.\ref{fig:legs-fix},  \ref{fig:wheels}, we compute the total average current by separating the contributions from the leg and wheel motors, as their mean currents differ significantly. Specifically, the mean current per leg motor is approximately \( I_{\text{mot,l}} = 6.05~\text{A} \), while for the wheel motors it is \( I_{\text{mot,w}} = 0.44~\text{A} \). With 8 leg motors and 4 wheel motors, the total average current becomes

\[
I_{\text{tot}} = 8 \times 6.05~\text{A} + 4 \times 0.44~\text{A} = 50.2~\text{A},
\]

assuming an operation time of \( T_{\text{o}} = 100~\text{s} \) and a supply voltage of \( V = 24~\text{V} \), the total energy consumption is estimated as

\[
P_{\text{c}} = V \times I_{\text{tot}} \times \frac{T_{\text{o}}}{3600} = 24 \times 50.2 \times \frac{100}{3600} \approx 34.9~\text{Wh},
\]

given a battery capacity of \( 99.90~\text{Wh} \), this corresponds to an estimated battery life of

\[
\frac{99.90~\text{Wh}}{34.9~\text{Wh}} \approx 2.9~\text{hours},
\]

which can be reasonably approximated as 3 hours.
However, even when considering a little motion of the legs, in this case, only lowering and lifting the base when the robot is moving using the wheels, shows an increase of 50 $\%$ in the current absorption by the leg motors, Fig. \ref{fig:legs}, which takes the estimated battery life to 2 hours. This indicates higher energy expenditure during active leg motions. Consequently, adopting a hybrid locomotion approach, which requires the robot to overcome different types of obstacles (combining wheeled and legged modes), yields an average battery life of approximately 1.5 hours, as detailed in Table \ref{spec_omniquad}.

\subsection{Comparison with state-of-the-art approaches}

Several approaches have been explored to develop wheel-leg composite control to optimize stability, efficiency, and adaptability to challenging terrains. In \cite{robo_cool}, the authors emphasize mechanical design, prioritizing a simplified control strategy with omnidirectional wheeled locomotion while minimizing leg movements. Similarly, \cite{NI2022} focuses on the mechanical structure to mitigate impact forces, with control strategies ensuring balance on uneven terrains. Meanwhile, \cite{8998367} concentrates on leg design to minimize impact, integrating a control strategy that stabilizes the robot’s base on rough surfaces. On the other hand, \cite{8642912} takes a different approach by developing an MPC controller to optimize travel time in legged locomotion, particularly in forward and backward movements, given the robot's car-like constraints in steering. In \cite{bitnaza}, an MPC-based whole-body stability control is implemented to handle the complexity of a six-legged robot with multiple degrees of freedom.

In our case, we aimed to develop a multimodal robot equipped with both mecanum wheels and legs to navigate challenging environments. Our focus was twofold: on the mechanical side, we sought to design a wheeled foot that is compact yet powerful enough to support traversal over difficult settings; and on the control side, we developed a strategy to leverage both wheels and legs for efficient climbing and manoeuvrability.

\color{black}
\section{Conclusion and Future Works}
We have developed a novel type of robot, named OmniQuad, which integrates both mecanum wheels and legs to enhance stability and maneuverability while mitigating critical steering issues. The robot features omnidirectional wheels attached to its feet, enabling smooth and precise movement in all directions. Additionally, the incorporation of legs provides OmniQuad with the capability to climb obstacles, traverse uneven terrain, and dynamically adjust its height to pass beneath obstructions or navigate confined spaces, significantly expanding its operational versatility. To govern its locomotion, we devised a control system grounded in the kinematics of both locomotion modes. In Tab. \ref{spec_omniquad} are reported the specifications of the robot.
\mygreen{Experiments conducted both indoors and in controlled scenarios showcased OmniQuad's adeptness at reliably navigating challenging paths and overcoming obstacles, either by effectively passing beneath or scaling over them. Notably, the robot demonstrated the capability to reduce its height by up to 180 mm, allowing it to seamlessly traverse low-clearance obstacles, as well as to ascend steps as high as 100 mm, even under varying payload conditions. Additionally, its agility in executing precise movements while consistently maintaining its orientation and directionality underscores its versatility and suitability for demanding inspection tasks, logistics operations, and search-and-rescue missions across structured and natural environments.}

The robot has also been presented at the Maker Faire Rome 2023 and at the European Robotics Forum 2024 held in Rimini, Italy.

In our future work, we intend to enhance the design of the wheeled foot. Specifically, we aim to reduce the total length of the component for a more compact configuration. Regarding the design part, we are planning to increase the number of DOF of each leg, introducing the hip abduction joint, in order to improve the stability of the robot. \mygreen{In this work, we achieve a fundamental level of autonomy for the Omniquad robot by leveraging encoder-based odometry and orientation data, allowing it to reliably navigate structured environments with static obstacles. Although the current autonomy is sufficient for simple scenarios, our future goals involve significantly enhancing its navigation capabilities. Specifically, we aim to incorporate perceptive sensors such as a depth camera and a LiDAR sensor, empowering the robot to autonomously navigate through unknown and dynamic environments. Additionally, we will employ an end-to-end approach \cite{rudin2022learning,Scaldaferri2025otto} to refine the control loop to enable hybrid motion \cite{Lee2024learning}, seamlessly integrating both legged and wheeled systems, and improving locomotion in natural environments by addressing challenges such as terrain recognition, dynamic balance management, and efficient transition strategies between locomotion modes.}

\begin{table}[t]
\begin{center}

\caption{\myRed{OmniQuad robot datasheet}}
\label{spec_omniquad}
\begin{tabular} {l@{\hspace{1.0cm}} l@{\hspace{4.0cm}} l@{\hspace{2.0cm}} }
\hline
   & \textbf{Property} & \textbf{Value}\\
\hline
   & Number of DOF & 12\\

   & Maximum forward velocity & 2 m/s\\

   & Maximum lateral velocity & 1.5 m/s\\

   & Battery capacity & 4500 mAh\\

   & Battery Voltage & 24 V\\

   & Battery Life & 1.5 h\\

   & Robot communication & Ethernet, Wi-Fi 5 GHz\\

   & Software & ROS2 compatible software\\

   & Low-level motor controller & PID anti-windup\\

   & Low-level controller frequency & 30 kHz\\
\hline
\end{tabular}
\end{center}
\vspace{-6mm}
\end{table}

\vspace{-2mm}
\section{Acknowledgment}
Published version available at: 

\noindent \url{https://www.sciencedirect.com/science/article/abs/pii/S0094114X25002149}.
\vspace{2mm}

This research is partially supported by the European Union’s HORIZON-MSCA-2023-SE-01-01 - MSCA Staff Exchanges 2023 Program under the Grant Agreement No. 101182891 (NEUTRAWEED), Horizon 2020 Research and Innovation Programme under Grant Agreement No. 101016970 (Natural Intelligence), and Horizon Research and Innovation Programme under Grant Agreement No. 101070596 (euROBIN), in part by the ARCADIT project, funded by the National Center of Research on High Performance Computing, Big Data and Quantum Computing , through an Innovation Grant assigned to UNIPISA and Italferr (CUP: I53C22000690001), and in part by the Italian Ministry of Education and Research in the framework of the “FoReLab” (Future-oriented Research Lab) Project (Departments of Excellence). F. Iotti is within the Italian Doctorate of Robotics and Intelligent Machines. The authors would like to thank Lorenzo Martignetti, Giuseppe Alfonso, and Jacopo Cioni from the University of Pisa for their suggestions on the system development.
\vspace{-4mm}
\bibliographystyle{IEEEtran}
\bibliography{sample,SOA}

@article{momaro,
author = {Schwarz, Max and Rodehutskors, Tobias and Droeschel, David and Beul, Marius and Schreiber, Michael and Araslanov, Nikita and Ivanov, Ivan and Lenz, Christian and Razlaw, Jan and Schüller, Sebastian and Schwarz, David and Topalidou-Kyniazopoulou, Angeliki and Behnke, Sven},
year = {2017},
month = {03},
pages = {400-425},
title = {NimbRo Rescue: Solving Disaster-Response Tasks through Mobile Manipulation Robot Momaro},
volume = {34},
journal = {Journal of Field Robotics},
doi = {10.1002/rob.21677}
}

@INPROCEEDINGS{omniwheg, 
author={Cao, Ruixiang and Gu, Jun and Yu, Chen and Rosendo, Andre}, 
booktitle={2022 IEEE/RSJ International Conference on Intelligent Robots and Systems (IROS)},  
title={OmniWheg: An Omnidirectional Wheel-Leg Transformable Robot}, 
year={2022}, 
volume={}, 
number={}, 
pages={5626-5631}, 
doi={10.1109/IROS47612.2022.9982030}}

@article{wheels-knees, 
author = {Li, Wujing and Wei, Linchao and Zhang, Xiaochen}, 
year = {2022}, 
month = {09}, 
pages = {9239}, 
title = {A Wheels-on-Knees Quadruped Assistive Robot to Carry Loads},
volume = {12}, 
journal = {Applied Sciences}, 
doi = {10.3390/app12189239}}

@ARTICLE{Ranjan2025,
  author={Ranjan, Alok and Iotti, Francesco and Angelini, Franco and Garabini, Manolo},
  journal={IEEE/ASME Transactions on Mechatronics}, 
  title={TraQuad: A Modular Tracked Legged Multimodal Quadrupedal Robot}, 
  year={2025},
  volume={},
  number={},
  pages={1-13},
  doi={10.1109/TMECH.2025.3560078}}

@ARTICLE{zhou2024,
  author={Zhou, Qinqin and Yang, Sicheng and Jiang, Xinyang and Zhang, Dongsheng and Chi, Wanchao and Chen, Ke and Zhang, Shenghao and Li, Jie and Zhang, Jingfan and Wang, Rui and Li, Jingchen and Zhang, Yufeng and Wang, Haitao and Wang, Shuai and Xiang, Lingzhu and Zheng, Yu and Zhang, Zhengyou},
  journal={IEEE Transactions on Automation Science and Engineering}, 
  title={Max: A Wheeled-Legged Quadruped Robot for Multimodal Agile Locomotion}, 
  year={2024},
  volume={21},
  number={4},
  pages={7562-7582},
  doi={10.1109/TASE.2023.3345876}}

@ARTICLE{8642912,
  author={Bjelonic, Marko and Bellicoso, C. Dario and de Viragh, Yvain and Sako, Dhionis and Tresoldi, F. Dante and Jenelten, Fabian and Hutter, Marco},
  journal={IEEE Robotics and Automation Letters}, 
  title={Keep Rollin’—Whole-Body Motion Control and Planning for Wheeled Quadrupedal Robots}, 
  year={2019},
  volume={4},
  number={2},
  pages={2116-2123},
  doi={10.1109/LRA.2019.2899750}}

@ARTICLE{robo_cool,
  author={Yun, Seong-Ho and Park, Jaehyun and Seo, Jiwon and Kim, Yong-Jae},
  journal={IEEE Robotics and Automation Letters}, 
  title={Development of an Agile Omnidirectional Mobile Robot With GRF Compensated Wheel-leg Mechanisms for Human Environments}, 
  year={2021},
  volume={6},
  number={4},
  pages={8301-8308},
  doi={10.1109/LRA.2021.3098954}}

@ARTICLE{mec_hexa,
  author={Orozco-Magdaleno, Ernesto C. and Gómez-Bravo, Fernando and Castillo-Castañeda, Eduardo and Carbone, Giuseppe},
  journal={IEEE/ASME Transactions on Mechatronics}, 
  title={Evaluation of Locomotion Performances for a Mecanum-Wheeled Hybrid Hexapod Robot}, 
  year={2021},
  volume={26},
  number={3},
  pages={1657-1667},
  doi={10.1109/TMECH.2020.3027259}}

@ARTICLE{Centauro,
  author={Kashiri, Navvab and Baccelliere, Lorenzo and Muratore, Luca and Laurenzi, Arturo and Ren, Zeyu and Hoffman, Enrico Mingo and Kamedula, Malgorzata and Rigano, Giuseppe Francesco and Malzahn, Jorn and Cordasco, Stefano and Guria, Paolo and Margan, Alessio and Tsagarakis, Nikos G.},
  journal={IEEE Robotics and Automation Letters}, 
  title={CENTAURO: A Hybrid Locomotion and High Power Resilient Manipulation Platform}, 
  year={2019},
  volume={4},
  number={2},
  pages={1595-1602},
  doi={10.1109/LRA.2019.2896758}}

@INPROCEEDINGS{Hytro,
  author={Lu, Dongping and Dong, Erbao and Liu, Chunshan and Xu, Min and Yang, Jie},
  booktitle={2013 IEEE/RSJ International Conference on Intelligent Robots and Systems}, 
  title={Design and development of a leg-wheel hybrid robot “HyTRo-I”}, 
  year={2013},
  volume={},
  number={},
  pages={6031-6036},
  doi={10.1109/IROS.2013.6697232}}

@INPROCEEDINGS{omni_feet,
  author={Bu, Yongzhong and Bu, Yongjian and Li, Haoran and Mao, Shixin and Zhu, Haifei},
  booktitle={2022 IEEE International Conference on Robotics and Biomimetics (ROBIO)}, 
  title={Development of Wheel-leg Hybrid Climbing Robot with Switchable Permanent Magnetic Omni-wheels as Feet}, 
  year={2022},
  volume={},
  number={},
  pages={1-6},
  doi={10.1109/ROBIO55434.2022.10011830}}

@INPROCEEDINGS{9636371,
  author={Bjelonic, Marko and Grandia, Ruben and Harley, Oliver and Galliard, Cla and Zimmermann, Samuel and Hutter, Marco},
  booktitle={2021 IEEE/RSJ International Conference on Intelligent Robots and Systems (IROS)}, 
  title={Whole-Body MPC and Online Gait Sequence Generation for Wheeled-Legged Robots}, 
  year={2021},
  volume={},
  number={},
  pages={8388-8395},
  doi={10.1109/IROS51168.2021.9636371}}

@INPROCEEDINGS{9734076,
  author={Zhang, Haolin and Zhuang, Zheming and Wei, Wei and Guan, Yuntao and Li, Jinfu and Dai, Jiansheng},
  booktitle={2022 8th International Conference on Mechatronics and Robotics Engineering (ICMRE)}, 
  title={Structure Design and Motion Control of a Hybrid Quadruped Robot with Wheels and Legs}, 
  year={2022},
  volume={},
  number={},
  pages={122-128},
  doi={10.1109/ICMRE54455.2022.9734076}}

@ARTICLE{9204432,
  author={Chen, Jiawei and Xu, Kun and Ding, Xilun},
  journal={IEEE/ASME Transactions on Mechatronics}, 
  title={Roller-Skating of Mammalian Quadrupedal Robot With Passive Wheels Inspired by Human}, 
  year={2021},
  volume={26},
  number={3},
  pages={1624-1634},
  doi={10.1109/TMECH.2020.3025942}}

@article{CHEN2021367,
title = {Control strategy of stable walking for a hexapod wheel-legged robot},
journal = {ISA Transactions},
volume = {108},
pages = {367-380},
year = {2021},
issn = {0019-0578},
author = {Zhihua Chen and Shoukun Wang and Junzheng Wang and Kang Xu and Tao Lei and Hao Zhang and Xiuwen Wang and Daohe Liu and Jinge Si}
}

@ARTICLE{9794482,
  author={He, Jun and Sun, Yanlong and Yang, Limin and Sun, Jiaze and Xing, Yan and Gao, Feng},
  journal={IEEE/ASME Transactions on Mechatronics}, 
  title={Design and Control of TAWL—A Wheel-Legged Rover With Terrain-Adaptive Wheel Speed Allocation Capability}, 
  year={2022},
  volume={27},
  number={4},
  pages={2212-2223},
  doi={10.1109/TMECH.2022.3176638}}

@ARTICLE{8998367,
  author={Bouton, Arthur and Grand, Christophe and Benamar, Faïz},
  journal={IEEE/ASME Transactions on Mechatronics}, 
  title={Design and Control of a Compliant Wheel-on-Leg Rover Which Conforms to Uneven Terrain}, 
  year={2020},
  volume={25},
  number={5},
  pages={2354-2363},
  doi={10.1109/TMECH.2020.2973752}}

@INPROCEEDINGS{9561579,
  author={Wang, Shuai and Cui, Leilei and Zhang, Jingfan and Lai, Jie and Zhang, Dongsheng and Chen, Ke and Zheng, Yu and Zhang, Zhengyou and Jiang, Zhong-Ping},
  booktitle={2021 IEEE International Conference on Robotics and Automation (ICRA)}, 
  title={Balance Control of a Novel Wheel-legged Robot: Design and Experiments}, 
  year={2021},
  volume={},
  number={},
  pages={6782-6788},
  doi={10.1109/ICRA48506.2021.9561579}}

@ARTICLE{10502024,
  author={Godden, Thomas and Mulvey, Barry W. and Redgrave, Ellen and Nanayakkara, Thrishantha},
  journal={IEEE Robotics and Automation Letters}, 
  title={PaTS-Wheel: A Passively-Transformable Single-Part Wheel for Mobile Robot Navigation on Unstructured Terrain}, 
  year={2024},
  volume={9},
  number={6},
  pages={5512-5519},
  doi={10.1109/LRA.2024.3389828}}

@article{bitnaza,
title = {Whole-body stability control with high contact redundancy for wheel-legged hexapod robot driving over rough terrain},
journal = {Mechanism and Machine Theory},
volume = {181},
pages = {105199},
year = {2023},
issn = {0094-114X},
doi = {https://doi.org/10.1016/j.mechmachtheory.2022.105199},
author = {Kang Xu and Yanqun Lu and Lei Shi and Jianyong Li and Shoukun Wang and Tao Lei},
}

@article{NI2022,
title = {Parameters uncertainty analysis of posture control of a four-wheel-legged robot with series slow active suspension system},
journal = {Mechanism and Machine Theory},
volume = {175},
pages = {104966},
year = {2022},
issn = {0094-114X},
doi = {https://doi.org/10.1016/j.mechmachtheory.2022.104966},
author = {Liwei Ni and Liang Wu and Hongsong Zhang},
}

@inproceedings{rudin2022learning,
  title={Learning to walk in minutes using massively parallel deep reinforcement learning},
  author={Rudin, Nikita and Hoeller, David and Reist, Philipp and Hutter, Marco},
  booktitle={Conference on Robot Learning},
  pages={91--100},
  year={2022},
  organization={PMLR}
}

@article{Lee2024learning,
author = {Joonho Lee  and Marko Bjelonic  and Alexander Reske  and Lorenz Wellhausen  and Takahiro Miki  and Marco Hutter },
title = {Learning robust autonomous navigation and locomotion for wheeled-legged robots},
journal = {Science Robotics},
volume = {9},
number = {89},
pages = {eadi9641},
year = {2024},
doi = {10.1126/scirobotics.adi9641},
}

@ARTICLE{Scaldaferri2025otto,
  author={Scaldaferri, Antonello and Tolomei, Simone and Iotti, Francesco and Gambino, Paolo and Pierallini, Michele and Angelini, Franco and Garabini, Manolo},
  journal={IEEE Open Journal of the Industrial Electronics Society}, 
  title={Otto - Design and Control of an 8-DoF SEA-driven Quadrupedal Robot}, 
  year={2025},
  volume={},
  number={},
  pages={1-21},
}

@book{robot,
  title={Springer handbook of robotics},
  author={Siciliano, Bruno and Khatib, Oussama and Kr{\"o}ger, Torsten},
  volume={200},
  year={2008},
  publisher={Springer}
}

@online{moteus,
author = {J. Pieper},
title = {Mjbots/Moteus },
year = 2018,
url = {https://github.com/mjbots/moteus},
urldate = {2023-06-30}
}

@INPROCEEDINGS{rover_mars,
  author={Lindemann, R.A. and Voorhees, C.J.},
  booktitle={2005 IEEE International Conference on Systems, Man and Cybernetics}, 
  title={Mars Exploration Rover mobility assembly design, test and performance}, 
  year={2005},
  volume={1},
  number={},
  pages={450-455},
}

@inproceedings{Lunar_rover,
  title={ {Mobility performance evaluation of lunar rover and optimization of rocker-bogie suspension parameters}},
  author={ {Li, Suojun and Gao, Haibo and Deng, Zongquan}},
  booktitle={ {2008 2nd International symposium on systems and control in aerospace and astronautics}},
  pages={ {1-6}},
  year={ {2008}},
  organization={ {IEEE}}
}

@INPROCEEDINGS{9305643,
  author={Dosoftei, Catalin and Horga, Vasile and Doroftei, Ioan and Popovici, Tudor and Custura, Stefan},
  booktitle={2020 International Conference and Exposition on Electrical And Power Engineering (EPE)}, 
  title={Simplified Mecanum Wheel Modelling using a Reduced Omni Wheel Model for Dynamic Simulation of an Omnidirectional Mobile Robot}, 
  year={2020},
  volume={},
  number={},
  pages={721-726},
}

@INPROCEEDINGS{Tlale2008,
  author={{Tlale, Nkgatho and de Villiers, Mark}},
  booktitle={{2008 15th International Conference on Mechatronics and Machine Vision in Practice}}, 
  title={{Kinematics and Dynamics Modelling of a Mecanum Wheeled Mobile Platform}}, 
  year={{2008}},
  volume={},
  number={},
  pages={{657-662}},
  doi={{10.1109/MMVIP.2008.4749608}}}

@book{mob_rob_apl,
  title={ {Applications of Mobile Robots}},
  author={ {Hurtado, Efren Gorrostieta}},
  year={ {2019}},
  publisher={ {BoD--Books on Demand}}
}

@article{TAHERI2020103958,
title = {{Omnidirectional mobile robots, mechanisms and navigation approaches}},
author = {{Hamid Taheri and Chun Xia Zhao}},
journal = {{Mechanism and Machine Theory}},
volume = {{153}},
pages = {{103958}},
year = {{2020}},
issn = {{0094-114X}},

}

@article{three_omni,
author = {Yunardi, Riky and Arifianto, Deny and Bachtiar, Farhan and Prananingrum, Jihan},
year = {2021},
month = {03},
pages = {},
title = {Holonomic Implementation of Three Wheels Omnidirectional Mobile Robot using DC Motors},
volume = {2},
journal = {Journal of Robotics and Control (JRC)},

}

@inproceedings{survey,
  title={A survey of wheeled-legged robots},
  author={Bjelonic, Marko and Klemm, Victor and Lee, Joonho and Hutter, Marco},
  booktitle={Climbing and walking robots conference},
  pages={83--94},
  year={2022},
  organization={Springer}
}

@INPROCEEDINGS{1242089,
  author={Loh, W.K. and Low, K.H. and Leow, Y.P.},
  booktitle={2003 IEEE International Conference on Robotics and Automation (Cat. No.03CH37422)}, 
  title={Mechatronics design and kinematic modelling of a singularityless omni-directional wheeled mobile robot}, 
  year={2003},
  volume={3},
  number={},
  pages={3237-3242},
}

@article{NI,
  author={Angelini, Franco and Angelini, Pierangela and Angiolini, Claudia and Bagella, Simonetta and Bonomo, Fabio and Caccianiga, Marco and Santina, Cosimo Della and Gigante, Daniela and Hutter, Marco and Nanayakkara, Thrishantha and Remagnino, Paolo and Torricelli, Diego and Garabini, Manolo},
  journal={IEEE Access}, 
  title={Robotic Monitoring of Habitats: The Natural Intelligence Approach}, 
  year={2023},
  volume={11},
  number={},
  pages={72575-72591},
  doi={10.1109/ACCESS.2023.3294276}}

@INPROCEEDINGS{5980575,
  author={Bischoff, Rainer and Huggenberger, Ulrich and Prassler, Erwin},
  booktitle={2011 IEEE International Conference on Robotics and Automation}, 
  title={KUKA youBot - a mobile manipulator for research and education}, 
  year={2011},
  volume={},
  number={},
  pages={1-4},
}

@article{warehouse_robot,
  title={ {Robotized and automated warehouse systems: Review and recent developments}},
  author={ {Azadeh, Kaveh and De Koster, Ren{\'e} and Roy, Debjit}},
  journal={ {Transportation Science}},
  volume={ {53}},
  number={ {4}},
  pages={ {917-945}},
  year={ {2019}},
  publisher={ {INFORMS}}
}

@article{four_omni,
author = {Kundu, Ananda and Mazumder, Oishee and Lenka, Prasanna and Bhaumik, Subhasis},
year = {2017},
pages = {289-295},
title = {Design and Performance Evaluation of 4 Wheeled Omni Wheelchair with Reduced Slip and Vibration},
volume = {105},
journal = {Procedia Computer Science},

}

@article{mec_apl,
author = {Adascalitei, Florentina and Doroftei, Ioan},
year = {2011},
pages = {21-29},
title = {Practical applications for mobile robots based on Mecanum wheels - a systematic survey},
journal = {Romanian Review Precision Mechanics, Optics and Mechatronics}
}

@misc{ilon1975wheels,
  title={Wheels for a course stable selfpropelling vehicle movable in any desired direction on the ground or some other base},
  author={Ilon, Bengt Erland},
  year={1975},
  month=apr # "~8",
  publisher={Google Patents},
  note={US Patent 3,876,255}
}

@INPROCEEDINGS{5442595,
  author={Ramirez-Serrano, A. and Kuzyk, R.},
  booktitle={2010 Sixth International Conference on Autonomic and Autonomous Systems}, 
  title={Modified Mecanum Wheels for Traversing Rough Terrains}, 
  year={2010},
  volume={},
  number={},
  pages={97-103},
  doi={10.1109/ICAS.2010.35}}

@article{des_opt_mec,
author = {Bae, Jong-Jin and Kang, Namcheol},
year = {2016},
pages = {1-8},
title = {Design Optimization of a Mecanum Wheel to Reduce Vertical Vibrations by the Consideration of Equivalent Stiffness},
volume = {2016},
journal = {Shock and Vibration},
doi = {10.1155/2016/5892784}
}

@article{mec_intro,
author = {Doroftei, Ioan and Stirbu, Bogdan},
year = {2010},
pages = {173-178},
title = {Design, Modeling and Control of an Omni-Directional Mobile Robot},
volume = {166-167},
journal = {Solid State Phenomena},
doi = {10.4028/www.scientific.net/SSP.166-167.173}
}

@article{SUN2021107128,
title = {Path-following control of Mecanum-wheels omnidirectional mobile robots using nonsingular terminal sliding mode},
journal = {Mechanical Systems and Signal Processing},
volume = {147},
pages = {107-128},
year = {2021},
issn = {0888-3270},
author = {Zhe Sun and Hao Xie and Jinchuan Zheng and Zhihong Man and Defeng He},

}

@ARTICLE{6930798,
  author={Hutter, Marco and Gehring, Christian and Höpflinger, Mark A. and Blösch, Michael and Siegwart, Roland},
  journal={IEEE Transactions on Robotics}, 
  title={Toward Combining Speed, Efficiency, Versatility, and Robustness in an Autonomous Quadruped}, 
  year={2014},
  volume={30},
  number={6},
  pages={1427-1440},
  doi={10.1109/TRO.2014.2360493}}

@book{Leg_system,
  title={Legged robotic systems},
  author={Carbone, Giuseppe and Ceccarelli, Marco},
  year={2005},
  publisher={INTECH Open Access Publisher}
}

@ARTICLE{8913831,
  author={Guizzo, Erico},
  journal={IEEE Spectrum}, 
  title={By leaps and bounds: An exclusive look at how Boston dynamics is redefining robot agility}, 
  year={2019},
  volume={56},
  number={12},
  pages={34-39},
  doi={10.1109/MSPEC.2019.8913831}}

@INPROCEEDINGS{7758092,
  author={Hutter, Marco and Gehring, Christian and Jud, Dominic and Lauber, Andreas and Bellicoso, C. Dario and Tsounis, Vassilios and Hwangbo, Jemin and Bodie, Karen and Fankhauser, Peter and Bloesch, Michael and Diethelm, Remo and Bachmann, Samuel and Melzer, Amir and Hoepflinger, Mark},
  booktitle={2016 IEEE/RSJ International Conference on Intelligent Robots and Systems (IROS)}, 
  title={ANYmal - a highly mobile and dynamic quadrupedal robot}, 
  year={2016},
  volume={},
  number={},
  pages={38-44},
  doi={10.1109/IROS.2016.7758092}}

@INPROCEEDINGS{8793865,
  author={Katz, Benjamin and Carlo, Jared Di and Kim, Sangbae},
  booktitle={2019 International Conference on Robotics and Automation (ICRA)}, 
  title={Mini Cheetah: A Platform for Pushing the Limits of Dynamic Quadruped Control}, 
  year={2019},
  volume={},
  number={},
  pages={6295-6301},
  doi={10.1109/ICRA.2019.8793865}}

@article{doi:10.1177/0278364915578839,
author = {Claudio Semini and Victor Barasuol and Thiago Boaventura and Marco Frigerio and Michele Focchi and Darwin G. Caldwell and Jonas Buchli},
title ={Towards versatile legged robots through active impedance control},
journal = {The International Journal of Robotics Research},
volume = {34},
number = {7},
pages = {1003-1020},
year = {2015},
}

@article{Sprwitz2018OncillaRA,
  title={Oncilla Robot: A Versatile Open-Source Quadruped Research Robot With Compliant Pantograph Legs},
  author={Alexander Spr{\"o}witz and Alexandre Tuleu and Mostafa Ajaoolleian and Massimo Vespignani and Rico Moeckel and Peter Eckert and Michiel D'Haene and Jonas Degrave and Arne Nordmann and Benjamin Schrauwen and Jochen J. Steil and Auke Jan Ijspeert},
  journal={Frontiers in Robotics and AI},
  year={2018},
  volume={5},
}

@article{Sprwitz2013TowardsDT,
  title={Towards dynamic trot gait locomotion: Design, control, and experiments with Cheetah-cub, a compliant quadruped robot},
  author={Alexander Spr{\"o}witz and Alexandre Tuleu and Massimo Vespignani and Mostafa Ajallooeian and Emilie Badri and Auke Jan Ijspeert},
  journal={The International Journal of Robotics Research},
  year={2013},
  volume={32},
  pages={932 - 950}
}

@ARTICLE{Em-act,
  author={Gopanunni, Ramesh krishnan Muttathil and Martignetti, Lorenzo and Iotti, Francesco and Ranjan, Alok and Angelini, Franco and Garabini, Manolo},
  journal={IEEE Open Journal of the Industrial Electronics Society}, 
  title={EM-Act: A Modular Series Elastic Actuator for Dynamic Robots}, 
  year={2024},
  volume={5},
  number={},
  pages={468-480}
  }

@article{10.1115/1.4011045,
    author = {Denavit, J. and Hartenberg, R. S.},
    title = "{A Kinematic Notation for Lower-Pair Mechanisms Based on Matrices}",
    journal = {Journal of Applied Mechanics},
    volume = {22},
    number = {2},
    pages = {215-221},
    year = {2021},
    month = {06}

}

@article{lin2013,
author = {Lin, Lih-Chang and Shih, Hao-Yin},
year = {2013},
month = {01},
pages = {166-179},
title = {Modeling and Adaptive Control of an Omni-Mecanum-Wheeled Robot},
volume = {04},
journal = {Intelligent Control and Automation},
doi = {10.4236/ica.2013.42021}
}

@inproceedings{gor2013,
  title={Dynamic modeling and simulation of compliant legged quadruped robot},
  author={Gor, Mehul and Pathak, Pushparaj Mani and Kumar Samantaray, A and Yang, JM},
  booktitle={proceedings of the 1st international and 16th national conference on machines and mechanisms (iNaCoMM2013), IIT Roorkee, India},
  pages={7--11},
  year={2013}
}

@article{review_quadruped,
title = {A study on quadruped mobile robots},
journal = {Mechanism and Machine Theory},
volume = {190},
pages = {105448},
year = {2023},
issn = {0094-114X},
doi = {https://doi.org/10.1016/j.mechmachtheory.2023.105448},
author = {Hamid Taheri and Nasser Mozayani},
}

@article{ZHANG201518,
title = {Research on a tracked omnidirectional and cross-country vehicle},
journal = {Mechanism and Machine Theory},
volume = {87},
pages = {18-44},
year = {2015},
issn = {0094-114X},
doi = {https://doi.org/10.1016/j.mechmachtheory.2014.12.016},
author = {Yunan Zhang and Tao Huang},
}

@ARTICLE{Zheng2024,
  author={Zheng, Chuanqi and Sane, Siddharth and Lee, Kangneoung and Kalyanram, Vishnu and Lee, Kiju},
  journal={IEEE Transactions on Robotics}, 
  title={$\mathbf{\alpha }$-WaLTR: Adaptive Wheel-and-Leg Transformable Robot for Versatile Multiterrain Locomotion}, 
  year={2023},
  volume={39},
  number={2},
  pages={941-958},
}
\vspace{-12mm}

\vspace{-12mm}

\end{document}